\documentclass[twoside,11pt]{article}

\usepackage{jmlr2e}

\jmlrheading{18}{2017}{1-52}{9/15; Revised 3/17}{12/17}{15-464}{Michael Riis Andersen, Aki Vehtari, Ole Winther and Lars Kai Hansen}

\ShortHeadings{Bayesian Inference for Spatio-temporal Spike-and-Slab Priors}{Andersen, Vehtari, Winther, and Hansen}
\firstpageno{1}
\usepackage{amsmath}
\usepackage{bm}
\usepackage{xcolor}

\newcommand{\ab}{\bm{a}}
\newcommand{\bb}{\bm{b}}
\newcommand{\e}{\bm{e}}
\newcommand{\m}{\bm{m}}
\newcommand{\p}{\bm{p}}
\newcommand{\x}{\bm{x}}
\newcommand{\y}{\bm{y}}
\newcommand{\z}{\bm{z}}

\newcommand{\A}{\bm{A}}
\newcommand{\B}{\bm{B}}
\newcommand{\C}{\bm{C}}

\newcommand{\I}{\bm{I}}
\newcommand{\M}{\bm{M}}

\newcommand{\R}{\bm{R}}
\newcommand{\Ss}{\bm{S}}
\newcommand{\U}{\bm{U}}
\newcommand{\X}{\bm{X}}
\newcommand{\Y}{\bm{Y}}
\newcommand{\Z}{\bm{Z}}

\newcommand{\N}{\mathcal{N}}

\newcommand{\dd}{\text{d}}
\newcommand{\E}{\mathbb{E}}
\newcommand{\V}{\mathbb{V}}
\newcommand{\KL}{\text{KL}}
\DeclareMathOperator{\argmin}{\text{argmin}\;}

\usepackage{subfigure}
\usepackage{placeins}
\usepackage{amsmath}
\usepackage{lipsum}
\usepackage{booktabs}
\usepackage[plain]{algorithm}
\usepackage{microtype}

\DeclareMathOperator{\Ber}{Ber}

\newcommand{\dtuaddr}{\addr Department of Applied Mathematics and Computer Science\\
       Technical University of Denmark\\
DK-2800 Kgs. Lyngby, Denmark}

\begin{document}

\title{Bayesian Inference for Spatio-temporal Spike-and-Slab Priors}

\author{\name Michael Riis Andersen\thanks{Work done mainly while at Department of Applied Mathematics and Computer Science,  Technical University of Denmark}  \email michael.andersen@aalto.fi \\
		\name Aki Vehtari \email Aki.Vehtari@aalto.fi\\
		\addr Helsinki Institute for Information Technology HIIT\\
		Department of Computer Science, Aalto University\\
		P.O. Box 15400, FI-00076, Finland\\
		\AND
		\name Ole Winther \email olwi@dtu.dk \\
		\name Lars Kai Hansen\email lkai@dtu.dk \\
		\dtuaddr
       }

\editor{Lawrence Carin}

\maketitle

\begin{abstract}
In this work, we address the problem of solving a series of underdetermined linear inverse problemblems subject to a sparsity constraint. We generalize the spike-and-slab prior distribution to encode a priori correlation of the support of the solution in both space and time by imposing a transformed Gaussian process on the spike-and-slab probabilities. 
An expectation propagation (EP) algorithm for posterior inference under the proposed model is derived. For large scale problems, the standard EP algorithm can be prohibitively slow. We therefore introduce three different approximation schemes to reduce the computational complexity. Finally, we demonstrate the proposed model using numerical experiments based on both synthetic and real data sets. 
\end{abstract}


\section{Introduction}
\label{sec:introduction}
Many problems of practical interest in machine learning involve a high dimensional feature space and a relatively small number of observations. Inference is in general difficult for such underdetermined problems due to high variance and therefore regularization is often the key to extracting meaningful information from such problems \citep{Tibshirani94regressionshrinkage}. The classical approach is Tikhonov regularization (also known as $\ell_2$ regularization), but during the last few decades sparsity has been an increasingly popular choice of regularization for many problems, giving rise to methods such as the LASSO \citep{Tibshirani94regressionshrinkage}, Sparse Bayesian Learning \citep{Tipping:2001:SBL:944733.944741} and sparsity promoting priors \citep{mitchell1988a}.
\\
\\
In this work, we address the problem of finding sparse solutions to linear inverse problems of the form
\begin{align}\label{eq:intro_smv_problem}
	\y = \A\x + \e,
\end{align}
where $\x \in \mathbb{R}^D$ is the desired solution, $\y \in \mathbb{R}^N$ is an observed measurement vector, $\A \in \mathbb{R}^{N \times D}$ is a known forward model and $\e \in \mathbb{R}^N$ is additive measurement noise. We are mainly interested in the underdetermined regime, where the number of observations is smaller than the number of unknowns, that is $N < D$. In the sparse recovery literature, it has been shown that the sparsity constraint is crucial for recovering $\x$ from a small set of linear measurements \citep{candes2006robust}. Furthermore, the ratio between the number non-zero coefficients $K = \|\x\|_0$ and the dimension $D$ dictates the required number of measurements $N$ for robust reconstruction of $\x$ and this relationship has given rise to so-called \textit{phase transition curves} \citep{journals/pieee/DonohoT10}. A large body of research has been dedicated to improve these phase transition curves and these endeavors have lead to the concepts of \textit{multiple measurement vectors} \citep{cotter2005sparse} and \textit{structured sparsity} \citep{Huang:2009:LSS:1553374.1553429}. 
\\
\\
The multiple measurement vector problem (MMV) is a natural extension of eq. \eqref{eq:intro_smv_problem}, where multiple measurements $\y_1, \y_2, \hdots, \y_T$ are observed and assumed to be generated from a series of signals $\x_1, \x_2, \hdots, \x_T$, which share a common sparsity pattern. In matrix notation, we can write the problem as 
\begin{align} \label{eq:intro_mmv}
	\Y = \A\X + \bm{E},
\end{align}
where the desired solution is now a matrix $\X = \begin{bmatrix}\x_1&\x_2&\hdots&\x_T\end{bmatrix} \in \mathbb{R}^{D\times T}$ and similarly for the measurement matrix $\Y \in \mathbb{R}^{N \times T}$ and the noise term $\bm{E} \in \mathbb{R}^{N \times T}$. The assumption of \textit{joint sparsity} allows one to recover $\X$ with significantly fewer observations compared to solving each of the $T$ inverse problems in eq. \eqref{eq:intro_smv_problem} separately \citep{cotter2005sparse}. The MMV approach has also been generalized to problems, where the sparsity pattern is evolving slowly in time \citep{ziniel2013a}. Structured sparsity, on the other hand, is a generalization of simple sparsity and seeks to exploit the fact that the sparsity patterns of many natural signals contain a richer structure than simple sparsity, for example, \textit{group sparsity} \citep{Jacob:2009:GLO:1553374.1553431} or \textit{cluster structured sparsity} \citep{Yu2012259}.
\\
\\
In this paper, we combine these two approaches and focus on problems, where the sparsity pattern of $\X$ exhibits a spatio-temporal structure. In particular, we assume that the row and column indices of $\X$ can be associated with a set of spatial and temporal coordinates, respectively. This can equivalently be interpreted as a sparse linear regression problem, where the support of the regressors is correlated in both space and time. Applications of such a model include dynamic compressed sensing \citep{ziniel2013a}, background subtraction in computer vision \citep{NIPS2008_3487} and EEG source localization problem \citep{baillet2001a}.
\\
\\
We take a Bayesian approach to modeling this structure since it provides a natural way of incorporating such prior knowledge in a model. In particular, we propose a hierarchical probabilistic model for $\X$ based on the so-called spike-and-slab prior \citep{mitchell1988a}. We introduce a smooth latent variable controlling the spatio-temporal structure of the support of $\X$ by extending the work by \citet{NIPS2014_5464}. We aim for full Bayesian inference under the proposed probabilistic model, but inference w.r.t. the exact posterior distribution of interest is intractable. Instead we resort to approximate inference using Expectation Propagation \citep{Minka01, opper2000gaussian}, which has been shown to provide accurate inference for spike-and-slab priors \citep{JMLR:v14:hernandez-lobato13a, hernandez-lobato2010a,JMLR:v15:jylanki14a,Peltola2014-gy}. Our model formulation is generic and generalizes easily to other types of observations. In particular, we also combine the proposed prior with a probit observation model to model binary observations in a sparse linear classification setting. 
\\
\\
The contribution of this paper is three-fold. First we extend the structured spike-and-slab prior and the associated EP inference scheme to incorporate both spatial and temporal smoothness of the support. However, the computational complexity of the resulting EP algorithm is prohibitively slow for problems of even moderate sizes of signal dimension $D$ and length $T$. To alleviate the computational bottleneck of the EP algorithm we propose three different approximation schemes. Finally, we discuss several approaches for learning the hyperparameters and evaluate them based on synthetic and real data sets.

\subsection{Related Work}
In this section, we briefly review some of the most common approaches to simple sparsity and their generalization to structured sparsity. The classical approach to sparsity is the LASSO  \citep{Tibshirani94regressionshrinkage}, which operates by optimizing a least squares cost function augmented with an $\ell_1$ penalty on the regression weights. Several extensions have been proposed in the literature to generalize the LASSO to the structured sparsity setting, examples include group and graph LASSO \citep{Jacob:2009:GLO:1553374.1553431}. From a probabilistic perspective sparsity can be encouraged through the use of \textit{sparsity-promoting priors}. 
A non-exhaustive list of sparsity-promoting priors includes the Laplace prior \citep{park2008bayesian}, Automatic Relevance Determination prior \citep{Neal:1996a}, the horseshoe prior \citep{journals/jmlr/CarvalhoPS09} and the spike-and-slab prior \citep{mitchell1988a}. All of these were originally designed to enforce simply sparsity, but they have all been generalized to the structured sparsity setting. The general strategy is to extend univariate densities to correlated multivariate densities by augmenting the models with a latent multivariate variable, where the correlation structure can be controlled explicitly, for example, using Markov Random Fields \citep{NIPS2008_3487, miguel2011a} or multivariate Gaussian distributions \citep{Engelhardt2014-ln}. Here we limit ourselves to consider the latter.
\\
\\
From a probabilistic perspective, optimizing with an $\ell_1$ regularization term can be interpreted as maximum a posteriori (MAP) inference under an i.i.d. Laplace prior distribution on the regression weights \citep{park2008bayesian}. The univariate Laplace prior has been generalized to the multivariate Laplace (MVL) distribution, which couples the prior variance of the regression weights through a scale mixture formulation \citep{NIPS2009_3751}. 
\\
\\
Another approach is Automatic Relevance Determination (ARD) \citep{Neal:1996a}, which works by imposing independent zero mean Gaussian priors with individual precision parameters on the regression weights. These precision parameters are then optimized using a maximum likelihood type II and the idea is then that the precision parameters of irrelevant features will approach infinity and thereby forcing the weights of the irrelevant features to zero. \citet{wu2014sparse} extend the ARD framework to promote spatial sparsity by introducing a latent multivariate Gaussian distribution to impose spatial structure onto the precision parameters of ARD giving rise to \textit{dependent relevance determination priors}.
\\
\\
The horseshoe prior is defined as a scale mixture of Gaussians, where a half-Cauchy distribution is used as prior for the standard deviation of the Gaussian density \citep{journals/jmlr/CarvalhoPS09}. The resulting density has two very appealing properties for promoting sparsity, namely heavy tails and an infinitely large spike at zero. A generalization to the multivariate case has been proposed by \citet{NIPS2013_5212}. 
\\
\\
The spike-and-slab prior is an increasingly popular choice of sparsity promoting prior and is given by a binary mixture of two components: a Dirac delta distribution (spike) at zero and Gaussian distribution (slab) \citep{mitchell1988a,Carbonetto2012-tl}. The spike-and-slab prior has been generalized to the group setting by \citet{JMLR:v14:hernandez-lobato13a}, to clustered sparsity setting by \citet{Yu2012259} and spatial structures by \citet{NIPS2014_5464}, \citet{BIOM:BIOM12126}, and \citet{Engelhardt2014-ln}. \citet{BIOM:BIOM12126} induce the spatial structure using basis functions and \citet{NIPS2014_5464} impose the structure using a multivariate Gaussian density. The latter is the starting point of this work.
\\
\\
Our work is closely related to the work on the multivariate Laplace prior (MVL) \citep{NIPS2009_3751} as mentioned above and the work on the network-based sparse Bayesian classification algorithm (NBSBC) \citep{miguel2011a}. The former also uses EP for approximating the posterior distribution of a Gaussian linear model with the MVL prior, where the structure of the support is encoded into the model using a sparse precision matrix. The NBSBC method also uses EP to approximate the posterior distribution of linear model with coupled spike-and-slab priors, but the structure of the support is encoded in a network using a Markov Random Field (MRF) prior. In contrast, we can inject a priori knowledge of the structure into the model using generic covariance functions rather than clique potentials as in the MRF-based models, which makes it easier to interpret interesting quantities like the characteristic lengthscale etc.

\subsection{Structure of Paper}
This paper is organized as follows. In Section \ref{sec:structured_spike_and_slab} we review the structured spike-and-slab prior and in Section \ref{sec:spatio-temporal_spike_and_slab} we discuss different ways of extending the model to include the temporal structure as well. After introducing the models we propose an algorithm for approximate inference based on the expectation propagation (EP) framework. We review the basics of EP and describe the proposed algorithm in Section \ref{sec:inference}. In Section \ref{sec:approximations} we introduce three simple approximation schemes to speed of the inference process and discuss their properties. Finally, in Section \ref{sec:experiments} we demonstrate the proposed method using synthetic and real data sets. 

\subsection{Notation}
We use bold uppercase letters to denote matrices and bold lowercase letters to denote vectors. Unless stated otherwise, all vectors are column vectors. Furthermore, we use the notation $\bm{a}_{n,\cdot} \in \mathbb{R}^{1 \times D}$ and $\bm{a}_{
\cdot,i} \in \mathbb{R}^{N \times 1}$ for the $n$'th row and $i$'th column in the matrix $\A \in \mathbb{R}^{N \times D}$, respectively. $\left[K\right]$ denotes the set of integers from $1$ to $K$, that is $\left[K\right] = \left\lbrace 1, 2, .., K\right\rbrace$. We use the notation $\ab \circ \bb$ to denote the element-wise Hadamard product of $\ab$ and $\bb$ and $\A \otimes \B \in \mathbb{R}^{MN \times MN}$ for the Kronecker product of matrices $\A \in \mathbf{R}^{M \times M}$ and $\B \in \mathbf{R}^{N \times N}$.  We use $\N\left(\x|\m,\bm{V}\right)$ to denote a multivariate Gaussian density over $\x$ with mean vector $\m$ and covariance matrix $\bm{V}$ and $\Ber\left(z|p\right)$ denotes a Bernoulli distribution on $z$ with probability of $p(z = 1) = p$.

\section{The Structured Spike-and-Slab Prior}
\label{sec:structured_spike_and_slab}
The purpose of this section is to describe the \textit{structured spike-and-slab prior} \citep{NIPS2014_5464}, but first we briefly review the conventional spike-and-slab prior \citep{mitchell1988a}. For $\x \in \mathbb{R}^D$, the spike-and-slab prior distribution is given by
\begin{align} \label{eq_spike_and_slab}
p(\x\big|p_0, \rho_0, \tau_0)  = \prod_{i=1}^D \left[(1-p_0)\delta\left(x_i\right) + p_0\N\left(x_i| \rho_0, \tau_0\right)\right],
\end{align}
where $\delta\left(x\right)$ is the Dirac delta function and $p_0, \rho_0$ and $\tau_0$ are hyperparameters. In particular, $p_0$ is the prior probability of a given variable being active, that is $p(x_i \neq 0) = p_0$, and $\rho_0, \tau_0$ are the prior mean and variance, respectively, of the active variables. The spike-and-slab prior in eq. \eqref{eq_spike_and_slab} is also known as the Bernoulli-Gaussian prior since the prior can decomposed as
\begin{align} \label{eq_bern_gauss}
p(\x\big| p_0, \rho_0, \tau_0)  &= \sum_{\z} \prod_{i=1}^D \left[(1-z_i)\delta\left(x_i\right) + z_i\N\left(x_i| \rho_0, \tau_0\right)\right]\prod_{i=1}^D \text{Ber}\left(z_i|p_0\right),
\end{align}
where the sum is over all the binary variables $z_i$ for $i \in \left[D\right]$. Thus, the latent binary variable $z_i \in \left\lbrace 0, 1 \right\rbrace$ can interpreted as an indicator variable for the event $x_i \neq 0$. We will refer to $\z$ as the \textit{sparsity pattern} or the \textit{support} of $\x$. In eq. \eqref{eq_spike_and_slab} and \eqref{eq_bern_gauss} we condition explicitly on the hyperparameters $p_0, \rho_0, \tau_0$, but to ease the notation we will omit this in the remainder of this paper.
\\
\\
The variables $x_i$ and $x_j$ are assumed to be independent for $i \neq j$ as seen in eq. \eqref{eq_spike_and_slab} and \eqref{eq_bern_gauss}. This implies that the number of active variables follows a binomial distribution and hence, the marginal probability of $x_i$ and $x_j$ being jointly active, is given by $p(x_i \neq 0, x_j \neq 0) = p_0^2$ for all $i \neq j$. However, in many applications the variables $\left\lbrace x_k \right\rbrace_{k=1}^D$ might a priori have an underlying topographic relationship such as a spatial or temporal structure. Without loss of generality we will assume a spatial relationship, where $\bm{d}_i$ denotes the spatial coordinates of $x_i$. For such models, it is often a reasonable assumption that $p(x_i \neq 0, x_j \neq 0)$ should depend on $\|\bm{d}_i - \bm{d}_j\|$. For instance, neighboring voxels in functional magnetic resonance imaging (fMRI) analysis \citep{citeulike:791498} are often more likely to be active simultaneously compared to two voxels far apart. Such a priori knowledge is neglected by the conventional spike-and-slab prior in eq. \eqref{eq_spike_and_slab}.
\\
\\
The structured spike-and-slab model is capable of modeling such structure and is given in terms of a hierarchical model
\begin{align}
p(\x\big|\z) &= \prod_{i=1}^D \left[\left(1-z_i\right)\delta\left(x_i\right) + z_i\N\left(x_i\big|\rho_0, \tau_0\right)\right], \label{eq:prior_on_x}\\
p(\z\big|\bm{\gamma}) &= \prod_{i=1}^D \Ber\left(z_i\big|\phi\left(\gamma_i\right)\right), \quad\quad \phi: \mathbb{R} \rightarrow \left(0, 1\right),\label{eq:prior_on_z}\\
p(\bm{\gamma}) &= \N\left(\bm{\gamma}\big|\bm{\mu}_0, \bm{\Sigma}_0\right), \label{eq:prior_on_gamma}
\end{align}
where $\bm{\gamma}$ is a latent variable controlling the structure of the sparsity pattern. Using this model prior knowledge of the structure of the sparsity pattern can be encoded using $\bm{\mu}_0$ and $\bm{\Sigma}_0$. The mean value $\bm{\mu}_0$ controls the expected degree of sparsity and the covariance matrix $\bm{\Sigma}_0$ determines the prior correlation of the support. The map $\phi: \mathbb{R} \rightarrow \left(0, 1\right)$ serves the purpose of squeezing $\gamma_i$ into the unit interval and thereby $\phi\left(\gamma_i\right)$ represents the probability of $z_i = 1$. Here we choose $\phi$ to be the standard normal cumulative distribution function (CDF), but other choices, such as the logistic function, are also possible.
\\
\\
Using this formulation, the marginal prior probability of the $i$'th variable being active is given by
\begin{align}
p(z_i = 1) = \int p(z_i = 1\big|\gamma_i)p(\gamma_i) \text{d} \gamma_i = \int \phi(\gamma_i) \N\left(\gamma_i\big|\mu_i, \Sigma_{0, ii}\right) \text{d} \gamma_i = \phi\left(\frac{\mu_i}{\sqrt{1+\Sigma_{0, ii}}}\right) \label{eq:marginal_prior_prob}.
\end{align}
From this expression it is seen that when $\mu_i = 0$, the prior belief of $z_i$ is unbiased and $p(z_i = 1) = 0.5$, but when $\mu_i < 0$ the variable $z_i$ is biased toward zero and vice versa. If a subset of features $\left\lbrace x_j | j \in \mathcal{J} \subset \left[D\right] \right\rbrace$ is a priori more likely to explain the observed data $\y$, then this information can be encoded in the prior distribution by assigning the prior mean of $\bm{\gamma}$ such that $\mu_j > \mu_i$ for all $j \in \mathcal{J}$ and for all $i \notin \mathcal{J}$. However, in the remainder of this paper we will assume that the prior mean is constant, that is $\mu_i = \nu_0$ for some $\nu_0 \in \mathbb{R}$. For more details on the prior distribution, see Appendix \ref{appendix:prior}.
\\
\\
The prior probability of two variables, $x_i$ and $x_j$, being jointly active is
\begin{align}
p(z_i = 1, z_j = 1) = \int \phi(\gamma_i)\phi\left(\gamma_j\right)\N\left(\bm{\gamma}\big|\bm{\mu}, \bm{\Sigma}_0\right) \text{d} \bm{\gamma}.
\end{align}
If $\bm{\Sigma}_0$ is a diagonal matrix, $\gamma_i$ and $\gamma_j$ become independent and we recover the conventional spike-and-slab prior. On the other hand, if we choose $\bm{\Sigma}_0$ to be a covariance matrix of the form $\Sigma_{0,ij} = g\left(\|\bm{d}_i - \bm{d}_j\|\right)$, we see that the joint activation probabilities indeed depend on the spatial distance as desired. Finally, we emphasize that this parametrization it not limited to nearest neighbors-type structures. In fact, this parametrization supports general structures that can be modeled using generic covariance functions. 

\section{The Spatio-temporal Spike-and-Slab Prior}
\label{sec:spatio-temporal_spike_and_slab}
In the following we will extend the structured spike-and-slab prior distribution to model temporal smoothness of the sparsity pattern as well. Let $t \in \left[T\right]$ be the time index, then $\x_t$, $\z_t$ and $\bm{\gamma}_t$ are the signal coefficients, the sparsity patterns and the latent structure variable at time $t$. Furthermore, we define the corresponding matrix quantities $\X = \begin{bmatrix}\x_1&\x_2&\x_T\end{bmatrix}$, $\Z = \begin{bmatrix}\z_1&\z_2&\z_T\end{bmatrix}$ and $\bm{\Gamma} = \begin{bmatrix}\bm{\gamma}_1&\bm{\gamma}_2&\hdots&\bm{\gamma}_T\end{bmatrix}$. 
\\
\\
There are several natural temporal extensions of the model. The simplest extension is to assume that $\left\lbrace \bm{\gamma}_t \right\rbrace_{t=1}^T$ is independent in time, so that $p(\Z, \bm{\Gamma}) = \prod_{t=1}^T p(\z_t\big|\bm{\gamma}_t)\prod_{t=1}^T p(\bm{\gamma}_ t)$, which is equivalent to solving each of the $T$ regressions problems in eq. \eqref{eq:intro_smv_problem} independently. Another simple extension is to use the so-called \textit{joint sparsity} assumption \citep{cotter2005sparse,zhang2011sparse,ziniel2013efficient} and assume that the sparsity pattern is static across time, and thus all $\left\lbrace \x_t \right\rbrace_{t=1}^T$ vectors share a common binary support vector $\z$, and $p(\X\big|\z) = \prod_{t=1}^T\prod_{i=1}^D \left[\left(1-z_i\right)\delta\left(x_{i, t}\right) + z_i\N\left(x_{i, t}\big|\rho_0, \tau_0\right)\right]$. A more interesting and flexible model is to assume that the support is slowly changing in time, by modelling the temporal evolution of $\bm{\gamma}_t$ using a first order Gauss-Markov process of the form $p\left(\bm{\gamma}_t\big|\bm{\gamma}_{t-1}\right) = \N\left(\bm{\gamma}_t\big|\left(1-\alpha\right)\bm{\mu}_0 + \alpha \bm{\gamma}_{t-1}, \beta \bm{\Sigma}_0\right)$, where the hyperparameters $\alpha \in \left[0, 1\right]$ and $\beta > 0$ control the temporal correlation and the ``innovation'' of the process, respectively. 
\\
\\
The first order model has the advantage that it factorizes across time, which makes the resulting inference problem much easier. On the other hand, first order Markovian dynamics is often not sufficient for capturing long range correlations. Imposing a Gaussian process distribution on $\bm{\Gamma}$ with arbitrary covariance structure would facilitate modeling of long range correlations in both time and space. Therefore, the hierarchical prior distribution for $\X$ becomes
\begin{align}\label{eq:kronecker_prior_start}
p(\X\big| \Z) &= \prod_{t=1}^T\prod_{i=1}^D \left[\left(1-z_{i, t}\right)\delta\left(x_{i, t}\right) + z_{i, t}\N\left(x_{i, t}\big|\rho_0, \tau_0\right)\right],\\
p(\Z\big|\bm{\Gamma})&=\prod_{t=1}^T \Ber\left(\z_t\big|\phi\left(\bm{\gamma}_t\right)\right),\\
p(\bm{\Gamma}) &= \N\left(\bm{\Gamma}\big|\bm{\mu}_0, \bm{\Sigma}_{0}\right),\label{eq:kronecker_prior_end}
\end{align}
where the mean $\bm{\mu}_0 \in \mathbb{R}^{TD \times 1}$ and covariance matrix $\bm{\Sigma}_0 \in \mathbb{R}^{TD \times TD}$ are now defined for the full $\bm{\Gamma}$-space. This model is more expressive, but the resulting inference problem becomes infeasible for even moderate sizes of $D$ and $T$. But if we assume that the underlying spatio-temporal grid can be written in Cartesian product form, then covariance matrix simplifies to a Kronecker product
\begin{align} 
p(\bm{\Gamma}) = \N\left(\bm{\Gamma}\big|\bm{\mu}_0, \bm{\Sigma}_{\text{temporal}} \otimes \bm{\Sigma}_{\text{spatial}}\right),
\end{align}
where $\bm{\Sigma}_{\text{temporal}} \in \mathbb{R}^{T \times T}$ and $\bm{\Sigma}_{\text{spatial}} \in \mathbb{R}^{D \times D}$.
This decomposition leads to more efficient inference schemes as we will discuss in Section \ref{sec:approximations}. In the remainder of the paper, we will focus on the model with Kronecker structure, but we refer to \citep{Andersen2015} for more details on the first order model and joint sparsity model.
\\
\\
The coefficients $\left\lbrace x_{i, t} \right\rbrace$ are conditionally independent given the support $\left\lbrace z_{i,t}  \right\rbrace$. For some applications it could be desirable to impose either spatial smoothness, temporal smoothness or both on the non-zero coefficients themselves \citep{NIPS2014_5233,ziniel2013a}, but in this work we only assume  a priori knowledge of the structure of the support. Although temporal smoothness of $x_{i,t}$ could easily be incorporated into the models described above.

\section{Inference Using Spatio-temporal Priors}
\label{sec:inference}
In the previous sections we have described the structured spike-and-slab prior and how to extend it to model temporal smoothness as well. We now turn our attention on how to perform inference using these models. We focus our discussion on the most general formulation using as given in eq. \eqref{eq:kronecker_prior_start}-\eqref{eq:kronecker_prior_end}. Let $\Y = \begin{bmatrix}
\y_1&\y_2&\hdots&\y_T\end{bmatrix}$ be an observation matrix, where $\y_t \in \mathbb{R}^N$ is an observation vector for time $t$. We assume that the distribution on $\Y$ factors over time and is given by
\begin{align}
p(\Y\big|\X) = \prod_{t=1}^T p(\y_t\big|\x_t).
\end{align}
We consider two different noise models: an isotropic Gaussian noise model and a probit noise model. The Gaussian noise model $p(\y_t\big|\x_t) = \N\left(\y_t\big|\A\x_t, \sigma_2 \I\right)$ is suitable for linear inverse problems with forward model $\A \in \mathbb{R}^{N \times D}$ or equivalently sparse linear regression problems with design matrix $\A \in \mathbb{R}^{N \times D}$.  On the other hand, the probit model is suitable for modeling binary observations, with $y_{t,n} \in \left\lbrace -1, 1\right\rbrace$, and is given by $p(\y_t\big|\x_t) = \prod_{n=1}^N \phi\left(y_{t,n}\ab_{n,\cdot} \x_t\right)$, where $\ab_{n,\cdot}$ is the $n$'th row of $\A$.  For both models we further assume that the matrix $\A$ is constant across time. However, this assumption can be easily relaxed to have $\A$ depend on $t$.
\\
\\
For both noise models the resulting joint distribution becomes
\begin{align} \label{eq:general_joint}
p\left(\Y, \X, \Z, \bm{\Gamma}\right) &= p(\Y\big|\X)p(\X\big|\Z)p(\Z\big|\bm{\Gamma})p(\bm{\Gamma})\\
&= \prod_{t=1}^T p(\y_t\big|\x_t)\prod_{t=1}^T \left[\left(1-\z_t\right)\circ\delta\left(\x_t\right) + \z_t\circ\N\left(\x_t\big|0, \tau \I\right)\right]\nonumber\\
&\quad\prod_{t=1}^T \Ber\left(\z_t\big|\phi\left(\bm{\gamma}_t\right)\right)\N\left(\bm{\Gamma}\big|\bm{\mu}_0, \bm{\Sigma}_{0} \right).
\end{align}
We seek the posterior distribution of the parameters $\X, \Z$ and $\bm{\Gamma}$ conditioned on the observations $\Y$, which is obtained by applying Bayes's Theorem to the joint distribution in eq. \eqref{eq:general_joint}
\begin{align}
p\left(\X, \Z, \bm{\Gamma}\big|\Y\right)&= \frac{1}{Z}\prod_{t=1}^T p(\y_t\big|\x_t)\prod_{t=1}^T \left[\left(1-\z_t\right)\circ\delta\left(\x_t\right) + \z_t\circ\N\left(\x_t\big|0, \tau \I\right)\right]\nonumber\\
&\quad\prod_{t=1}^T \Ber\left(\z_t\big|\phi\left(\bm{\gamma}_t\right)\right)\N\left(\bm{\Gamma}\big|\bm{\mu}_0, \bm{\Sigma}_0\right),
\end{align}
where $Z = p(\Y)$ is the marginal likelihood of $\Y$. Due to the product of mixtures in the distribution $p(\X\big|\Z)$, the expression for the marginal likelihood $Z$ involves a sum over $2^{DT}$ terms. This renders the computation of the normalization constant $Z$ intractable for even small $D$ and $T$. Hence, the desired posterior distribution is also intractable and we have to resort to approximate inference.
\\
\\
In the literature researchers have applied a whole spectrum of approximate inference methods for spike-and-slab priors, for example, Monte Carlo-methods \citep{mitchell1988a}, mean-field variational inference \citep{titsias2011a}, approximate message passing \citep{vila2013expectation} and expectation propagation \citep{JMLR:v14:hernandez-lobato13a,NIPS2014_5464}. We use the latter since expectation propagation has been shown to have good performance for linear models with spike-and-slab priors \citep{DBLP:journals/ml/Hernandez-Lobato15} and it has been shown to provide a much better approximation of the first and second moment posterior moment for spike-and-slab models \citep{Peltola2014-gy}.

\subsection{The Expectation Propagation Framework}
In this section, we briefly review expectation propagation for completeness. Expectation propagation (EP) \citep{Minka01, opper2000gaussian} is a deterministic framework for approximating probability distributions. Consider a probability distribution over the variable $\x \in \mathbb{R}^D$ that factorizes into $N$ components
\begin{align}
f(\x) = \prod_{i=1}^N f_i(\x_{i}),
\end{align}
where $\x_i$ is taken to be a subvector of $\x$. EP takes advantage of this factorization and approximates $f$ with a distribution $Q$ that shares the same factorization
\begin{align}
Q(\x) = \prod_{i=1}^N \tilde{f}_i(\x_{i}).
\end{align}
EP approximates each \textit{site term} $f_i$ with a (scaled) distribution $\tilde{f}_i$ from the exponential family. Since the exponential family is closed under products, the \textit{global approximation} $Q$ will also be in the exponential family. Consider the product of all $\tilde{f}_i$ terms except the $j$'th term
\begin{align}\label{eq:cavity}
Q^{\backslash j}(\x) = \prod_{i \neq j} \tilde{f}_i(\x_{i}) = \frac{Q(\x)}{\tilde{f}_j(\x_{j})}.
\end{align}
The core of the EP framework is to choose $\tilde{f_j}$ such that $\tilde{f}_j(\x_j) Q^{\backslash j}(\x_j) \approx f_j\left(\x_j\right)Q^{\backslash j}(\x_j)$. By approximating $f_j$ with $\tilde{f}_j$ in the context of $Q^{\backslash j}$, we ensure that the approximation is most accurate in the region of high density according to the \textit{cavity distribution} $Q^{\backslash j}$. This scheme is implemented by iteratively minimizing the Kullbach-Leibler divergence $\KL\left(f_j\left(\x_j\right)Q^{\backslash j}(\x)\big|\big|\tilde{f}_j\left(\x_j\right)Q^{\backslash j}(\x)\right)$.
Since $\tilde{f}_j\left(\x_j\right)Q^{\backslash j}(\x)$ belongs to the exponential family, the unique solution is obtained by matching the expected sufficient statistics \citep{bishop2006a}. Once the solution,
\begin{align} \label{eq:kl_optim}
Q^* = \underset{q}{\argmin} \KL\!\left(f_j\left(\x_j\right)Q^{\backslash j}(\x)\big|\big|q\right),
\end{align}
is obtained, the $j$'th site approximation is updated as
\begin{align} \label{eq:ep_site_update}
\tilde{f}^*_j\left(\x_j\right) \propto \frac{Q^*\left(\x\right)}{Q^{\backslash j}(\x)}.
\end{align}
The steps in eq. \eqref{eq:cavity}, \eqref{eq:kl_optim} and \eqref{eq:ep_site_update} are repeated sequentially for all $j \in \left[D\right]$ until convergence is achieved.

\subsection{The Expectation Propagation Approximation}
The EP framework provides flexibility in the choice of the approximating factors.  This choice is a trade-off between analytical tractability and sufficient flexibility for capturing the important characteristics of the true density. Consider the desired posterior density of interest
\begin{align} \label{eq:general_posterior}
p\left(\X, \Z, \bm{\Gamma}\big|\Y\right)&\propto \underbrace{\prod_{t=1}^T p(\y_t\big|\x_t)}_{f_1\left(\X\right)}\underbrace{\prod_{t=1}^T \left[\left(1-\z_t\right)\circ\delta\left(\x_t\right) + \z_t\circ\N\left(\x_t\big|0, \tau \I\right)\right]}_{f_2\left(\X, \Z\right)}\nonumber\\
&\quad\underbrace{\prod_{t=1}^T \Ber\left(\z_t\big|\phi\left(\bm{\gamma}_t\right)\right)}_{f_3\left(\Z, \bm{\Gamma}\right)}\underbrace{\N\left(\bm{\Gamma}\big|\bm{\mu}_0, \bm{\Sigma}_0\right)}_{f_4\left(\bm{\Gamma}\right)}.
\end{align}
This posterior density is decomposed into four terms $f_i$ for $i = 1, .., 4$, where the first three terms can be further decomposed. The term $f_1\left(\X\right)$ is decomposed into $T$ terms of the form $f_{1,t}\left(\x_t\right) = p(\y_t\big|\x_t)$, whereas the terms $f_2$ and $f_3$ are further decomposed as follows
\begin{align}
f_{1}\left(\X\right) &= \prod_{t=1}^T \tilde{f}_{1,}\left(\x_t\right) = \prod_{t=1}^T p(\y_t\big|\x_t),\\
f_{2}\left(\X, \Z\right) & = \prod_{t=1}^T\prod_{i=1}^D  f_{2,i,t}\left(x_{i,t}, z_{i,t}\right) = \prod_{t=1}^T\prod_{i=1}^D \left[\left(1-z_{i,t}\right)\circ\delta\left(x_{i,t}\right) + z_{i,t}\circ\N\left(x_{i,t}\big|\rho, \tau\right)\right],\\
f_{3}\left(\Z, \bm{\Gamma}\right) &= \prod_{t=1}^T\prod_{i=1}^D  f_{3,i,t} \left(z_{i,t}, \gamma_{i,t}\right) = \prod_{t=1}^T\prod_{i=1}^D\Ber\left(z_{i,t}\big|\phi\left(\gamma_{i,t}\right)\right).
\end{align}
Each $f_{1,t}$ term only depends on $\x_t$, $f_{2,i,t}$ only depends on $x_{i,t}$ and $z_{i,t}$ and $f_{3,j,t}$ only depends on $z_{i,t}$ and $\gamma_{i,t}$. Furthermore, the terms $f_{2,i,t}$ couple the variables $x_{i,t}$ and $z_{i,t}$, while $f_{3,i,t}$ couple the variables $z_{i,t}$ and $\gamma_{i,t}$. Based on these observations, we choose $\tilde{f_{1,t}}$, $\tilde{f}_{2,i,t}$ and $\tilde{f}_{3,j,t}$ to have the following forms
\begin{align}
\tilde{f}_{1,t}\left(\x_t\right) &= \N\left(\x_t\big|\hat{\m}_{1,t}, \hat{\bm{V}}_{1,t}\right),\\
\tilde{f}_{2,i,t}\left(x_{i,t}, z_{i,t}\right) &= \N\left(x_{i,t}\big|\hat{m}_{2,i,t}, \hat{v}_{2,i,t}\right)\Ber\left(z_{i,t}\big|\phi\left(\hat{\gamma}_{2,i,t}\right)\right),\\
\tilde{f}_{3,i,t}\left(z_{i,t}, \gamma_{i,t}\right) &= \N\left(\gamma_{i,t}\big|\hat{\mu}_{3,j,t}, \hat{\sigma}_{3,i,t}\right)\Ber\left(z_{i,t}\big|\phi\left(\hat{\gamma}_{3,j,t}\right)\right).
\end{align}
The exact term $f_1$ is a distribution wrt. $\y$ conditioned on $\x$, whereas the approximate term $\tilde{f}_1$ is a function of $\x$ that depends on the data $\y$ through $\hat{\m}_1$ and $\hat{\bm{V}}_1$ etc. 
Finally, $f_4$ already belongs to the exponential family and does therefore not have to be approximated by EP. That is, $\tilde{f_4}\left(\bm{\Gamma}\right) = {f_4}\left(\bm{\Gamma}\right) = \N\left(\bm{\Gamma}\big|\mu_{0}, \bm{\Sigma}_0\right)$. 
\\
\\
Define $\hat{\m}_{2,t} = \begin{bmatrix}
\hat{m}_{2,t,1}&\hat{m}_{2,t,2}&\hdots&\hat{m}_{2,t,D}\end{bmatrix}^T$, $\hat{\bm{V}}_{2,t} = \text{diag}\begin{pmatrix}
\hat{v}_{2,t,1}&\hat{v}_{2,t,2}&\hdots&\hat{v}_{2,t,D}\end{pmatrix}^T$ and $\hat{\bm{\gamma}}_{2,t} = \begin{bmatrix}
\hat{\gamma}_{2,t,1}&\hat{\gamma}_{2,t,2}&\hdots&\hat{\gamma}_{2,t,D}\end{bmatrix}$ and similarly for $\hat{\bm{\mu}}_{3,t}$, $\hat{\bm{\Sigma}}_{3,t}$ and $\hat{\bm{\gamma}}_{3,t}$, then the resulting global approximation becomes
\begin{align}
Q\left(\X, \Z, \bm{\Gamma}\right) &\propto \prod_{t=1}^T \underbrace{\N\left(\x_t\big|\hat{\m}_{1,t}, \hat{\bm{V}}_{1,t}\right)}_{\tilde{f}_1, t} \prod_{t=1}^T \underbrace{\N\left(\x_t\big|\hat{\m}_{2,t}, \hat{\bm{V}}_{2,t}\right)\Ber\left(\z_t\big|\phi\left(\hat{\bm{\gamma}}_{2,t}\right)\right)}_{\tilde{f}_{2,t}}\nonumber\\
&\quad \prod_{t=1}^T \underbrace{\N\left(\bm{\gamma}_{t}\big|\hat{\bm{\mu}}_{3,t}, \hat{\bm{\Sigma}}_{3,t}\right)\Ber\left(\z_{t}\big|\phi\left(\hat{\bm{\gamma}}_{3,t}\right)\right)}_{\tilde{f}_{3,t}}\underbrace{\N\left(\bm{\Gamma}\big|\bm{\mu}_0, \bm{\Sigma}_0\right)}_{\tilde{f}_4}\nonumber\\
&\propto \prod_{t=1}^T \N\left(\x_t\big|\hat{\m}_t, \hat{\bm{V}}_t\right) \prod_{t=1}^T \Ber\left(\z_t\big|\phi\left(\hat{\bm{\gamma}_t}\right)\right) \N\left(\bm{\Gamma}\big|\hat{\bm{\mu}}, \hat{\bm{\Sigma}}\right), \label{eq:global_approximation}
\end{align}
where the parameters of the global approximation are obtained by summing the natural parameters. In terms of mean and variance, we get
\begin{align}
\hat{\bm{V}}_t &= \left[\hat{\bm{V}}_{1,t}^{-1} + \hat{\bm{V}}_{2,t}^{-1}\right]^{-1},\label{eq:update_x_v}\\
\hat{\bm{m}}_t &= \hat{\bm{V}}_t\left[\hat{\bm{V}}_{1,t}^{-1}\hat{\m}_{1,t} + \hat{\bm{V}}_{2,t}^{-1}\hat{\m}_{2,t}\right],\label{eq:update_X_m}\\
\hat{\bm{\Sigma}} &= \left[\bm{\Sigma}_0^{-1} + \hat{\bm{\Sigma}}_{3}^{-1}\right]^{-1}\label{eq:update_gam_sigma},\\
\hat{\bm{\mu}} &= \hat{\bm{\Sigma}}\left[\bm{\Sigma}_0^{-1}\bm{\mu}_0 + \hat{\bm{\Sigma}}_{3}^{-1}\hat{\bm{\mu}}_{3}\right],\label{eq:update_gam_mu}\\
\phi\left(\hat{{\gamma}}_{i,t}\right) &= \frac{\phi\left(\hat{{\gamma}}_{2,i,t}\right)\phi\left(\hat{{\gamma}}_{3,i,t}\right)}{\left(1 - \phi\left(\hat{{\gamma}}_{2,i,t}\right)\right)\left(1-\phi\left(\hat{{\gamma}}_{3,j,t}\right)\right) + \phi\left(\hat{{\gamma}}_{2,i,t}\right)\phi\left(\hat{{\gamma}}_{3,i,t}\right)},
\end{align}
where $\hat{\bm{\Sigma}}_{3} \in \mathbb{R}^{TD \times TD}$ is a diagonal matrix, whose the diagonal is obtained by stacking the site variances $\hat{\bm \Sigma}_{3,t}$ for each time point and $\hat{\bm{\mu}}_{3} \in \mathbb{R}^{TD}$ is a vector obtained by stacking the site means $\hat{\bm \mu}_{3,t}$ for each time point. To compute the global approximation, we need to estimate the parameters $\hat{\m}_{1,t}$, $\hat{\bm{V}}_{1,t}$, $\hat{\m}_{2,t}$, $\hat{\bm{V}}_{2,t}$, $\hat{\bm{\mu}}_{3,t}$, $\hat{\bm{\Sigma}}_{3,t}$, $\hat{\bm{\gamma}}_{2,t}$ and $\hat{\bm{\gamma}}_{3,t}$ for all $t \in \left[T\right]$ using EP. The estimation procedure of $\hat{\m}_{1,t}$ and $\hat{\bm{V}}_{1,t}$ depends on the observation model being used, whereas the estimation procedure of the remaining parameters are independent on the choice of observation model. 
\\
\\
In principle, we could choose the approximate posterior distribution of $\bm{\Gamma}$ in eq. \eqref{eq:global_approximation} from a family of distributions that factorizes across space, time or both to reduce the computational complexity. This choice would indeed reduce the computational burden, but in contrast to classical variational inference schemes, the correlation structure of the prior would be ignored in the EP scheme and thus, the resulting posterior approximation would be meaningless for this specific model. 
\\
\\
In the conventional EP algorithm, the site approximations are updated in a sequential manner meaning that the global approximation is updated every time a single site approximation \citep{Minka01} is refined. In this work, we use the parallel update scheme to reduce the computational complexity of the algorithm. That is, we first update all the site approximations of the form $\tilde{f}_{2,i,t}$ for $i \in \left[D\right]$, $t \in \left[T\right]$, and then we update the global approximation w.r.t. $\x_t$ and similarly for the $\tilde{f}_{3,i,t}$ and the global approximation w.r.t. $\bm{\gamma}_t$. From a message passing perspective this can be interpreted as a particular scheduling of messages \citep{Minka05divergencemeasures}.
The proposed algorithm is summarized in Algorithm \ref{fig:algorithm}.

\begin{algorithm}[th]
\scriptsize
\centering
\fbox{\parbox[t][\height][t]{0.9\textwidth}{
\begin{itemize}
\item Initialize approximation terms $\tilde{f}_a$ for $a = 1, 2,3,4$ and $Q$
\item Repeat until stopping criteria
\begin{itemize}
\item For each $\tilde{f}_{1,n,t}$ (\textit{For non-Gaussian likelihoods only}):
\begin{itemize}
\item Compute cavity distribution: $Q^{\backslash 1,n,t} \propto \frac{Q}{\tilde{f}_{1,n,t}}$
\item Minimize: KL$\left(f_{1,n,t}Q^{\backslash 1,n,t}\big|\big|Q^{1, t,\text{new}}\right)$ w.r.t. $Q^{\text{new}}$
\item Compute: $\tilde{f}_{1,n,t} \propto \frac{Q^{1,t,\text{new}}}{Q^{\backslash 1,n,t}}$ to update parameters $\hat{m}_{1,n,t}, \hat{v}_{1,n,t}$ and $\hat{\gamma}_{1,n,t}$.
\end{itemize}

\item For each $\tilde{f}_{2,i,t}$:
\begin{itemize}
\item Compute cavity distribution: $Q^{\backslash 2,i,t} \propto \frac{Q}{\tilde{f}_{2,i,t}}$
\item Minimize: KL$\left(f_{2,i,t}Q^{\backslash 2,i,t}\big|\big|Q^{2, t,\text{new}}\right)$ w.r.t. $Q^{\text{new}}$
\item Compute: $\tilde{f}_{2,i,t} \propto \frac{Q^{2,t,\text{new}}}{Q^{\backslash 2,i,t}}$ to update parameters $\hat{m}_{2,i,t}, \hat{v}_{2,i,t}$ and $\hat{\gamma}_{2,i,t}$.
\end{itemize}
\item Update joint approximation parameters: $\hat{\m}, \hat{\bm{V}}$ and $\hat{\bm{\gamma}}$ 
\item For each $\tilde{f}_{3,i,t}$:
\begin{itemize}
\item Compute cavity distribution: $Q^{\backslash 3,i,t} \propto \frac{Q}{\tilde{f}_{3,i,t}}$
\item Minimize: KL$\left(f_{3,i,t}Q^{\backslash 3,i,t}\big|\big|Q^{3,t,\text{new}}\right)$ w.r.t. $Q^{3,t,\text{new}}$
\item Compute: $\tilde{f}_{3,i,t} \propto \frac{Q^{3,t,\text{new}}}{Q^{\backslash 3,i,t}}$ to update parameters $\hat{\mu}_{3,i,t}, \hat{\sigma}_{3,i,t}$ and $\hat{\gamma}_{3,i,t}$
\end{itemize}
\item Update joint approximation parameters: $\hat{\bm{\mu}}, \hat{\bm{\Sigma}}$ and $\hat{\bm{\gamma}}$
\end{itemize}
\item Compute marginal likelihood approximation
\end{itemize}
} 
}
\caption{Proposed algorithm for approximating the joint posterior distribution over $\X, \Z$ and $\bm{\Gamma}$ conditioned on $\Y$ using parallel EP.}
\label{fig:algorithm}
\end{algorithm}

\subsection{Estimating Parameters for $\tilde{f}_{1,t}$}
The estimation procedure for $\tilde{f}_{1,t}$ depends on the choice of observation model. Here we consider two different observation models, namely the isotropic Gaussian and the probit models. Both of these models lead to closed form update rules, but this is not true for all choices of $p(\y_t|\x_t)$. In general if $p(\y_t|\x_t)$ factorizes over $n$ and each term only depends on $\x_t$ through $\A\x_t$, then the resulting moment integrals are 1-dimensional and can be solved relatively fast using numerical integration procedures \citep{jylanki2011robust} if no closed form solution exists.
\\
\\
Under the Gaussian noise model, we have 
\begin{align}
f_{1,t}\left(\x_t\right) = p(\y_t\big|\x_t) = \N\left(\y_t\big|\A\x_t, \sigma^2 \I\right).
\end{align}
Thus, $f_{1,t}$ is already in the exponential family for all $t \in \left[T\right]$ and does therefore not have to be approximated using EP. In particular, the parameters for $\tilde{f}_{1,t}$ are determined by the relations $\hat{\bm{V}}^{-1}_{1,t} = \frac{1}{\sigma^2}\A^T\A$ and $\hat{\bm{V}}^{-1}_{1,t}\hat{\m}_{1,t} = \frac{1}{\sigma^2}\A^T\y_t$. For simplicity we also assume that the noise variance is constant for all $t$. 
\\
\\
Under the probit likelihood the term $f_{1,t}$ decompose to $f_{1,t} = \prod_{n=1}^N f_{1,t,n}$. In this case, the update of each site approximation $\tilde{f}_{1,t,n}$ resembles the updates for Gaussian process classification using EP, see appendix \ref{app:probit} for details.

\subsection{Estimating Parameters for $\tilde{f}_{2,t}$}
The terms $\tilde{f}_{2,t} = \prod_{i=1}^D \tilde{f}_{2,i,t}\left(x_{i,t}, z_{i,t}\right)$ factor over $i$, which implies that we only need the marginal cavity distributions of each pair of $x_{i,t}$ and $z_{i,t}$. Consider the update of the $j$'th term at time $t$, that is $\tilde{f}_{2,j,t}\left(x_{j,t}, z_{j,t}\right)$. The first step is to compute the marginal cavity distributions by removing the contribution of $\tilde{f}_{2,j,t}\left(x_{j,t}, z_{j,t}\right)$ from the marginal of the global approximation $Q$ using eq. \eqref{eq:cavity}
\begin{align}
Q^{\backslash 2, j, t}\left(x_{j,t}, z_{j,t}\right) &= \frac{Q^{\backslash 2, j, t}\left(x_{j,t}, z_{j,t}\right)}{\tilde{f}_{2,j,t}\left(x_{j,t}, z_{j,t}\right)} \propto \N\left(x_{j,t}\big|\mu^{\backslash 2,j,t}, \Sigma^{\backslash 2,j,t}\right)\Ber\left(z_{j,t}\big|\phi\left( \gamma^{\backslash 2,j,t} \right)\right).
\end{align}
When the approximate distribution belongs to the exponential family, the cavity distribution is simply obtained by computing the differences in natural parameters. Expressed in terms of mean and variance, we get
\begin{align} \label{eq:cav_f2tj_start}
\hat{v}^{\backslash 2, j, t} &= \left[ \hat{V}^{-1}_{t,jj} - \hat{v}^{-1}_{2,j,t}  \right]^{-1},\\
\hat{m}^{\backslash 2, j, t} &= \hat{v}^{\backslash 2, j, t} \left[ \hat{V}^{-1}_{t,jj}\hat{m}_{j,t} - \hat{v}^{-1}_{2,j,t}\hat{m}_{2,j,t}  \right],\\
\hat{\gamma}^{\backslash 2, j, t}  &= \hat{\gamma}_{3,j,t}. \label{eq:cav_f2tj_end}
\end{align}
The cavity parameter for $\gamma_{j,t}$ in $f_{2,j,t}$ is simply equal to $\hat{\gamma}_{3,j,t}$ (and vice versa) since $\hat{\gamma}_{2,j,t}$ and $\hat{\gamma}_{3,j,t}$ are the only two terms contributing to the distribution over $\z_{j,t}$. Next, we form the \textit{tilted} distribution $f_{2,j,t}Q^{\backslash 2, j, t}$ and compute the solution to the KL minimization problem in eq. \eqref{eq:kl_optim} by matching the expected sufficient statistics. This amounts to computing the zeroth, first and second moments w.r.t. $x_{j,t}$
\begin{align} \label{eq:inf_f2_int1}
X_m = \sum_{z_{j,t}} \int x_{j,t}^m \cdot f_{2,j,t}\left(x_{j,t}, z_{j,t}\right) Q^{\backslash 2, j, t}\left(x_{j,t}, z_{j,t}\right) \dd x_{j,t} \quad\text{for}\quad m = 0,1,2,
\end{align}
and the first moment of $z_{j,t}$
\begin{align} \label{eq:inf_f2_int2}
Z_1 = \sum_{z_{j,t}} \int z_{j,t}\cdot f_{2,j,t}\left(x_{j,t}, z_{j,t}\right) Q^{\backslash 2, j, t}\left(x_{j,t}, z_{j,t}\right) \dd x_{j,t}.
\end{align}
For notational convenience we have dropped the dependencies of $X_m$ and $Z_1$ on the indices $t$ and $j$. Alternatively, the moments could be obtained by computing the partial derivatives of the log normalizer of the tilted distribution.
\\
\\
The central moments of $Q^*$ in eq.\eqref{eq:kl_optim} are given by
\begin{align}
E\left[x_{j,t}\right] = \frac{X_1}{X_0}, && V\left[x_{j,t}\right] = \frac{X_2}{X_0} - \frac{X_1^2}{X_0^2}, &&
E\left[z_{j,t}\right] = \frac{Z_1}{X_0}.
\end{align}
Refer to Appendix \ref{app:moments_for_f2tj} for analytical expressions for these moments. Once $Q^*$ has been obtained, we can compute the new update site approximation for $\tilde{f}_{2,j,t}$ using eq. \eqref{eq:ep_site_update} as follows
\begin{align} \label{eq:f2_site_update0}
\tilde{f}^*_{2,j,t}\left(x_{j,t}, z_{j,t}\right) &= \frac{Q^*\left(x_{j,t}, z_{j,t}\right)}{Q^{\backslash 2, j, t}\left(x_{j,t}, z_{j,t}\right)} \propto \N\left(x_{j,t}\big|\hat{m}^*_{2, j,t}, \hat{v}^*_{2, j,t}\right)\Ber\left(z_{j,t}\big|\phi\left(\hat{\gamma}^*_{2,j,t}\right)\right),
\end{align}
where the new site parameters $\hat{m}^*_{2, j,t}$ and $\hat{v}^*_{2, j,t}$, are obtained by computing differences in natural parameters in the same manner as for the cavity parameters in eq. \eqref{eq:cav_f2tj_start} - \eqref{eq:cav_f2tj_end}
\begin{align} \label{eq:f2_site_update1}
\hat{v}^*_{2, j,t} &= \left[V\left[x_{j,t}\right]^{-1} - \left(\hat{v}^{\backslash 2, j, t}\right)^{-1}\right]^{-1},\\
\hat{m}^*_{2, j,t} &= \hat{v}^*_{2, j,t}\left[V\left[x_{j,t}\right]^{-1}E\left[x_{j,t}\right] - \left(\hat{v}^{\backslash 2, j, t}\right)^{-1}\hat{m}^{\backslash 2, j, t}\right].\label{eq:f2_site_update2}
\end{align}
The new site parameters for $z_{j,t}$ are obtained as (see Appendix \ref{app:moments_for_f2tj} for details)
\begin{align} \label{eq:f2_site_update3}
\phi\left(\hat{\gamma}^*_{2,j,t}\right) &\stackrel{(a)}{=} \frac{\frac{E\left[z_{j,t}\right]}{\phi\left(\hat{\gamma}^{\backslash 2, j, t} \right)}}{\frac{1-E\left[z_{j,t}\right]}{1-\phi\left(\hat{\gamma}^{\backslash 2, j, t} \right)} + \frac{E\left[z_{j,t}\right]}{\phi\left(\hat{\gamma}^{\backslash 2, j, t} \right)}} \stackrel{(b)}{=} \frac{\N\left(0\big|\hat{m}^{\backslash 2,i} - \rho_0, \hat{V}^{\backslash 2, j,t}+\tau_0\right)}{\N\left(0\big|\hat{m}^{\backslash 2,i}, \hat{V}^{\backslash 2, i}\right) + \N\left(0\big|\hat{m}^{\backslash 2,i} - \rho_0, \hat{V}^{\backslash 2, j,t}+\tau_0\right)},
\end{align}
where $(a)$ follows from forming the quotient of the two Bernoulli distributions and $(b)$ follows from straightforward algebraic reduction after substituting in the expression for the expectation of $z_{j,t}$.

\subsection{Estimating Parameters for $\tilde{f}_{3,t}$}
The procedure for updating $\tilde{f}_{3,t} = \prod_{i=1}^D \tilde{f}_{3,j,t}$ is completely analogously to the procedure for $\tilde{f}_{2,t}$. Consider the update for the $j$'th term at time $t$, that is $\tilde{f}_{3,j,t}$. After computing the cavity distribution in the same manner as in eq. \eqref{eq:cav_f2tj_start}-\eqref{eq:cav_f2tj_end}, we now compute the moments w.r.t. $\gamma_{j,t}$ and $z_{j,t}$ of the (unnormalized) tilted distribution
\begin{align} \label{eq:inf_f3_int1}
G_m &= \sum_{z_{j,t}} \int \gamma^m_{j,t}\cdot f_{3,j,t}\left(z_{j,t}, \gamma_{j,t}\right) Q^{\backslash 3,j,t}\left(z_{j,t}, \gamma_{j,t}\right) \dd \gamma_{j,t} \quad\text{for}\quad m = 0,1,2,\\
Z_1 &= \sum_{z_{j,t}} \int z_{j,t}\cdot f_{3,j,t}\left(z_{j,t}, \gamma_{j,t}\right) Q^{\backslash 3,j,t}\left(z_{j,t}, \gamma_{j,t}\right) \dd \gamma_{j,t}.\label{eq:inf_f3_int2}
\end{align}
Given these moments, we can obtain the central moments for $Q^*$ in eq. \eqref{eq:kl_optim}
\begin{align}
E\left[\gamma_{j,t}\right] = \frac{G_1}{G_0}, && V\left[\gamma_{j,t}\right] = \frac{G_2}{G_0} - \frac{G_1^2}{G_0^2}, &&
E\left[z_{j,t}\right] = \frac{Z_1}{G_0}.
\end{align}
Refer to Appendix \ref{app:moments_for_f3tj} for analytical expression of the moments. These moments completely determine $Q^*$ and the $j$'th site update at the $t$ is computed analogous to $\tilde{f}_{2,j,t}$ in eq. \eqref{eq:f2_site_update0} using eq. \eqref{eq:f2_site_update1}, \eqref{eq:f2_site_update2} and \eqref{eq:f2_site_update3}.

\subsection{The Computational Details}
In the previous sections, we have described how to use EP for approximate inference for the proposed model, and in this section, we discuss some of the computational details of the resulting EP algorithm. 

\subsubsection{Updating the Global Covariance Matrices} \label{subsec:update_cov}
Given a set of updated site approximations, $\tilde{f}_{2,t} = \prod_j \tilde{f}_{2,j,t}$, we can compute the parameters for the global approximate distribution of $\x_t$ using eq. \eqref{eq:update_x_v} and \eqref{eq:update_X_m}. Direct evaluation of eq.~\eqref{eq:update_x_v} results in a computational complexity of $\mathcal{O}\left(D^3\right)$. Recall, that $N$ is assumed to be smaller than $D$. This implies that $\hat{\bm{V}}_{1,t}^{-1} = \frac{1}{\sigma^2_0}\A^T\A$ has low rank. Furthermore, the matrix $\hat{\bm{V}_{2,t}}$ is diagonal, and therefore we can apply the matrix inversion lemma as follows
\begin{align}
\hat{\bm{V}}_{t} &= \hat{\bm{V}}_{2,t} - \hat{\bm{V}}_{2,t}\A^T \left(\sigma^2_0\I + \A\hat{\bm{V}}_{2,t}\A^T\right)^{-1} \A \hat{\bm{V}}_{2,t}. \label{eq:comp_details_matrix_inv_lemma}
\end{align}
The inverse of $\sigma^2_0\I + \A\hat{\bm{V}}_{2,t}\A^T = \bm{L}_t\bm{L}_t^T$ can be computed in $\mathcal{O}\left(N^3\right)$ using a Cholesky decomposition. Thus, for $N < D$ eq.~\eqref{eq:comp_details_matrix_inv_lemma} scales as $\mathcal{O}\left(ND^2\right)$. Moreover, eq.~\eqref{eq:cav_f2tj_start} shows that we only require the diagonal elements of $\tilde{\bm{V}}_{t}$ in order to update the site approximation parameters for $\tilde{f}_{2,t}$. Hence, we can further reduce the computational complexity by only computing the diagonal of $\hat{\bm{V}}_{t}$ as follows
\begin{align}
\text{diag}\left[\hat{\bm{V}}_{t}\right] &= \text{diag}\left[\hat{\bm{V}}_{2,t}\right] - \text{diag}\left[\hat{\bm{V}}_{2,t}\A^T \bm{L}^{-T}_t\bm{L}^{-1}_t \A \hat{\bm{V}}_{2,t}\right]\nonumber\\
  &= \text{diag}\left[\hat{\bm{V}}_{2,t}\right] - \text{diag}\left[\hat{\bm{V}}_{2,t}^2\right]  \circ\left(\bm{1}^T \left(\R_t \circ \R_t\right)\right), \label{eq:comp_details_v_diagonal}
\end{align}
where $\R_t \in \mathbb{R}^{N \times D}$ is defined as $\R_t = \bm{L}^{-1}_t\A$ and $\bm{1}$ is a column vector of ones. The resulting computational cost is $\mathcal{O}\left(N^2D\right)$. Similarly, the mean of the global approximate distribution of $\x_t$, can be efficiently evaluated as
\begin{align}
\hat{\bm{m}}_t &= \hat{\bm{V}}_{2,t}\bm{\eta}_t - \hat{\bm{V}}_{2,t} \R^T_t\R_t \hat{\bm{V}}_{2,t}\bm{\eta}_t, 
\end{align}
where $\bm{\eta}_t = \hat{\bm{V}}_{1,t}^{-1}\hat{\m}_{1,t} + \hat{\bm{V}}_{2,t}^{-1}\hat{\m}_{2,t}$. The total cost of updating the posterior distribution for $\x_t$ for all $t \in \left[T\right]$ is therefore $\mathcal{O}\left(TN^2D\right)$. 
\\
\\
Unfortunately, we cannot get the same speed up for the refinement of the global approximation of $\bm{\Gamma}$ since the prior covariance matrix $\bm{\Sigma}_0$ in general is full rank. However, we still only require the diagonal elements of the approximate covariance matrix $\hat{\bm{\Sigma}}$.  We implement the update as advocated by \citet{rasmussen2006a}, that is,
\begin{align}
\hat{\bm{\Sigma}} &= \left[\bm{\Sigma}_0^{-1} + \hat{\bm{\Sigma}}_{3}^{-1}\right]^{-1}\nonumber\\
&= \bm{\Sigma}_0 - \bm{\Sigma}_0\hat{\bm{\Sigma}}^{-\frac{1}{2}}_3\left(\hat{\bm{\Sigma}}^{-\frac{1}{2}}_3\bm{\Sigma}_0\hat{\bm{\Sigma}}^{-\frac{1}{2}}_3 + \I\right)^{-1}\hat{\bm{\Sigma}}^{-\frac{1}{2}}_3\bm{\Sigma}_0,
\end{align}
where the second equality follows from the matrix inverse lemma. Again, we compute the required inverse matrix using the Cholesky decomposition,  so that the total cost is $\mathcal{O}\left(D^3T^3\right)$.

\subsubsection{Initialization, Convergence and Negative Variances}
We initialize all the site terms to be rather uninformative, that is $\hat{m}_{2,i,t} = 0$, $\hat{v}_{2,i,t} = 10^4$, $\hat{\gamma}_{2,i,t} = 0$, $\hat{\mu}_{3,i,t} = 0$, $\hat{\sigma}_{3,i,t} = 10^4$, $\hat{\gamma}_{3,i,t} = 0$ for all $i \in \left[D\right]$ and $t \in \left[T\right]$ assuming standard scaling of the forward model $\A$.
\\
\\
There are in general no convergence guarantees for EP and the parallel version in particular can suffer from convergence problems \citep{Seeger2005-xa}. The standard procedure to overcome this problem is to use ``damping'' when updating the site parameters
\begin{align} \label{eq:damping}
\tilde{f}^* = \tilde{f}^{1-\alpha}_{\text{old}} \tilde{f}^{\alpha}_{\text{new}},
\end{align}
where $\alpha \in \left[0, 1\right]$ is the damping parameter and $\tilde{f}_{\text{old}}$ is the site approximation at the previous iteration. Since both $\tilde{f}_{\text{old}}$ and $\tilde{f}_{\text{new}}$ belongs to the exponential family, the update in eq. \eqref{eq:damping} corresponds to taking a convex combination of the previous and the new natural parameters of the site approximation. 
\\
\\
Negative variances occur ``naturally'' in EP \citep{bishop2006a} when updating the site approximations. However, this can lead to instabilities of the algorithm, non-positive semi-definitiveness of the posterior covariance matrices and convergence problems. We therefore take measures to prevent negative site variances. One way to circumvent this is to change a negative variance to $+\infty$, which corresponds to minimizing the KL divergence in eq. \eqref{eq:kl_optim} with the site variance constrained to be positive \citep{JMLR:v14:hernandez-lobato13a}. In practice, when encountering a negative variance after updating a given site we use $v_{\infty} = 10^2$ and $\sigma_{\infty} = 10^6$ for $\tilde{f}_{2,i,t}$ and $\tilde{f}_{3,i,t}$, respectively.

\section{Further Approximations}\label{sec:approximations}
As mentioned earlier, the updates of the global parameters for $\x_t$ and $\bm{\Gamma}$ are the dominating operations scaling as $\mathcal{O}\left(TN^2D\right)$ and $\mathcal{O}\left(D^3T^3\right)$, respectively. The latter term becomes prohibitive for moderate sizes of $D$ and $T$ and calls for further approximations. In this section, we introduce three simple approximations to reduce the computational complexity of the refinement of the posterior distribution for $\bm{\Gamma}$. The approximations and their computational complexities are summarized in table \ref{table:summary_approximations}.

\begin{table}[t]
\setlength{\tabcolsep}{12pt}
\centering
\begin{tabular}{lccc}
\hline
\bf{Approximation} & & \bf{Complexity} &  \bf{Storage}\\
\hline
Full EP & (EP) & $\mathcal{O}\left({T^3D^3}\right)$ & $\mathcal{O}\left({T^2D^2}\right)$\\
Low rank &(LR) & $\mathcal{O}\left({K^2TD}\right)$ & $\mathcal{O}\left({KTD}\right)$\\
Common precision &(CP) & $\mathcal{O}\left({TD^2 +DT^2}\right)$ & $\mathcal{O}\left({D^2 + T^2}\right)$\\
Group &(G) & $\mathcal{O}\left({T_g^3 D_g^3}\right)$ & $\mathcal{O}\left({T_g^2D_g^2}\right)$\\
\hline
\end{tabular}
\caption{Summary of approximation schemes for updating the global parameters for $\bm{\Gamma}$.}
\label{table:summary_approximations}
\end{table}

\subsection{The Low Rank Approximation}
The eigenvalue spectrum of many prior covariance structures of interest, for example simple neighborhoods, decay relatively fast. Therefore, we can approximate $\bm{\Sigma}_0$ with a low rank approximation plus a diagonal matrix $\bm{\Sigma}_0 \approx \bm{U}\Ss\bm{U}^T + \bm{\Lambda}$, where $\Ss \in \mathbb{R}^{K\times K}$ is a diagonal matrix containing $K$ largest eigenvalues, and $\bm{U}\in\mathbb{R}^{DT \times K}$ is a matrix containing the corresponding eigenvectors \citep{riihimaki2014laplace}. The diagonal matrix $\bm{\Lambda}$ is chosen such that the diagonal in the exact prior covariance matrix $\bm{\Sigma}_0$ is preserved. This allows us to apply the matrix inversion lemma to compute the update of the posterior covariance matrix for $\bm{\Gamma}$ (see Section \ref{subsec:update_cov}).
\\
\\
Computing the eigendecomposition of $\bm{\Sigma}_0 \in \mathbb{R}^{DT\times DT}$ scales in general as $\mathcal{O}\left(D^3T^3\right)$. However, when the prior covariance has Kronecker structure, the eigendecompositions of $\bm{\Sigma}_0 = \bm{\Sigma}_t \otimes \bm{\Sigma}_s$ can be efficiently obtained from the eigendecompositions of $\bm{\Sigma}_t \in \mathbb{R}^{T \times T}$ and $\bm{\Sigma}_s  \in \mathbb{R}^{D \times D}$. In this case, the eigendecomposition of $\bm{\Sigma}_0$ can be obtained in $\mathcal{O}\left(D^3 + T^3\right)$.
\\
\\
Using a $K$-rank approximation, the computational cost of refining the covariance matrix for $\bm{\Gamma}$ becomes $\mathcal{O}\left(K^2DT\right)$ and the memory footprint is $\mathcal{O}\left(TDK\right)$. For a fixed value of $K$ this scales linearly in both $D$ and $T$. However, to maintain a sufficiently good approximation $K$ can scale with both $D$ and $T$. 

\subsection{The Common Precision Approximation}
Rather than approximating the prior covariance matrix as done in the low rank approximation, we now approximate the EP approximation scheme itself. If the prior covariance matrix for $\bm{\Gamma}$ can be written in terms of Kronecker products, we can significantly speed up the computation of the posterior covariance matrix of $\bm{\Gamma}$ by approximating the site precisions with a single common parameter. Let $\tilde{\bm{\theta}}_{3} \in \mathbb{R}^{DT \times 1}$ be a vector containing the site precisions (inverse variances) for the site approximations $\left\lbrace f_{3, i,t} \right\rbrace$ for all $i \in \left[D\right]$ and for all $t \in \left[T\right]$, then we make the following approximation 
\begin{align}
	\tilde{\bm{\Sigma}}_{3} \approx \bar{\theta}^{-1}\I,
\end{align}
where $\bar{\theta}$ is the mean of value of $\tilde{\bm{\theta}}_{3}$. Assume the prior covariance matrix for $\bm{\Gamma}$ can be decomposed into a temporal part and a spatial part as follows $\bm{\Sigma}_0 = \bm{\Sigma}_t \otimes \bm{\Sigma}_s$. Let $\U_t$, $\U_s$ and $\Ss_t$, $\Ss_s$ be eigenvectors and eigenvalues for $\bm{\Sigma}_t \in \mathbb{R}^{T \times T}$ and $\bm{\Sigma}_s \in \mathbb{R}^{D \times D}$, respectively. The global covariance matrix is updated as $\tilde{\bm{\Sigma}} = \bm{\Sigma}_0\left(\bm{\Sigma}_0 + \tilde{\bm{\Sigma}}_3\right)^{-1}\tilde{\bm{\Sigma}}_3$. We now use the properties of eigendecompositions for Kronecker products to compute the inverse matrix
\begin{align}
\left(\bm{\Sigma}_t \otimes \bm{\Sigma}_s + \tilde{\bm{\Sigma}}_3\right)^{-1} &\approx \left(\bm{\Sigma}_t \otimes \bm{\Sigma}_s + \bar{\Sigma}_3 \I\right)^{-1}\nonumber\\
&= \left[\left(\U_t \otimes \U_s\right)\left(\Ss_t \otimes \Ss_s\right)\left(\U^T_t \otimes \U^T_s\right) + \bar{\Sigma}_3 \I\right]^{-1}\nonumber\\
&= \left(\U_t \otimes \U_s\right)\left(\Ss_t \otimes \Ss_s + \bar{\Sigma}_3 \I\right)^{-1}\left(\U^T_t \otimes \U^T_s\right) \label{eq:approx_common_precision},
\end{align}
where $\left(\Ss_t \otimes \Ss_s + \bar{\Sigma}_3 \I\right)$ is diagonal and therefore fast to invert. The \textit{common precision} approximation $\hat{\bm{\Sigma}}_{CP}$ is then obtained as
\begin{align}
\hat{\bm{\Sigma}}_{CP} &= \left(\bm{\Sigma}_t \otimes \bm{\Sigma}_s\right)\left(\bm{\Sigma}_t \otimes \bm{\Sigma}_s + \bar{\Sigma}_3 \I\right)^{-1}\bar{\Sigma}_3 \I\nonumber\\
&= \left(\U_t \otimes \U_s\right)\left(\Ss_t \otimes \Ss_s\right)\left(\Ss_t \otimes \Ss_s + \bar{\Sigma}_3 \I\right)^{-1}\left(\U^T_t \otimes \U^T_s\right)\bar{\Sigma}_3.
\end{align}
Let $\M \in \mathbb{R}^{TD \times 1}$ denote the diagonal of $\left(\Ss_t \otimes \Ss_s\right)\left(\Ss_t \otimes \Ss_s + \bar{\Sigma}_3 \I\right)^{-1}$, then we can compute the diagonal of $\hat{\bm{\Sigma}}_{CP}$ as follows
\begin{align}
\text{diag}\left[\hat{\bm{\Sigma}}_{CP}\right]_i &=  \bar{\Sigma}_3\sum_{k} \left(\U_t \otimes \U_s\right)_{ik}M_{k}\left(\U^T_t \otimes \U^T_s\right)_{ki}\nonumber\\
&= \bar{\Sigma}_3\sum_{k} \left(\U_t \otimes \U_s\right)^2_{ik}M_{k}\nonumber\\
\Rightarrow\quad\text{diag}\left[\hat{\bm{\Sigma}}_{CP}\right] &=  \bar{\Sigma}_3\left(\U_t \circ \U_t \otimes \U_s \circ \U_s\right) \M,
\end{align}
where $\circ$ is the Hadamard-product. We now see that the desired diagonal can be obtained by multiplying a Kronecker product with a vector and this can be computed efficiently using the identity
\begin{align}
\text{vec}\left[\A\B\C\right] = \left(\C^T \otimes \A\right)\text{vec}\left[\B\right]. \label{eq:kron_vec_mult}
\end{align}
Therefore, 
\begin{align}
\text{diag}\left[\hat{\Sigma}_{CP}\right] &= \bar{\Sigma}_3\cdot \text{vec}\left[\left(\U_s \circ \U_s\right)\text{vec}^{-1}\left[\M\right]\left(\U_t \circ \U_t\right)^T\right].
\end{align}
Since the Hadamard products can be precomputed, this scales as $\mathcal{
O}\left(D^2T + T^2D\right)$. During the EP iterations we only need to store $\U_s \in \mathbb{R}^{D \times D}$ and $\U_t \in \mathbb{R}^{T \times T}$, so the resulting memory footprint is $\mathcal{O}\left(D^2 + T^2\right)$. The posterior mean vector can also be computed efficiently by iteratively applying the result from eq. \eqref{eq:kron_vec_mult}
\begin{align}
\hat{\bm{\Sigma}}_{CP}\bm{\eta} &= \left(\U_t \otimes \U_s\right)\text{diag}\left[M\right]\left(\U^T_t \otimes \U^T_s\right)\bm{\eta},
\end{align}
where $\bm{\eta} = \hat{\bm{\Sigma}}_{3}^{-1}\hat{\bm{\mu}}_3 + \hat{\bm{\Sigma}}_0^{-1}\hat{\bm{\mu}}_{0}$. 
\\
\\
The proposed approximation reduces the cost from $\mathcal{O}\left(D^3T^3\right)$ to $\mathcal{O}\left(D^2T + T^2D\right)$. If the spatial covariance matrix is a Kronecker product itself, for example, $\bm{\Sigma}_s = \bm{\Sigma}_x \otimes \bm{\Sigma_y}$ or $\bm{\Sigma}_s = \bm{\Sigma}_x \otimes \bm{\Sigma}_y \otimes \bm{\Sigma}_z$, the computational complexity can be further reduced. Such covariance structures could occur in image application or in analysis of fMRI data.
\\
\\
This common precision approximation is closely related to the recently proposed \textit{Stochastic Expectation Propagation} (SEP) \citep{Li2015-ra}, where both the means and variances of the site approximation terms have been tied together. Tying both means and variances is reasonable when the site terms are approximating likelihood terms and $N \gg D$. In case of the present model, we expect positive values of $\Gamma_{i,t}$ for $z_{i,t} = 1$ and negative values of $\Gamma_{i,t}$ for $z_{i,t} = 0$, and thus enforcing a common mean for the site approximation terms $\tilde{f}_{3, i, t}$ would not make sense. 
\\
\\
From experiments we have observed that this common precision approach significantly increases the number of iterations until convergence. However, this problem can be mitigated by repeating the updates for the site approximations $\tilde{f}_{3, i, t}$ and the global approximation for $\bm{\Gamma}$ a few times before moving on to update the site approximations for $\tilde{f}_{2, i, t}$. Specifically, within each EP iteration we repeat the updates for posterior distribution of $\bm{\Gamma}$ 5 times. The added computational workload is still negligible compared to full EP. Furthermore, for some problem instances CP-EP can oscillate. The oscillation can be alleviated heuristically by decreasing the damping parameter $\alpha$ by $10\%$ if the approximate log likelihood decreases from one iteration to another after the first $100$ iterations.

\subsection{Grouping the Latent Structure Variables}
Consider a problem, where the spatial coordinates $\bm{d}_{i}$ for each  $x_{i,\cdot}$ form a uniformly spaced grid. Assume the characteristic length-scale of the sparsity pattern is large relative to the grid size, then support variables $\left\lbrace z_i \right\rbrace$ in a neighborhood could  ``share'' the same $\gamma$-variable with a little loss of accuracy \citep{Jacob2009,JMLR:v14:hernandez-lobato13a}. This \textit{grouping} of the latent variables could either be in the spatial, temporal or both dimensions. Let $G$ be the number of groups and $g: \left[D\right] \times \left[T\right] \rightarrow \left[G\right]$ be a grouping function that maps from a spatial and temporal index to a group index, then the grouped version of the prior is given by
\begin{align}\label{eq:group_prior_start}
p(\Z\big|\bm{\gamma})&=\prod_{t=1}^T\prod_{i=1}^D \Ber\left(z_{t,i}\big|\phi\left(\gamma_{g\left(i, t\right)}\right)\right),\\
p(\bm{\gamma}) &= \N\left(\bm{\gamma}\big|\bm{\mu}_0, \bm{\Sigma}_0\right),\label{eq:group_prior_end}
\end{align}
where $\bm{\mu}_0 \in \mathbb{R}^{G}$ and $\bm{\Sigma}_0 \in \mathbb{R}^{G \times G}$ are the prior mean and covariance for the new grouped model. The resulting computational complexity is indeed determined by the size of the groups. For example, assume that the support variable for a given problem have been grouped in groups of 2 in both the spatial dimension and temporal dimension, then the total number of groups becomes $G = \frac{1}{2}D\frac{1}{2}T = \frac{1}{4}DT$ and the resulting computational cost is reduced to a fraction of $\left(\frac{1}{4}\right)^3$ of the cost of the full EP scheme. Furthermore, if necessary both the low rank and the common precision approximation can be applied on top of this approximation. 
\FloatBarrier

\section{The Marginal Likelihood Approximation and Model Selection}
\label{sec:Hyperparameter}

The model contains several hyperparameters $\bm{\Omega} \in \mathbb{R}^L$, which include, for example, the hyperparameters of the kernel for $\bm{\Gamma}$.
In a fully Bayesian setting, the natural approach to handle hyperparameters is to impose prior distributions and marginalize over the hyperparameters. The exact, but generally intractable marginalization integral is given by
\begin{align}
	p(\X, \Z, \bm{\Gamma}|\Y) = \int p(\X, \Z, \bm{\Gamma}|\Y, \bm{\Omega}) p\left(\bm{\Omega}\big|\Y\right) \text{d} \bm{\Omega}, \label{eq:full_marginalization}
\end{align}
where $p\left(\bm{\Omega}\big|\Y\right) \propto p(\Y\big|\bm{\Omega})p(\bm{\Omega})$ for some prior distribution $p(\bm{\Omega})$. The true marginal likelihood $p(\Y\big|\bm{\Omega})$ is given by the following marginalization
\begin{align}
p(\Y\big|\bm{\Omega}) &= \int\! f_1\left(\X\big|\bm{\Omega}\right)\sum_{\Z}f_2\left(\X,\Z\big|\bm{\Omega}\right)\dd \X \!\int\! f_3\left(\Z,\bm{\Gamma}\big|\bm{\Omega} \right)f_4\left(\bm{\Gamma}\big|\bm{\Omega}\right) \dd \bm{\Gamma}. \label{eq:exact_likelihood}
\end{align}
The exact quantity is intractable, but the EP framework provides an approximation to the marginal likelihood conditioned on the hyperparameters, $p(\Y\big|\bm{\Omega}) \approx Q(\Y|\bm{\Omega})$. The approximation $Q(\Y|\bm{\Omega})$ is obtained by substituting the exact site terms, for example, $f_{2,i,t}$, with a scaled version of the corresponding site approximation, for example, $s_{2,i,t}\tilde{f}_{2,i,t}$, and then carrying out the marginalization analytically. The scaling constants, for example, $s_{2, i, t}$, are chosen such that
\begin{align}
	\mathbb{E}_{Q^{\backslash 2, i, t}}\left[f_{2,i,t}\left(x_{i,t}, z_{i,t}\right)\right] = s_{2,i,t}\mathbb{E}_{Q^{\backslash 2, i, t}}\left[\tilde{f}_{2,i,t}\left(x_{i,t}, z_{i,t}\right)\right]
\end{align}
and similarly for all the site terms $f_{a, i, t}$ for $a \in \left[4\right]$, $i \in \left[D\right]$, $t \in \left[T\right]$. In the following, we will describe three different approximation strategies based on the marginal likelihood approximation.

\subsection{Maximum Likelihood and MAP Estimation}
This simplest and most crude approximation is to use a point estimate of $\bm{\Omega}$ instead of integrating over the uncertainty. Specifically, we aim to locate the maximum a posteriori (MAP) value by maximizing $\ln Q\left(\bm{\Omega}|\Y\right) = \ln Q\left(\Y\big|\bm{\Omega}\right) + \ln p\left(\bm{\Omega}\right) + \text{constant}$ using gradient-based methods. A maximum likelihood type II estimate is obtained by choosing an (improper) flat prior $p\left(\bm{\Omega}\right) \propto 1$. For severely ill-posed problems, the marginal likelihood approximation can be completely non-informative with regard to one or more hyperparameters and thus, the maximum likelihood estimate can lead to suboptimal and unstable results for some problems. For some problem instances, it can also happen that the marginal likelihood solution with regard to the prior mean and variance of $\bm{\Gamma}$ is not in the interior of $\mathbb{R}^2$ and thus, gradient-based optimization with regard to these parameters will diverge. However, this problem is easily solved by imposing a weakly informative prior on the prior variance of $\bm{\Gamma}$ with little influence on the result (see Appendix \ref{appendix:prior} for more details).
\\
\\
The marginal likelihood approximation, $Q\left(\Y\big|\bm{\Omega}\right)$, depends on the hyperparameters $\bm{\Omega}$ directly as well as through the site parameters, but the latter dependency can be ignored in gradients computations when the EP fixed point conditions hold \citep{Seeger2005-xa}. The hyperparameter optimization procedure proceeds in an iterative two-stage fashion, where we first run EP until convergence and then we take a gradient step with regard to the hyperparameters and then repeat.


\subsection{Approximate Marginalization using Numerical Integration}
As a better approximation of eq.\,\eqref{eq:full_marginalization}, we propose to approximate the marginalization integral using numerical integration with a finite sum using a central composite design (CCD) grid \citep{Rue_undated-nu}. This method has previously been successfully applied for marginalizing over hyperparameters in Gaussian process based models and the accuracy is reported to be between empirical Bayes and full marginalization using a dense grid \citep{Vanhatalo2010-gw}. We approximate the marginal posterior distribution as follows
\begin{align}
	p(\X, \Z, \bm{\Gamma}|\Y) &\approx \int Q(\X, \Z, \bm{\Gamma}|\Y, \bm{\Omega}) Q\left(\bm{\Omega}\big|\Y\right) \text{d} \bm{\Omega}\\
	& \approx \sum_{m=1}^M Q(\X, \Z, \bm{\Gamma}|\Y, \bm{\Omega}_m) Q\left(\bm{\Omega}_m\big|\Y\right) w_m,
\end{align}
for a set of points $\left\lbrace \bm{\Omega}_m \right\rbrace_{m=1}^M$, a set of integration weights $\left\lbrace w_m \right\rbrace_{m=1}^M$. Thus, the resulting approximate marginal posterior distribution becomes a Gaussian mixture model with mixing weights $\pi_m = Q\left(\bm{\Omega}_m\big|\Y\right) w_m$ and components $Q(\X, \Z, \bm{\Gamma}|\Y,\bm{\Omega}_m)$.
\\
\\
To keep the computational burden to a minimum, we use a so-called Central Composite Design (CCD) to choose the points and weights.
Most of the hyperparameters are variance or scale parameters and hence constrained to be positive. Therefore, we first transform these parameters into an unconstrained space using a log transformation, $\lambda_i = \ln \Omega_i$. Next, we locate the mode in the transformed parameter space, $\hat{\bm{\lambda}}_{\text{MAP}}$, by optimizing $Q\left(\Y\big|\bm{\lambda}\right)$ with regard to $\bm{\lambda}$ using gradient-based optimization methods and numerically estimate the inverse Hessian, $\hat{\bm{S}} = \bm{H}^{-1}$, at the mode $\hat{\bm{\lambda}}_{\text{MAP}}$.
\\
\\
The CCD integration points are then obtained as $\bm{\lambda}_m = \hat{\bm{S}}^{\frac{1}{2}}\p_m + \hat{\bm{\lambda}_{\text{MAP}}}$, where $\left\lbrace \p_m \right\rbrace_{m=1}^M$ is a CCD design grid \citep{Rue_undated-nu} in $L$-dimensions. The points on the CCD grid consist of a fractional factorial design as well as $2K$ star points and a center point. All points, except for the center point, lives on the surface of a $L$-dimensional ball with radius $\sqrt{L}$. This specific design choice requires a much smaller number of points compared to a dense grid. For example, for $L = 2, 3, 4, 5$ parameters, the number of CCD points are $M = 9, 15, 25, 43$, respectively. The integration weights $\left\lbrace w_m \right\rbrace_{m=1}^M$ are chosen such that the integral match for the first three moments of a $L$-dimensional standardized Gaussian random variable, $\z \sim \mathcal{N}(\mathbf{0}, \I_L)$, and $\mathbb{E}\left[1\right] = 1$, $\mathbb{E}\left[\z\right] = 0$, $\mathbb{E}\left[\z^T \z\right] = L$.

\subsection{Bayesian Optimization}
There can be some challenges with gradient-based optimization of the marginal likelihood approximation. Firstly, the optimization problem is in general non-convex and thus, the results can suffer from poor local optima. Secondly, for some problem instances the marginal likelihood approximate can exhibit discontinuities (as discussed in experiment \ref{sec:experiment_1}).
\\
\\
To counter these issues, we consider Bayesian optimization \citep{Shahriari2016-bo,Snoek2012-oo} as a third strategy to model selection as it does not depend on gradient information. As indicated by the name, Bayesian optimization is a probabilistic approach to optimization, where the objective function is modelled as a random function. Thus, the approach allows us to model the potential discontinuities. Specifically, we use a Gaussian process to model log posterior density as follows
\begin{align}
	\ln Q (\bm{\Omega}|Y ) \sim \mathcal{GP}\left(\mu\left(\bm{\Omega}\right), k\left(\bm{\Omega}, \bm{\Omega}'\right)\right),
\end{align}
where $\mu: \mathbb{R}^K \rightarrow \mathbb{R}$ is a mean function and $k: \mathbb{R}^L \times \mathbb{R}^L \rightarrow \mathbb{R}$ is the kernel function. Rather than following the direction of the gradient, Bayesian optimization works by exploring values of $\bm{\Omega}$, that are likely to improve the value of the objective function as measured by a so-called acquisition function. For more details on Bayesian optimization, we refer to \citep{Shahriari2016-bo,Snoek2012-oo,Brochu2010-as} .

\FloatBarrier

\section{Numerical Experiments}\label{sec:experiments}
In this section, we conduct a series of experiments designed to investigate the properties of the proposed model and the associated EP inference scheme. 
\\
\\
We describe seven experiments with a Gaussian observation model and one experiment with a probit observation model. In the first five experiments, we focus on problem instances with a single measurement vector. Experiment 1 investigates the effect of the prior by analyzing a synthetic data set with a range of different values for the hyperparameters. In the second experiment, we compare the three different approximation schemes (low rank, common precision, group) to standard EP. Specifically, we analyze a synthetic data set with all four methods and compare the results. Experiment 3 is designed to investigate how the EP algorithms perform as a function of the undersampling ratio $N/D$ giving rise to the so-called \textit{phase transition curves} \citep{journals/pieee/DonohoT10}. In experiment 4, we apply the proposed model to a compressed sensing problem and in experiment 5, we apply our model to a binary classification task, where the goal is to discriminate between utterances of two different vowels using log-periodograms as features.
\\
\\
In Experiment 6-8, we turn our attention to problems with multiple measurement vectors. In the sixth experiment, we qualitatively study the properties of the proposed methods in the multiple measurement vector setting. We demonstrate the benefits of modeling both the spatial and temporal structure of the support and discuss the marginal likelihood approximation for hyperparameter tuning. Experiment 7 studies the performance of the EP algorithms as a function of the undersampling ratio when multiple measurement vectors are available and compare the results to competing methods. Finally, in Experiment 8 we apply the proposed method to an EEG source localization problem \citep{baillet2001a}. 
\\
\\
For the subset of experiments, where the ground truth solutions are available, we use the \textit{Normalized Mean Square Error (NMSE)} and the \textit{F-measure} \citep{Rijsbergen:1979:IR:539927} to quantify the performance of the algorithms. In particular, we compute the NMSE between the posterior mean $\hat{\X} = \E_{Q(\X|\Y)}\left[\X\right]$ and the true solution $\X_0$ to quantify the algorithms abilities to reconstruct the true signal $\X_0$ 
\begin{align}
	\text{NMSE} = \frac{\|\hat{\X} - \X_0\|^2_{\text{F}}}{\|\X_0\|^2_{\text{F}}},
\end{align}
where $\|\cdot\|_{\text{F}}$ is the Frobenius norm. We use the F-measure to quantify the algorithms' abilities to recover the true support set
\begin{align}
	F = \frac{2\cdot\text{precision}\cdot\text{recall}}{\text{precision}+\text{recall}},
\end{align}
where \textit{precision} (positive predictive value) is the fraction of non-zero weights found by the algorithm that are also non-zero in the true model, while recall \textit{sensitivity} is the fraction of non-zeros in the true model that have been identified by the algorithm. Here a given weight $x_{i,t}$ is identified as being non-zero if the posterior mean of $z_{i,t}$ is above 0.5.
\\
\\
The code is available at \url{https://github.com/MichaelRiis/SSAS}.

\subsection{Experiment 1: The Effect of the Prior}\label{sec:experiment_1}
\begin{figure}[ht]
\centering
\subfigure[Signal]{\includegraphics[width = 0.45\textwidth]{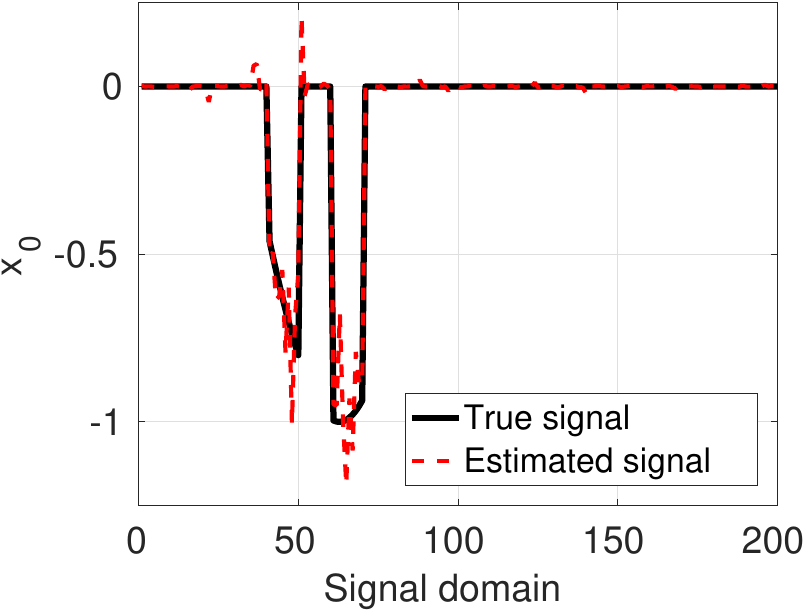}}
\subfigure[Support]{\includegraphics[width = 0.45\textwidth]{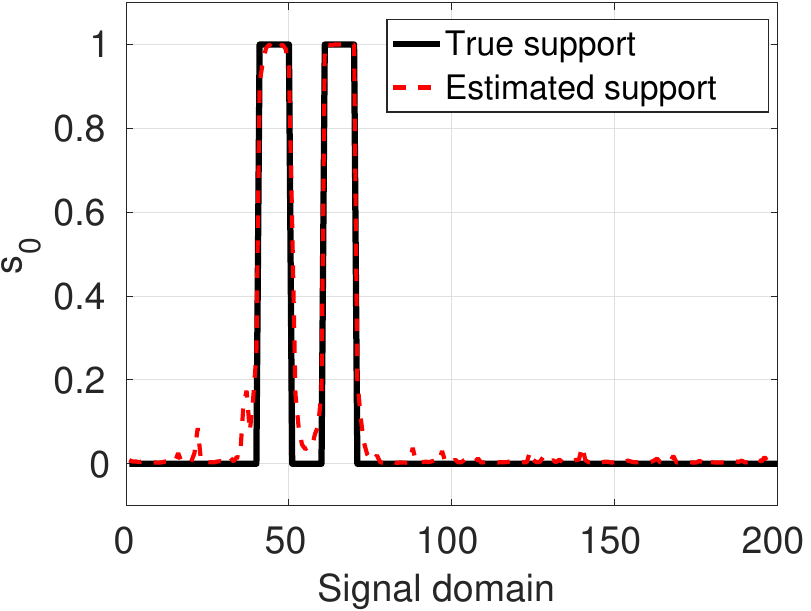}}
\caption{(a) Synthetic test signal $\x_0$ superimposed with the posterior mean of the test signal. The active coefficients are sampled from a cosine function (b) The support of the test signal superimposed with the posterior support probabilities. }
\label{fig:exp_test_signal}
\end{figure}

\begin{figure}[t]
\centering
\subfigure[Log likelihood]{\includegraphics[width = 0.45\textwidth]{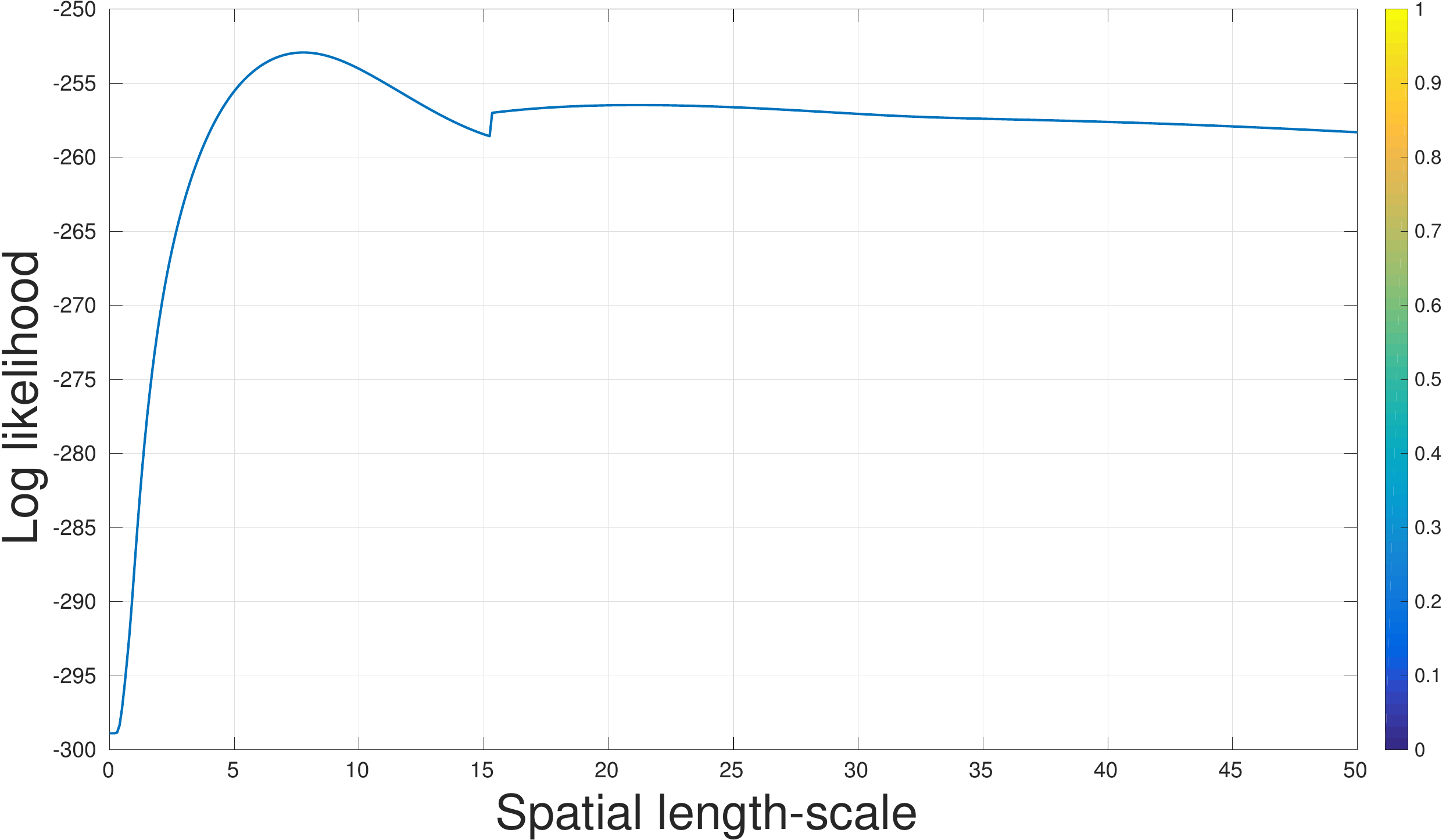}}
\subfigure[Log likelihood]{\includegraphics[width = 0.45\textwidth]{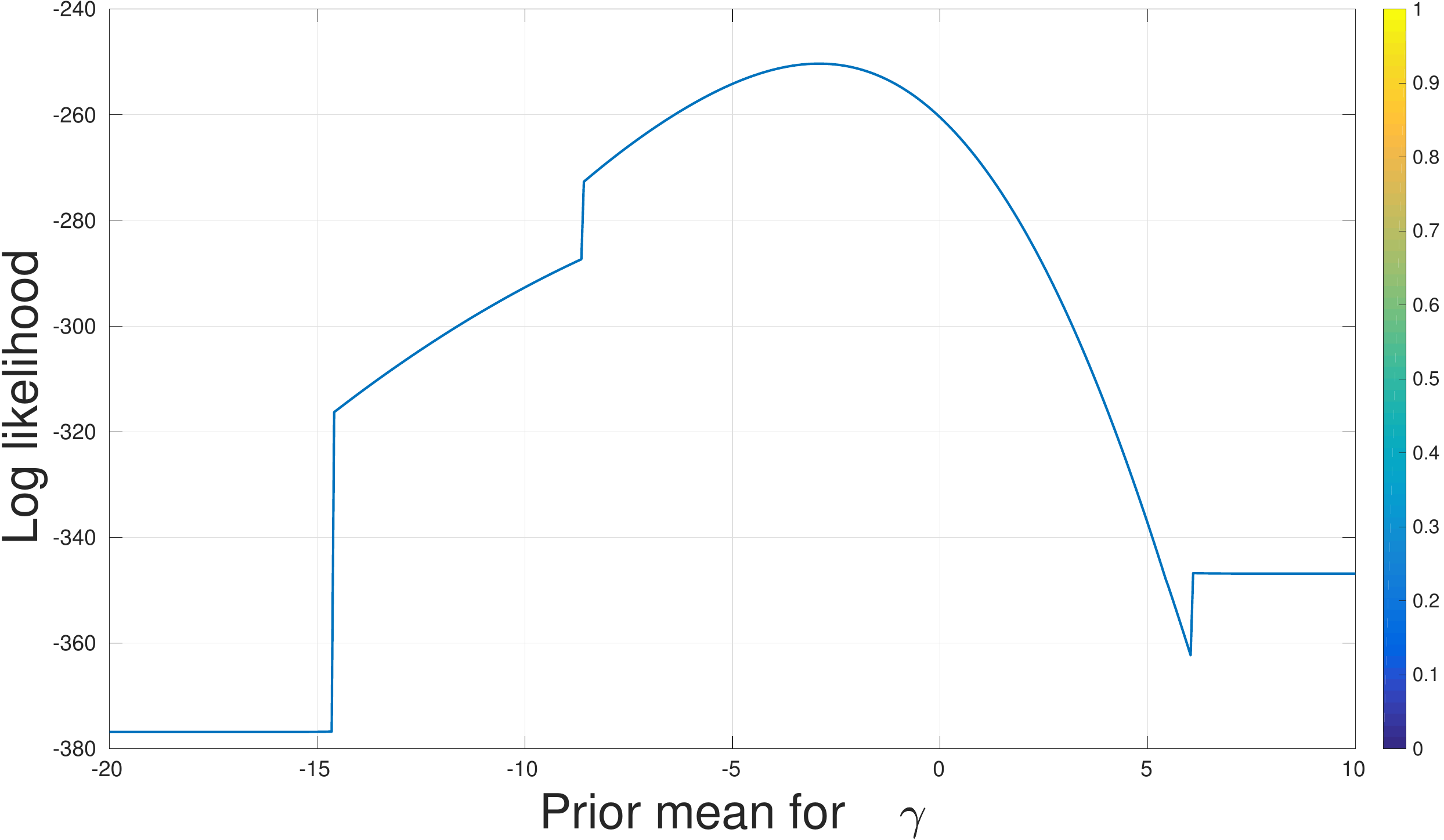}}

\subfigure[Posterior mean of $\gamma$]{\includegraphics[width = 0.45\textwidth]{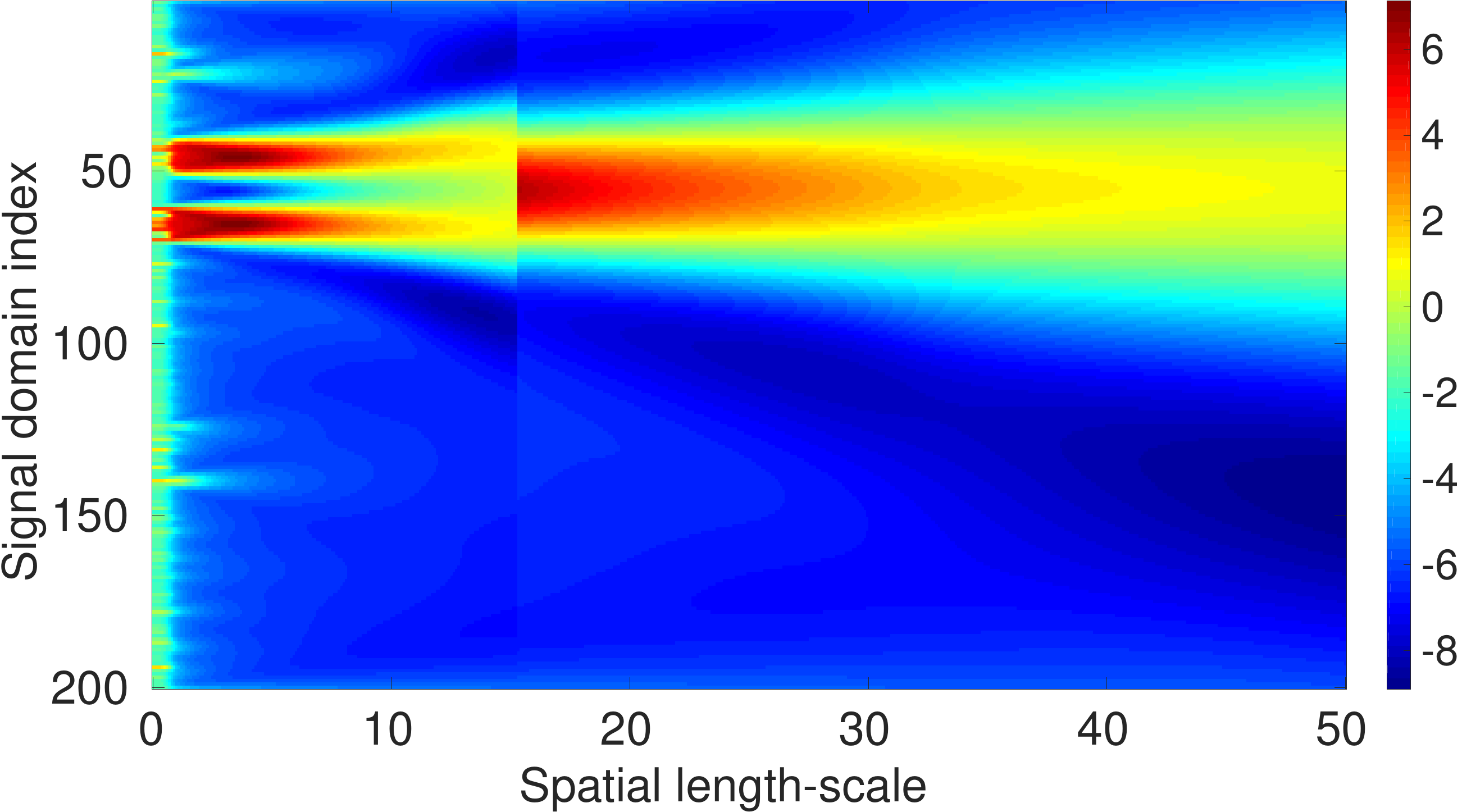}}
\subfigure[Posterior mean of $\gamma$]{\includegraphics[width = 0.45\textwidth]{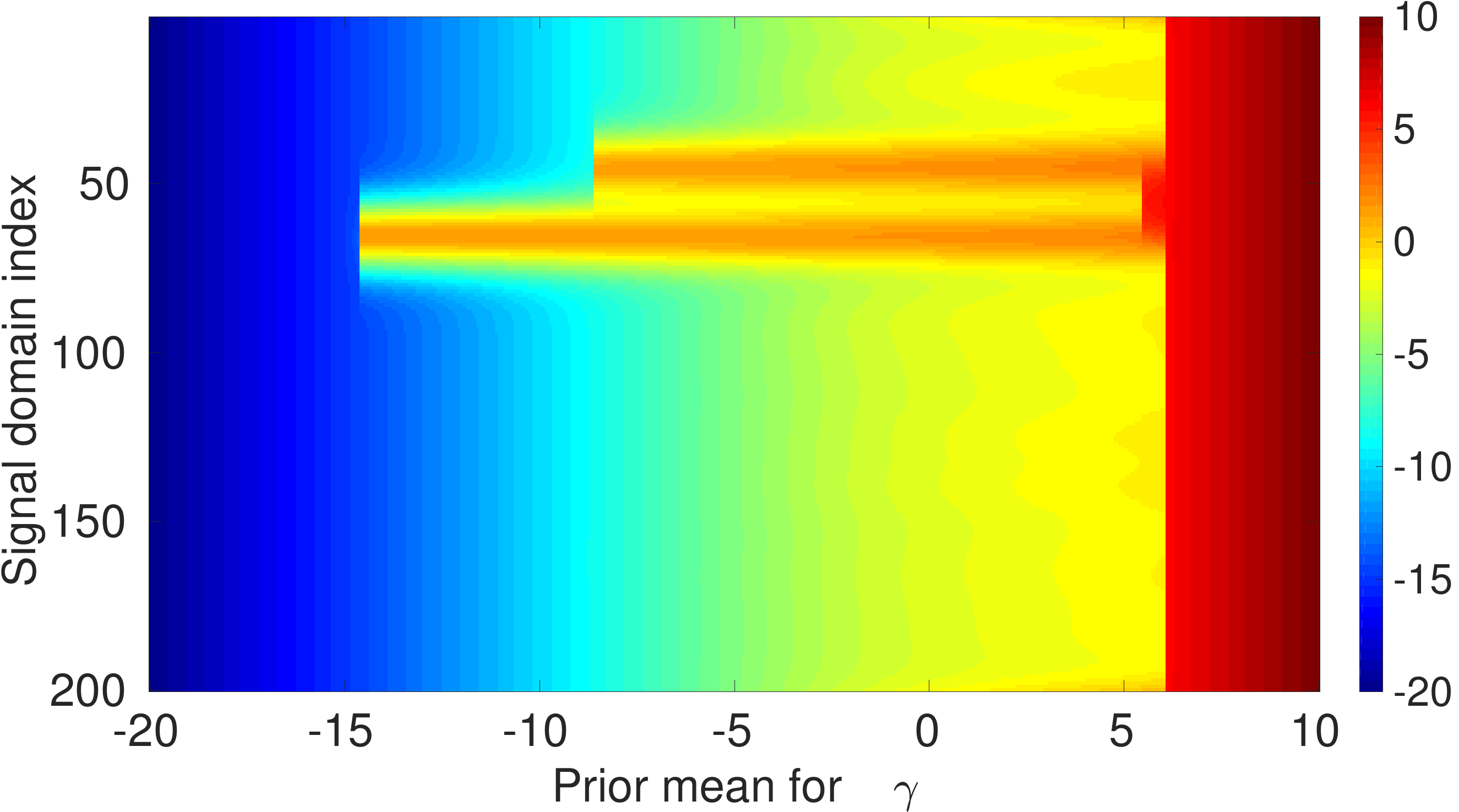}}

\subfigure[Posterior mean of $z$]{\includegraphics[width = 0.45\textwidth]{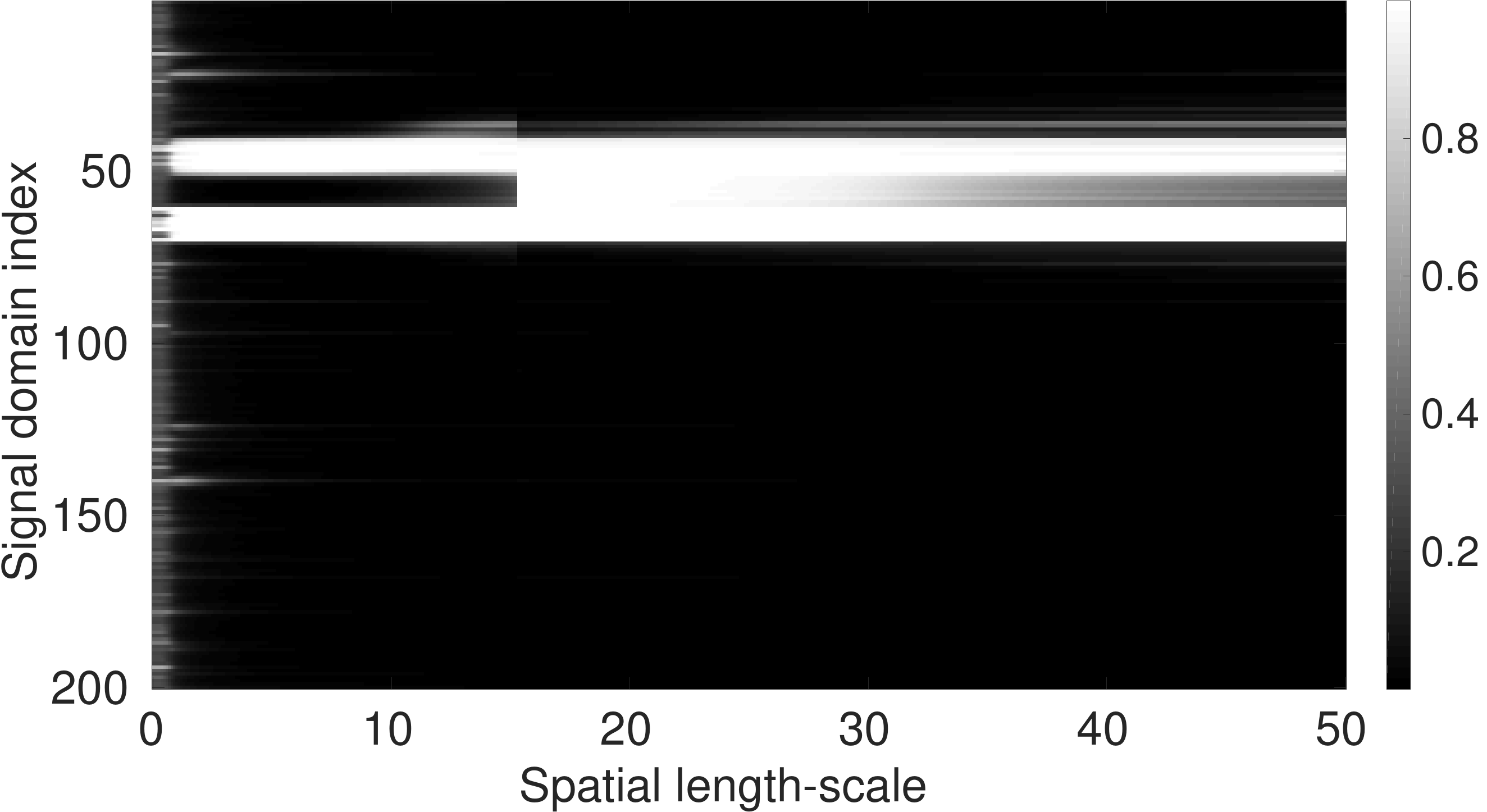}}
\subfigure[Posterior mean of $z$]{\includegraphics[width = 0.45\textwidth]{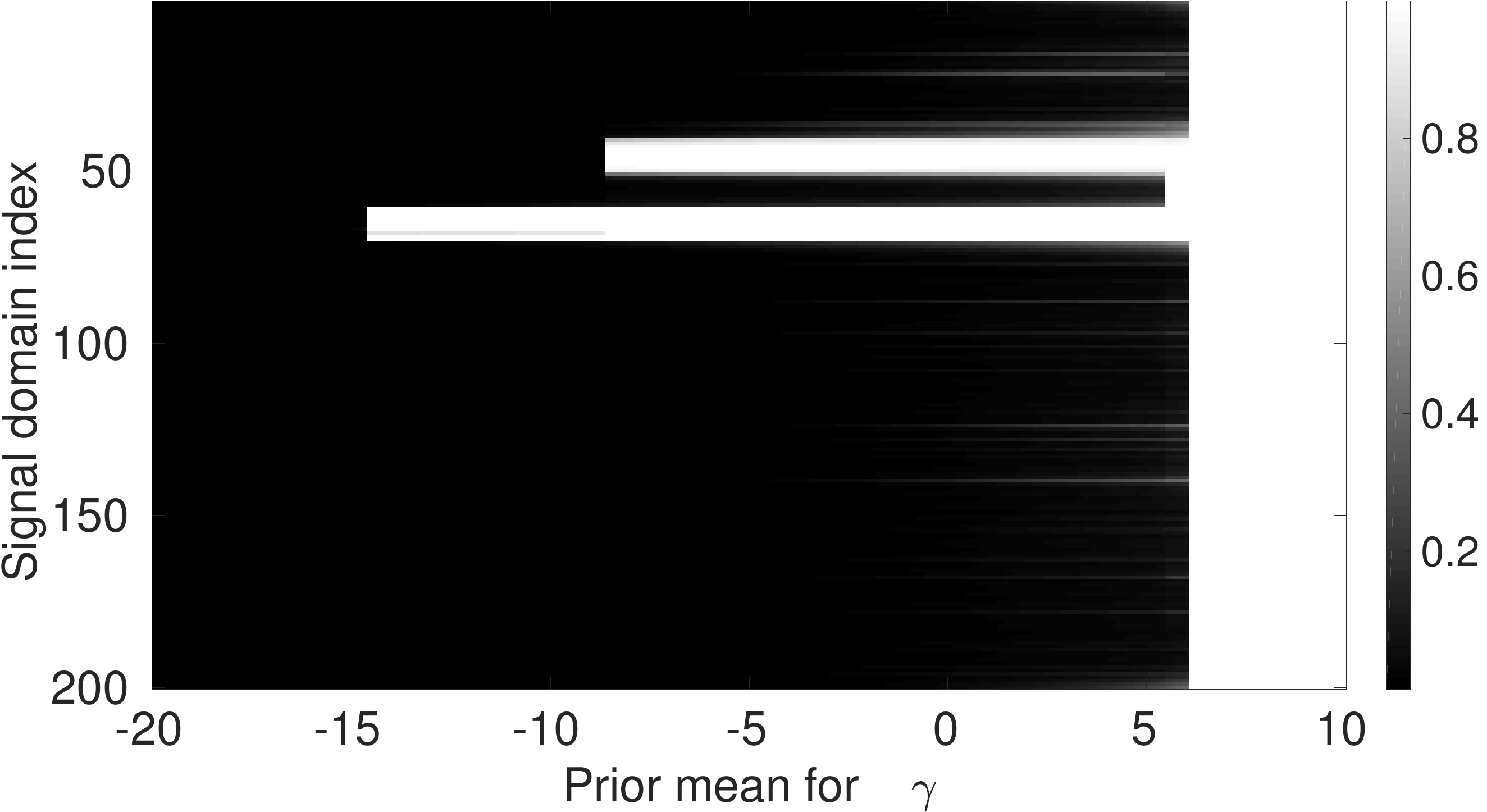}}
\caption{The effect of the spatial prior distribution. The left-most column shows the approximate marginal log likelihood, posterior mean of $\bm{\gamma}$ and posterior mean of $z$ as a function of the prior length-scale of $\bm{\gamma}$. The right-most column shows similar plots as a function of the prior mean $\nu_0$ of $\bm{\gamma}$. }
\label{fig:experiments_sweep1}
\end{figure}

In this experiment, we investigate the effect of the structured spike-and-slab prior on the reconstructed support set. For simplicity we only consider spatial structure, $T = 1$, and we further assume that the spatial coordinates are on a regular 1D grid. We construct a sparse 1D test signal $\x_0 \in \mathbb{R}^{200}$, where the active coefficients are sampled from a cosine function, see Figure \ref{fig:exp_test_signal}(a)--(b). Based on this signal we generate a synthetic data set using the linear model $\y = \A\x_0 + \e$, where $A_{ij} \sim \N\left(0 ,1\right)$, $\e \sim \mathcal{N}(0, 5\bm{I})$ is isotropic Gaussian noise (SNR $\approx 5$dB) and the number of samples is $N = 0.5D$. The prior on $\bm{\gamma}$ is of the form $p(\bm{\gamma}) = \N\left(\bm{\gamma}|\bm{\mu}, \bm{\Sigma}\right)$, where
$$\bm{\mu} = \nu \cdot \bm{1} \quad\text{and}\quad \bm{\Sigma}_{ij} = \kappa^2_1 \exp\left(-\frac{D^2_{ij}}{2\kappa_2^2}\right).$$
We sample the length-scale $\kappa_2$ equidistantly $100$ times in $\left[10^{-3}, 50\right]$ and run the algorithm on the synthetic data set for each value of $\kappa_2$. For this experiment we use the standard EP scheme with no further approximations. The noise variance is fixed to the true value and the remaining hyperparameters are fixed $\nu = 0$, $\tau = 1, \kappa^2_1 = 5$.
The posterior results are shown in the panels in leftmost column in Figure \ref{fig:experiments_sweep1}. The topmost panel shows the marginal likelihood approximation as a function of the spatial length scale $\kappa_2$. The panel in the middle shows the posterior mean $\E_{Q\left(z_i|\y\right)}\left[\gamma_i\right]$, as a function of the scale $\kappa_2$. That is, each column in the image corresponds to the posterior mean of $\bm{\gamma}$ for a specific value of $\kappa_2$. The panel in the bottom shows a similar plot for the posterior support probabilities $\E_{Q(z_i|\y)}\left[z_i\right]$. 
\\
\\
When $\kappa_2$ is close to zero the posterior mean vectors for both $\bm{\gamma}$ and $\z$ are very irregular and resemble the solution of an independent spike-and-slab prior. As the length-scale increases the posterior mean vector $\bm{\gamma}$ becomes more and more smooth and eventually give rise to well-defined clusters in the support. The algorithm recovers the correct support for $\kappa_2 \in \left[3, 15\right]$. However, at $\kappa_2 \approx 15$ a discontinuity is seen. Since the prior distribution on $\z$ does not exhibit any phase transitions with regard to $\kappa_2$, this is likely to be an effect of a unimodal approximation to a highly multimodal distribution. The discontinuity is also present in the marginal likelihood approximation as seen in the top panel and therefore one should be cautious when optimizing the marginal likelihood using line search based methods. We repeated this experiment for multiple realizations of the noise and the discontinuity was only present occasionally.
\\
\\
The rightmost column in Figure \ref{fig:experiments_sweep1} shows equivalent figures for a sweep over $\nu$, which is the prior mean of $\gamma_i$, where it is seen that the algorithm recovers the correct support for $\nu \in \left[-15, 0\right]$. It is seen that when $\nu$ is below some threshold $\nu_{\text{lower}}$, the posterior mean of $z_i$ is close to zero for all $i \in \left[D\right]$. The total number of active variables increases with $\nu$, until $\nu$ surpasses an upper threshold  $\nu_{\text{upper}}$, where all variables are included in the support set. It is also seen that variables are included cluster-wise rather than individually, which gives rise to discontinuities in the marginal likelihood in the topmost panel. 
\\
\\
Figure \ref{fig:exp_test_signal} shows the estimated signal and the estimated support probabilities for the optimal hyperparameter values in the top row of Figure \ref{fig:experiments_sweep1}, that is the prior mean $\nu = -2.93$ and lengthscale $\kappa_2 = 7.72$, where it is seen that both the estimated coefficients $\hat{\x}$ and the estimated support $\hat{\bm{s}}$ are high-quality approximations of the true quantities. We will make these relationships more quantitative in experiment 3.

\subsection{Experiment 2: Comparison of Approximation Schemes}
\begin{figure}[tp]
\centering
\newcommand{\figwidth}{2.2cm}
\subfigure[True $\gamma_0$]{\includegraphics[height = \figwidth]{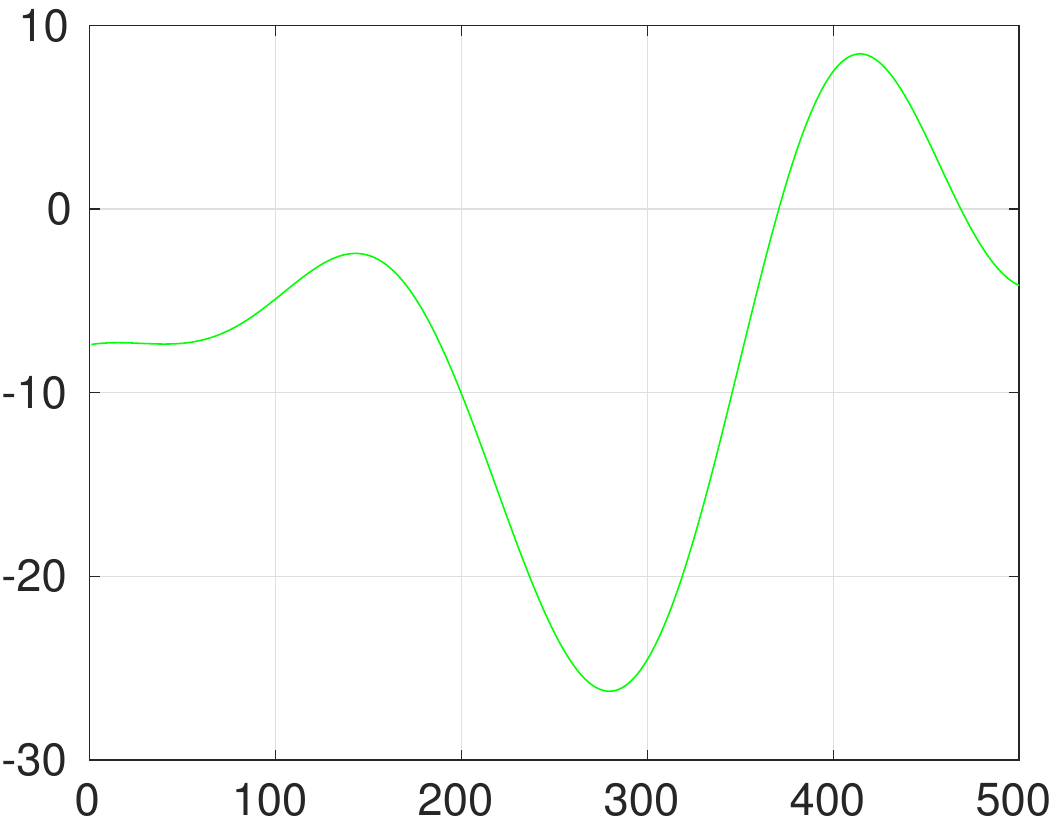}}
\subfigure[$\hat{\gamma}_{EP}$]{\includegraphics[height = \figwidth]{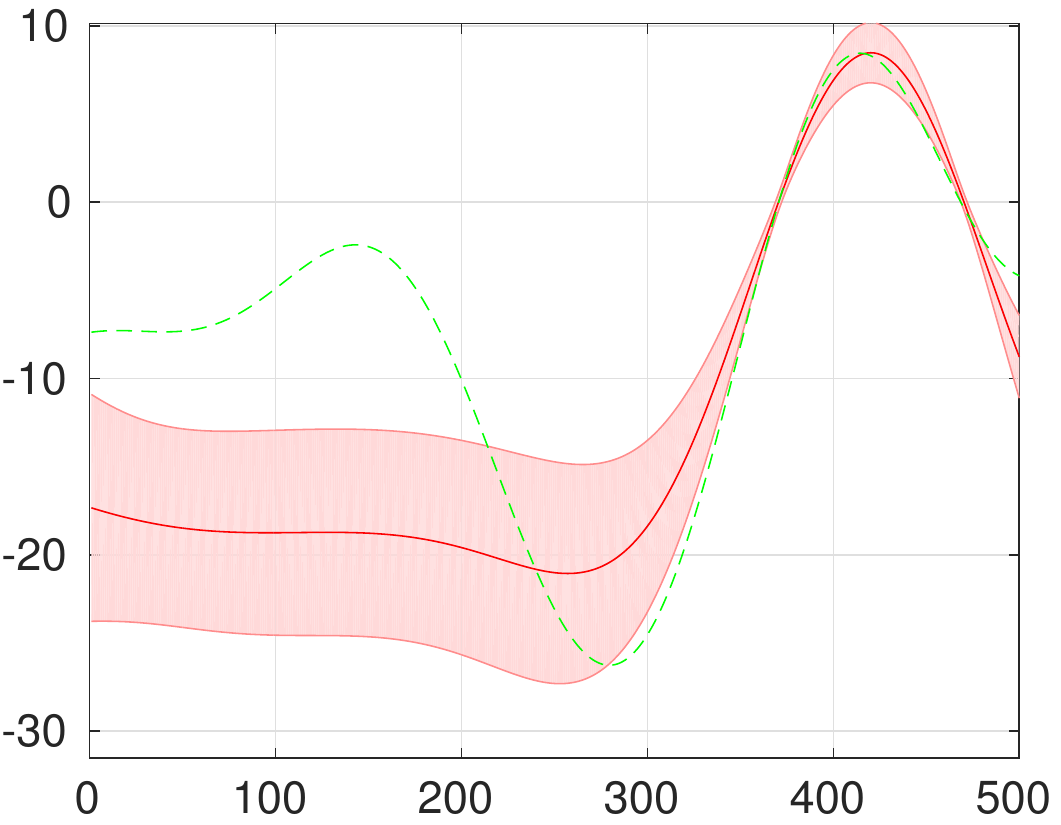}}
\subfigure[$\hat{\gamma}_{LR}$]{\includegraphics[height = \figwidth]{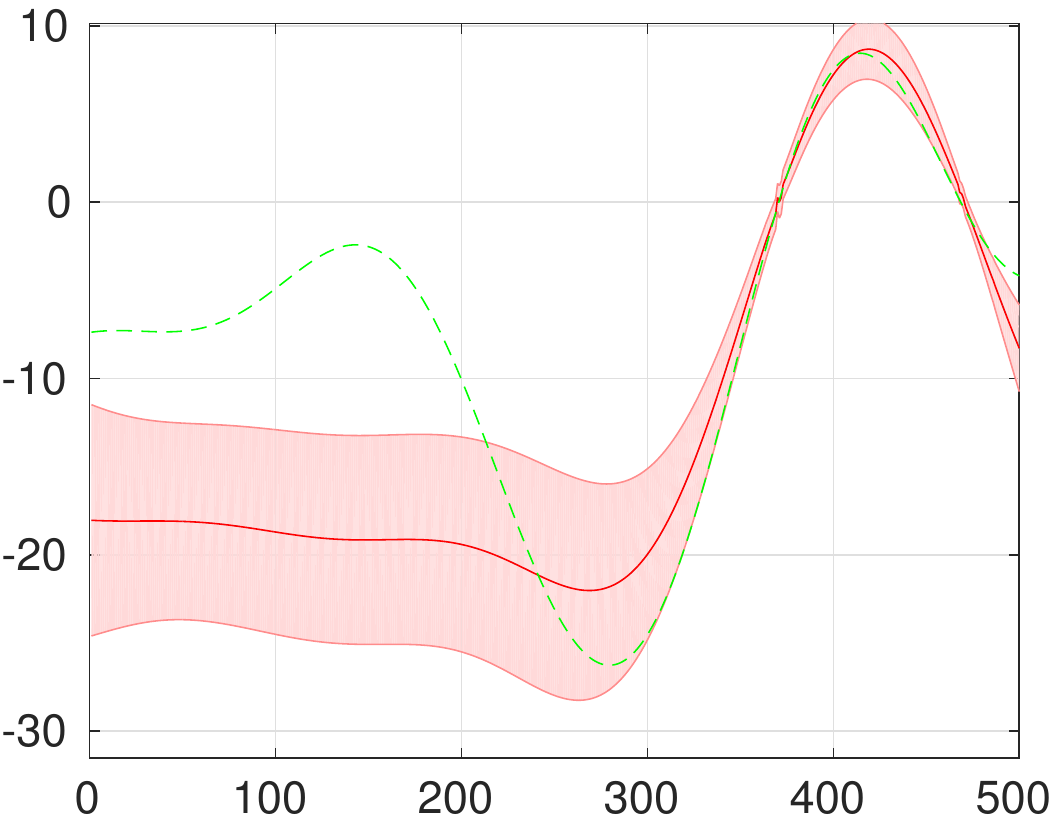}}
\subfigure[$\hat{\gamma}_{CP}$]{\includegraphics[height = \figwidth]{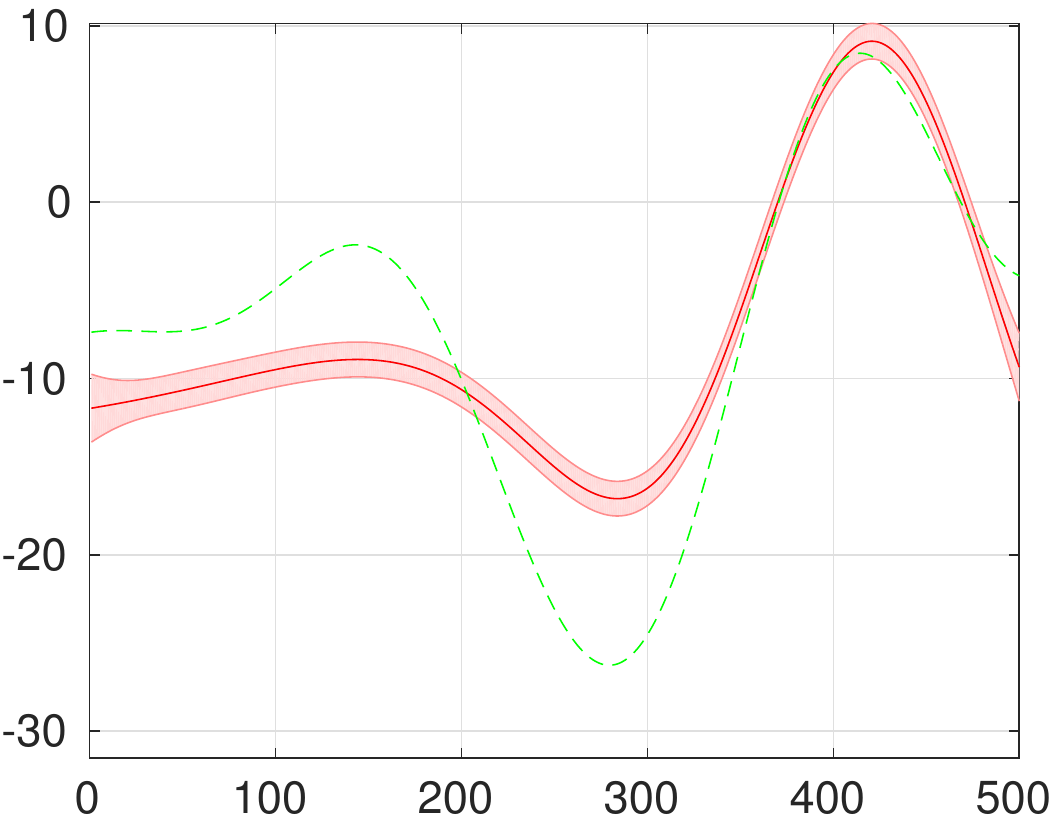}}
\subfigure[$\hat{\gamma}_{G}$]{\includegraphics[height = \figwidth]{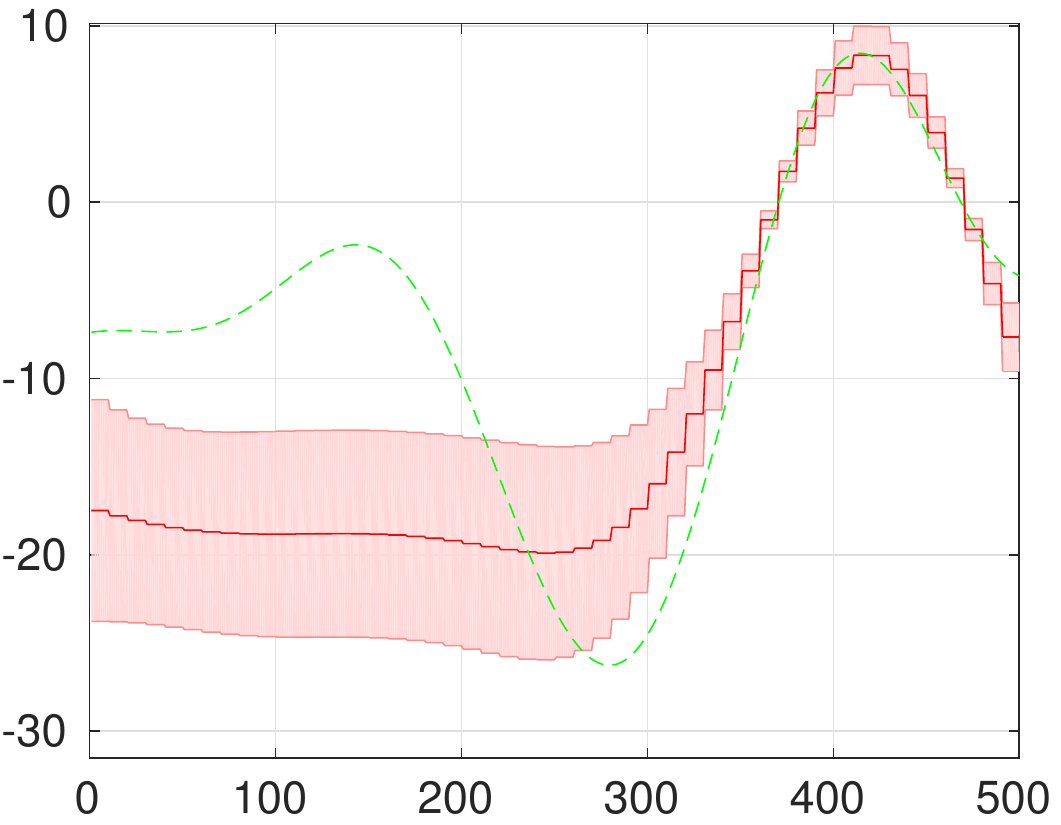}}
\subfigure[$\hat{\z}_0$]{\includegraphics[height = \figwidth]{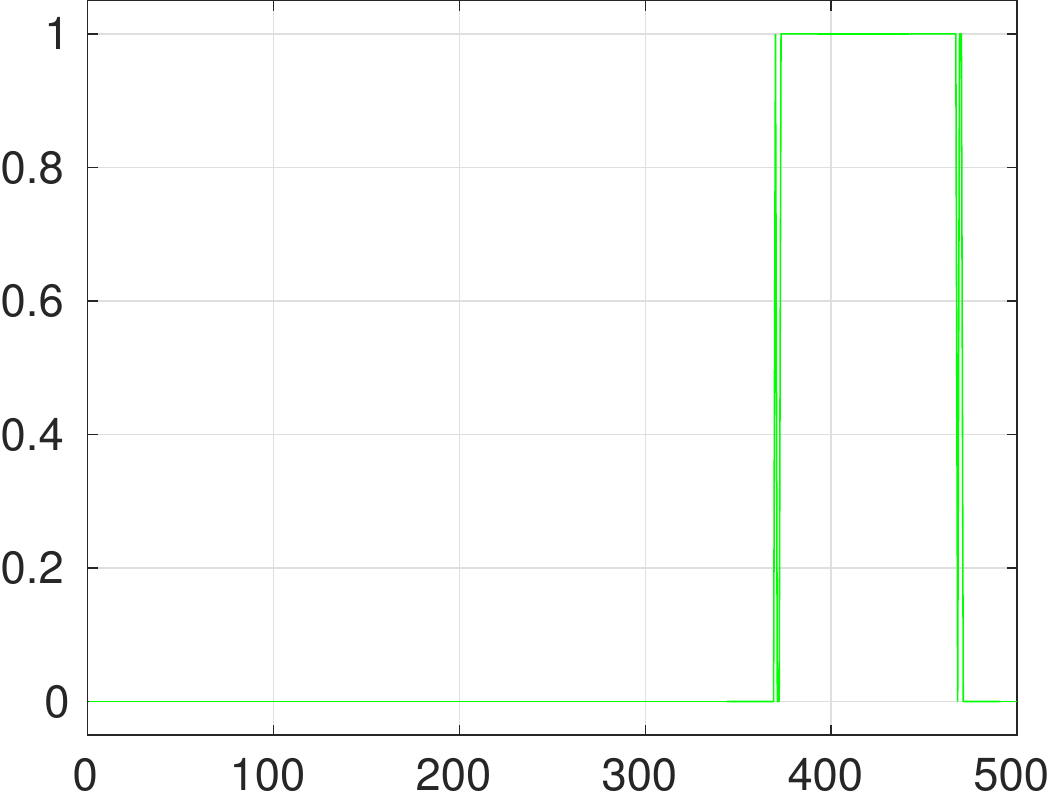}}
\subfigure[$\hat{\z}_{EP}$]{\includegraphics[height = \figwidth]{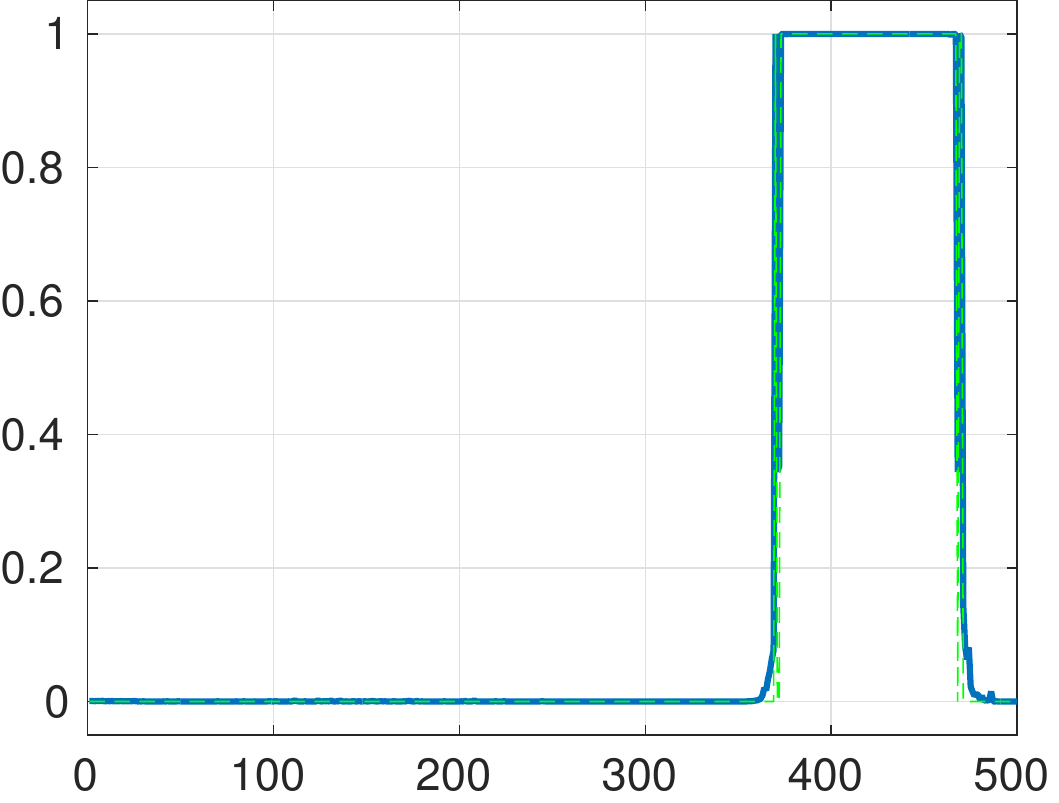}}
\subfigure[$\hat{\z}_{LR}$]{\includegraphics[height = \figwidth]{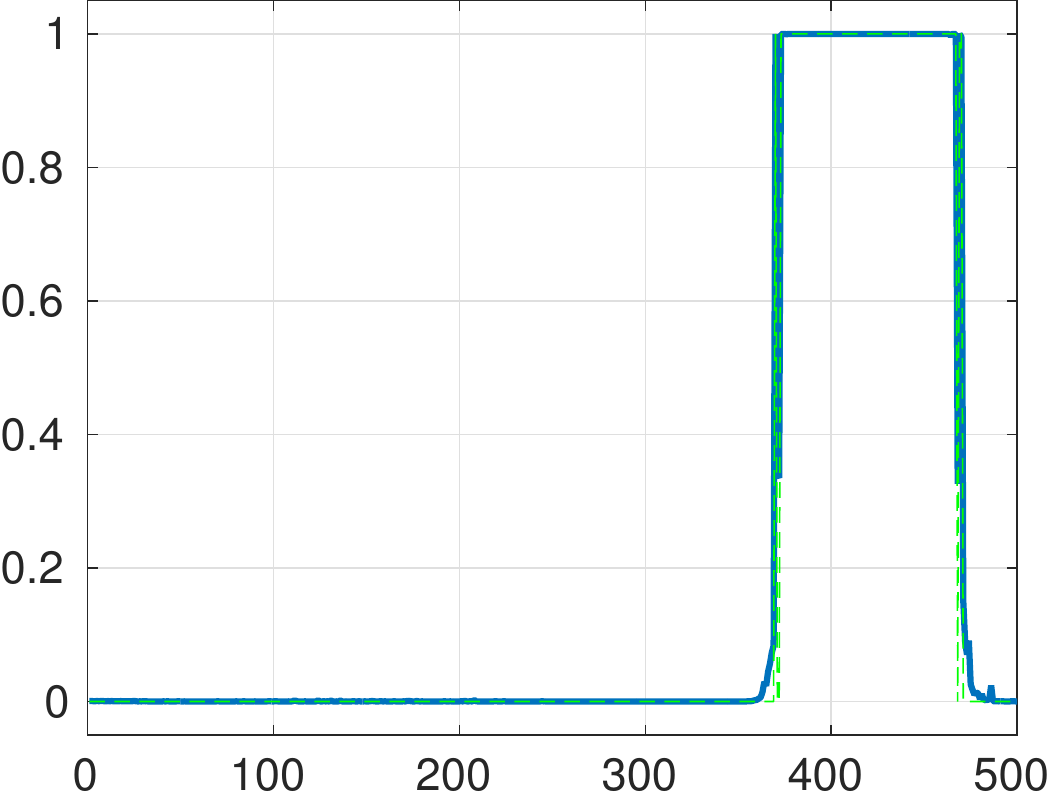}}
\subfigure[$\hat{\z}_{CP}$]{\includegraphics[height = \figwidth]{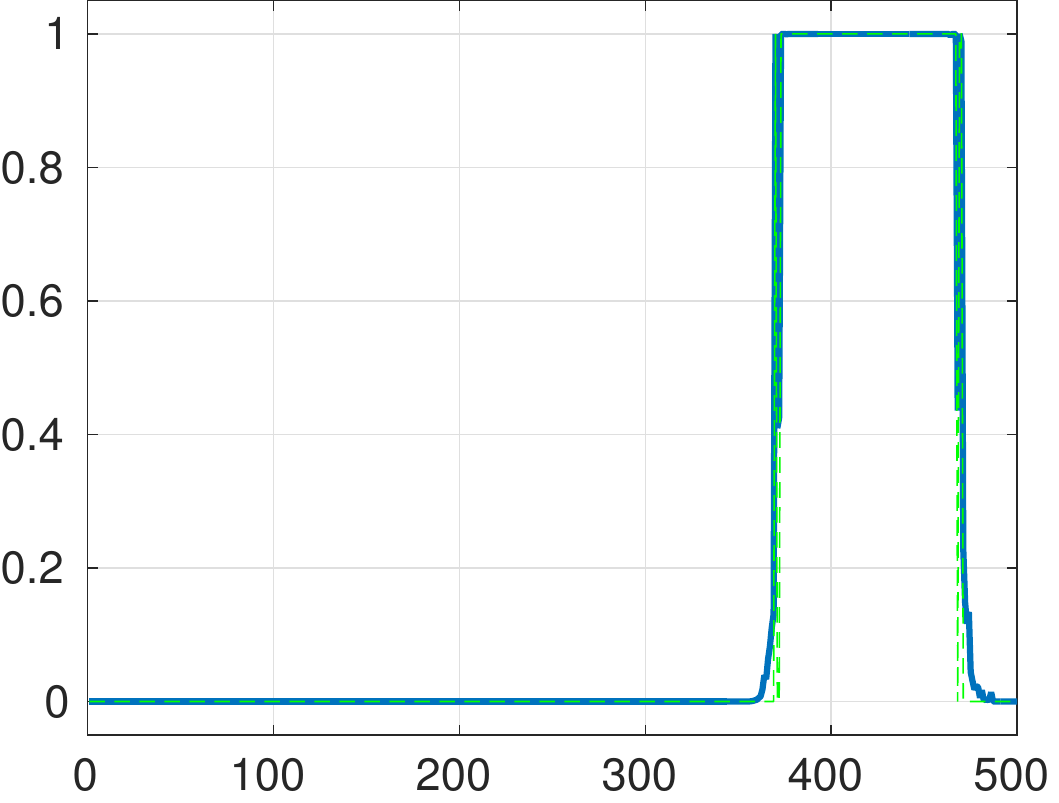}}
\subfigure[$\hat{\z}_{G}$]{\includegraphics[height = \figwidth]{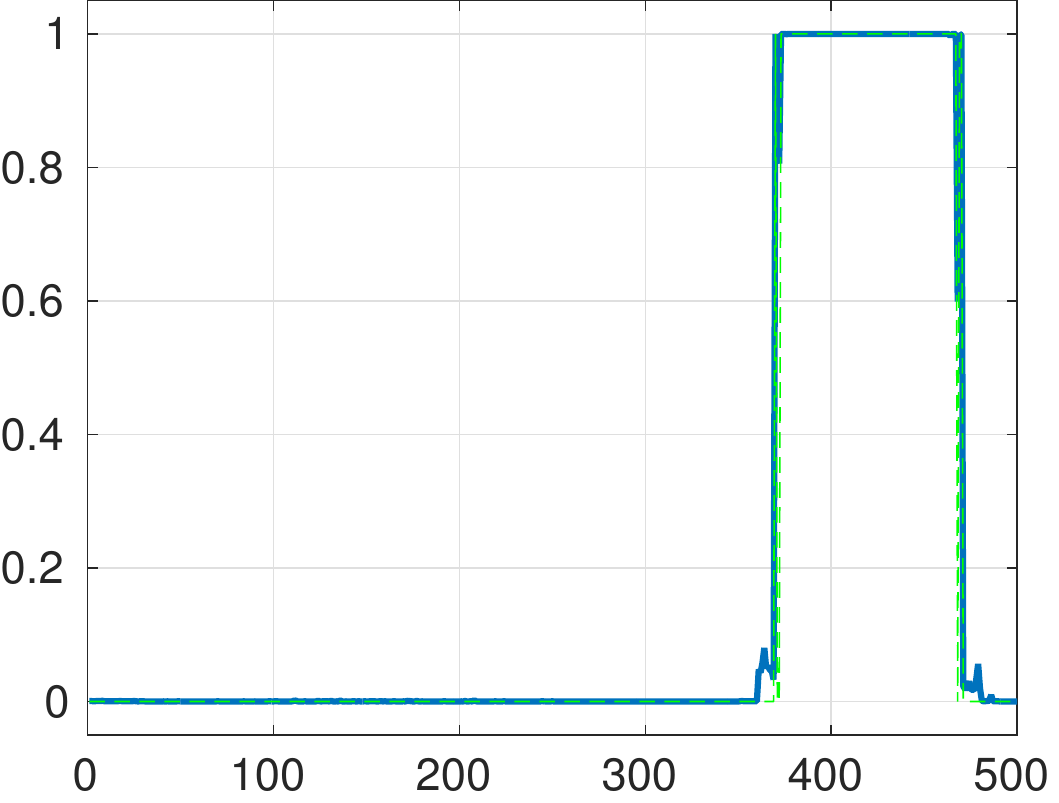}}\\
\subfigure[True $\x_0$]{\includegraphics[height = \figwidth]{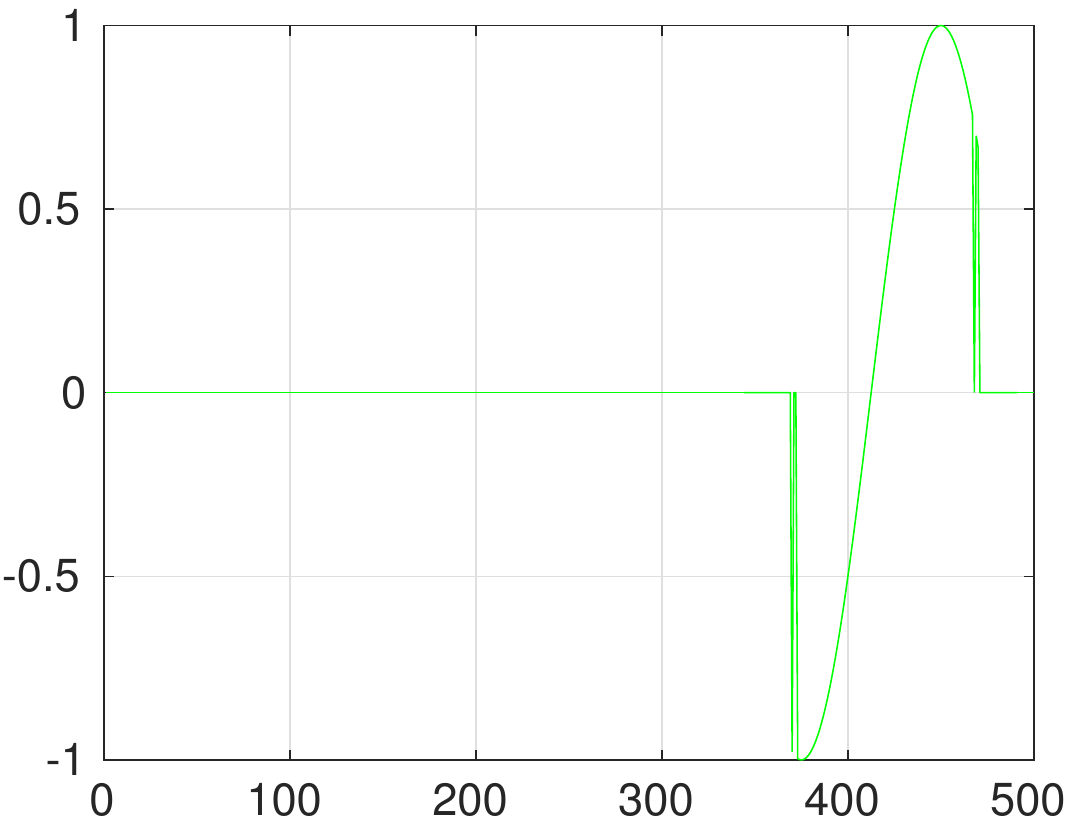}}
\subfigure[$\hat{\x}_{EP}$]{\includegraphics[height = \figwidth]{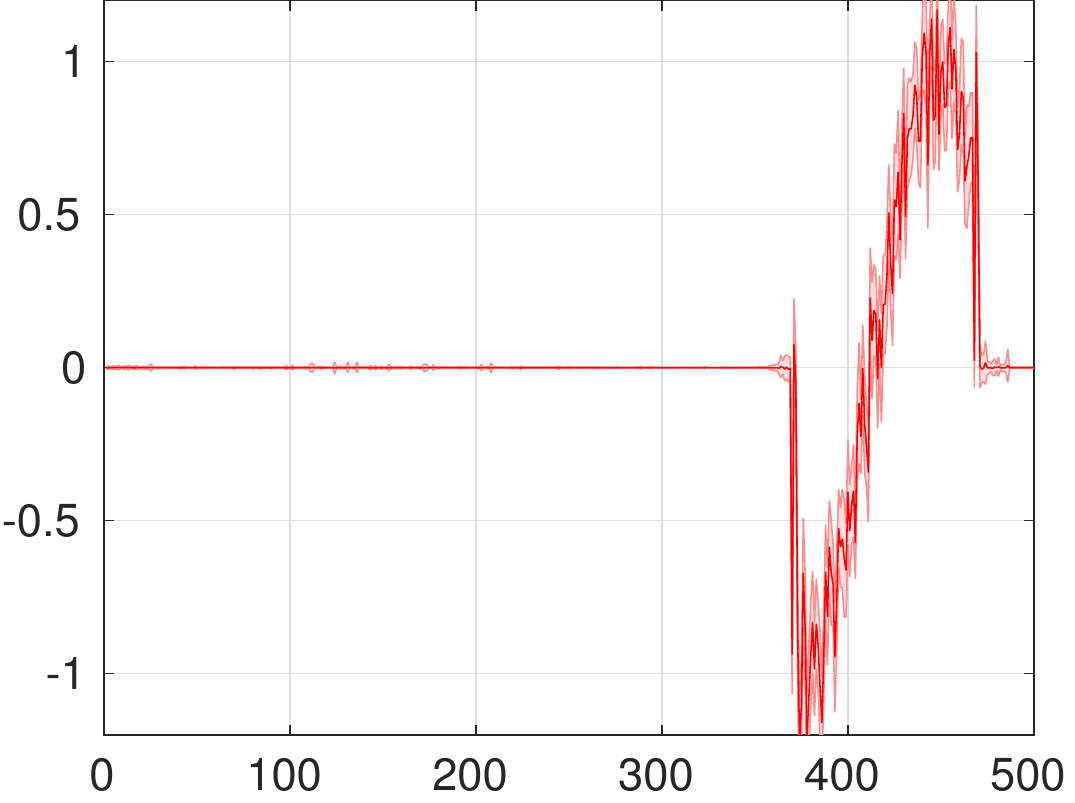}}
\subfigure[$\hat{\x}_{LR}$]{\includegraphics[height = \figwidth]{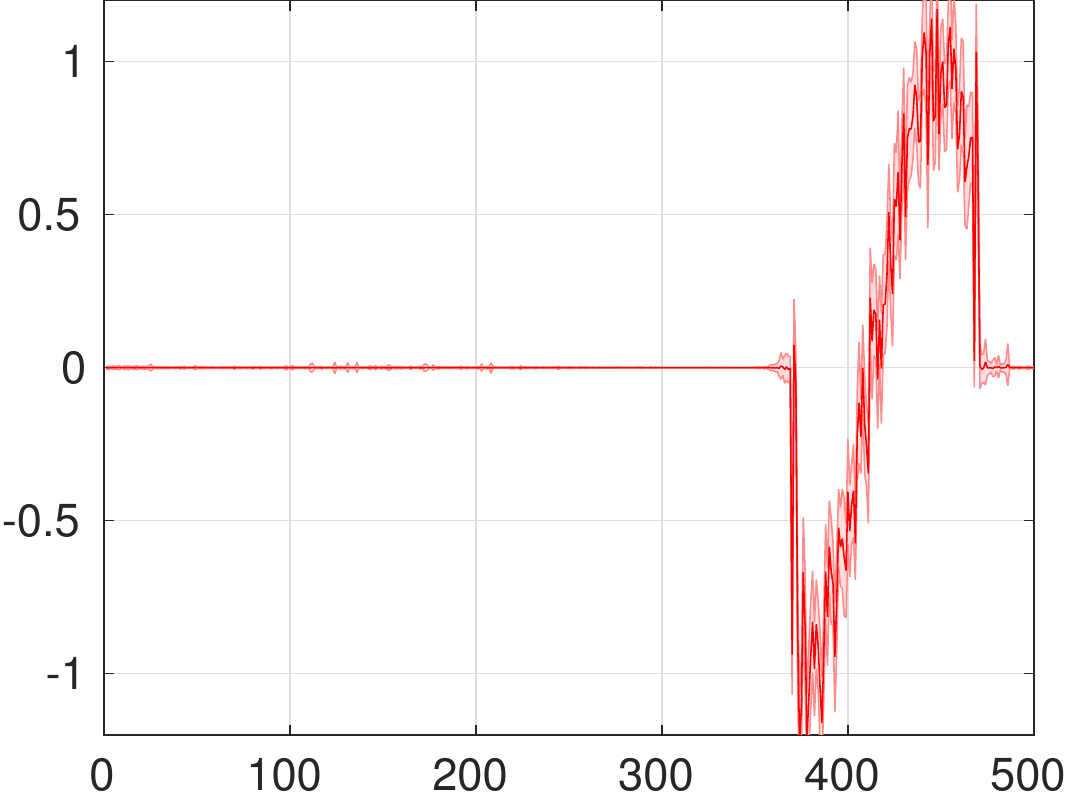}}
\subfigure[$\hat{\x}_{CP}$]{\includegraphics[height = \figwidth]{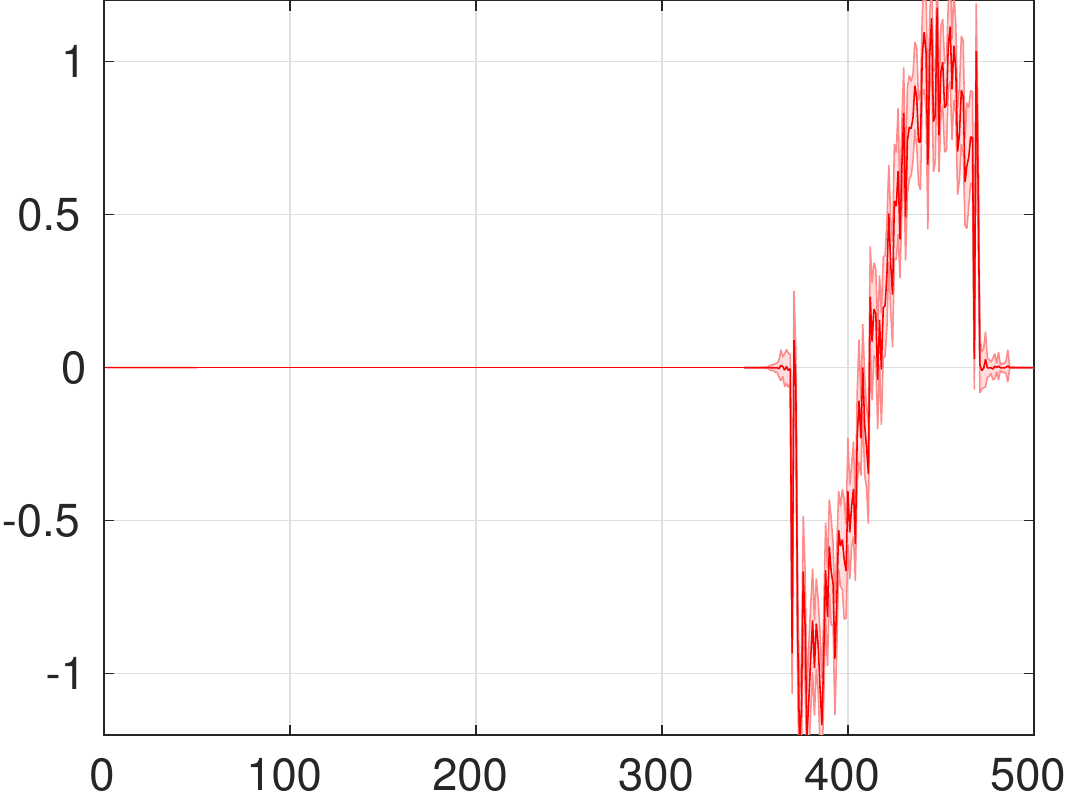}}
\subfigure[$\hat{\x}_{G}$]{\includegraphics[height = \figwidth]{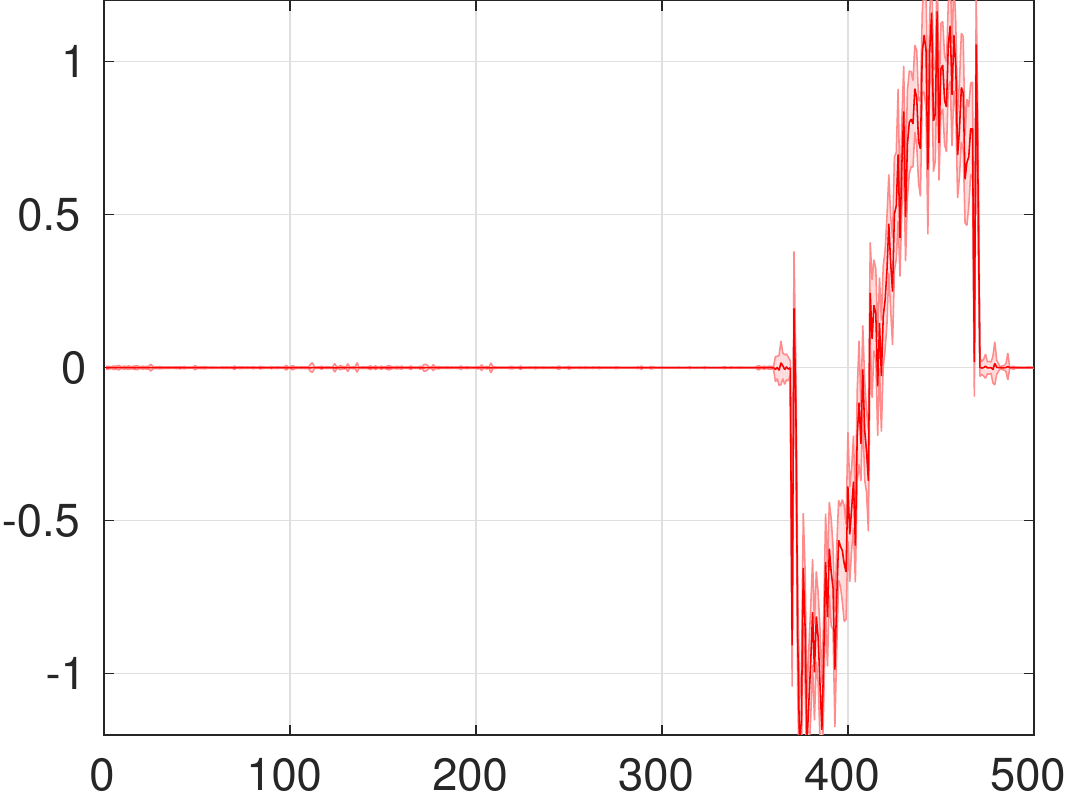}}
\caption{Comparison of approximation schemes. The panels in the first column shows a realization $\x, \z$, and $\bm{\gamma}$ from the prior distribution in eq. \eqref{eq:prior_on_x}-\eqref{eq:prior_on_gamma}. The columns 2-5 show the posterior mean quantities for EP, the low rank approximation (LR-EP), the common precision approximation (CP-EP) and the group approximation (G-EP), respectively. The pink shaded areas depict $\pm$ standard deviation.}
\label{fig:exp2}
\end{figure}

In this experiment, we investigate the properties of the proposed algorithm and the three approximation schemes: standard EP (EP), the low rank approximation (LR-EP), the common precision approximation (CP-EP) and the group approximation (G-EP). Using a similar setup as in Experiment 1, we generated a sample of $\bm{\gamma}_0$, $\z_0$ and $\x_0$ from the prior distribution specified in eq. \eqref{eq:prior_on_x}-\eqref{eq:prior_on_gamma} with $\rho_0 = 0, \tau_0 = 1$ and a squared exponential kernel with variance $\kappa^2_1 = 100$ and lengthscale $\kappa_2 = 75$. The generated sample is shown in the leftmost panels in Figure \ref{fig:exp2}. We generated observations from a linear measurement model $\y = \A\x_0 + \e$, where $A_{ij} \sim \mathcal{N}\left(0, 1\right)$ and the noise variance $\sigma^2$ is chosen such that the signal-to-noise is $20$dB. Next, we computed the posterior distributions of $\x_0$, $\z_0$ and $\bm{\gamma}_0$ from the observed measurements $\y$ using standard EP and the three approximation schemes. For the low rank approximation we included 7 eigenvectors corresponding to $99\%$ percent of the variance and for G-EP we used a group size of 10 variables. Columns 2-5 in Figure \ref{fig:exp2} show the posterior mean values for $\x, \z$, and $\bm{\gamma}$ for EP, LR-EP, CP-EP and G-EP, respectively.
Consider the posterior mean and standard deviation for $\bm{\gamma}$ for standard EP (topmost row, second column). In the region where $\bm{\gamma}_0$ is positive the posterior mean accurately recovers $\bm{\gamma}_0$ with high precision, whereas both the accuracy and the precision is lower in regions where $\bm{\gamma}_0$ is negative. The reason for the additional uncertainty is that negative values of $\gamma_i$ are in general associated with a small value of $|x_i|$, but $|x_i|$ can be small for two reason. Recall that each $x_i$ can be considered as a product $x_i = z_i\cdot c_i$, where $z_i \in \left\lbrace 0, 1\right\rbrace$ and $c_i \in \mathbb{R}$. If $z_i = 0$, then clearly $x_i = 0$, but we can still have that $x_i \approx 0$ even if $z_i = 1$ and $c_i \approx 0$ and thus the increased uncertainty. 
\\
\\
We can now compare the posterior distribution of $\bm{\gamma}_i$ for standard EP with the three approximations. Based on visual inspection one cannot tell the difference between the standard EP and EP with the low rank approximation, but the results for CP-EP and G-EP are quite different. For CP-EP it is seen that the posterior mean in the positive region is accurate, but the CP-EP approximation underestimates the uncertainty in general. The grouping effect for G-EP is clearly seen in the topmost panel in the last column, but despite the staircase pattern the posterior mean and variance are accurately recovered. The second and the third row in Figure \ref{fig:exp2} show the reconstructions of $\x$ and $\z$. We see that all of the four approaches accurately reconstruct the true quantities despite the approximation of the posterior distribution of $\bm{\gamma}$.

\subsection{Experiment 3: Phase Transitions for a Single Measurement Vector}
The purpose of this experiment is two-fold. The experiment serves to validate the inference algorithm, but it also serves to quantify the relationship between the recovery performance of the algorithm as a function of the undersampling ratio. It is well-known that the quality of the inferred solutions strongly depend on both the undersampling ratio $\delta = N/D$ and the number of non-zeros $K = ||\x||_0$ and that linear inverse problems exhibit a phase transition from almost perfect recovery to no recovery of solution $\x$ in the $\left(\delta, K\right)$-space \citep{journals/pieee/DonohoT10,journals/tit/DonohoMM11}. We hypothesize that the phase transition curves for signals with structured support can be improved, so that we can recover structured sparse signals using fewer measurements for a given number of non-zero coefficients $K$ by taking advantage of the structure. We investigate this hypothesis by measuring the recovery performance of the EP algorithms as a function of the undersampling ratio $N/D$ and compare with state-of-the-art probabilistic methods that ignore the structure of the support.
\\
\\
Using a squared exponential kernel for $\bm{\gamma}$ with variance $\kappa^2_1 = 50$ and lengthscale $\kappa_2 = 10$, we generated 100 realizations of $\x_0$ from the prior for $D = 500$. We fixed the expected sparsity to $K = \frac{1}{4}D = 125$ by choosing the prior mean of $\bm{\gamma}$ to $\nu = \phi^{-1}\left(\frac14\right)(1 + \kappa^2_1)$. As the recovery performance is very sensitive to the number of non-zero coefficients, we conditioned each sample of $\x$ on $||\x||_0 = K$ by discarding samples where $||\x||_0 \neq K$ to reduce the variance of the resulting curves for NMSE and F-measure. For each of the samples, we generated measurements $\y \in \mathbb{R}^N$ through the linear observation $\y = \A\x_0 + \e$ for a range of values for $N$. The forward model $\A$ is a Gaussian i.i.d. ensemble, where the column have been scaled to unit $\ell_2$-norm. The noise $e \sim \N\left(0, \sigma^2\right)$ is zero-mean Gaussian noise, where the noise variance $\sigma^2$ is chosen such that the signal-to-noise (SNR) ratio is fixed to $20$dB. We choose values of $N$ such that $\frac{N}{D} \in \left[0.05, 0.10, ..., 0.95\right]$.
\\
\\
We compare our methods with Bernoulli-Gaussian Approximate Message Passing (BG-AMP) \citep{vila2013expectation}, Orthogonal Matching Pursuit (OMP) \citep{Needell:2010:CIS:1859204.1859229} and an ``oracle'' estimator, which computes a ridge regression estimate based on knowledge of the true support. In this work, we use the BG-AMP method as baseline. It uses a (generalized) approximate message passing algorithm \citep{Rangan2010-uo} for inference in a probabilistic model with i.i.d. spike-and-slab priors and a Gaussian likelihood. The AMP algorithm is closely related to EP algorithm \citep{Meng2015-sx}, and the phase transition curve for BG-AMP is state of the art to the best of our knowledge. The OMP algorithm is a non-probabilistic greedy method, that iteratively select the column of $\A$ that correlate best with the current residuals until a pre-specified number of columns have been selected. The regularization parameters for the ridge regression is fixed to $\lambda = 10^{-3}$ for all runs. Finally, for comparison we also apply the proposed EP algorithm with a diagonal prior covariance matrix, which correspond to the conventional independent spike-and-slab prior (IEP). We provide BG-AMP and OMP with prior knowledge of the true number of non-zero variables in $\x_0$ and the noise variance used to generate the observations. The results are shown in Figure \ref{fig:experiment 3}.
\newcommand{\figheight}{5cm}
\begin{figure}[tp]
\centering
\subfigure[NMSE]{\includegraphics[height = \figheight]{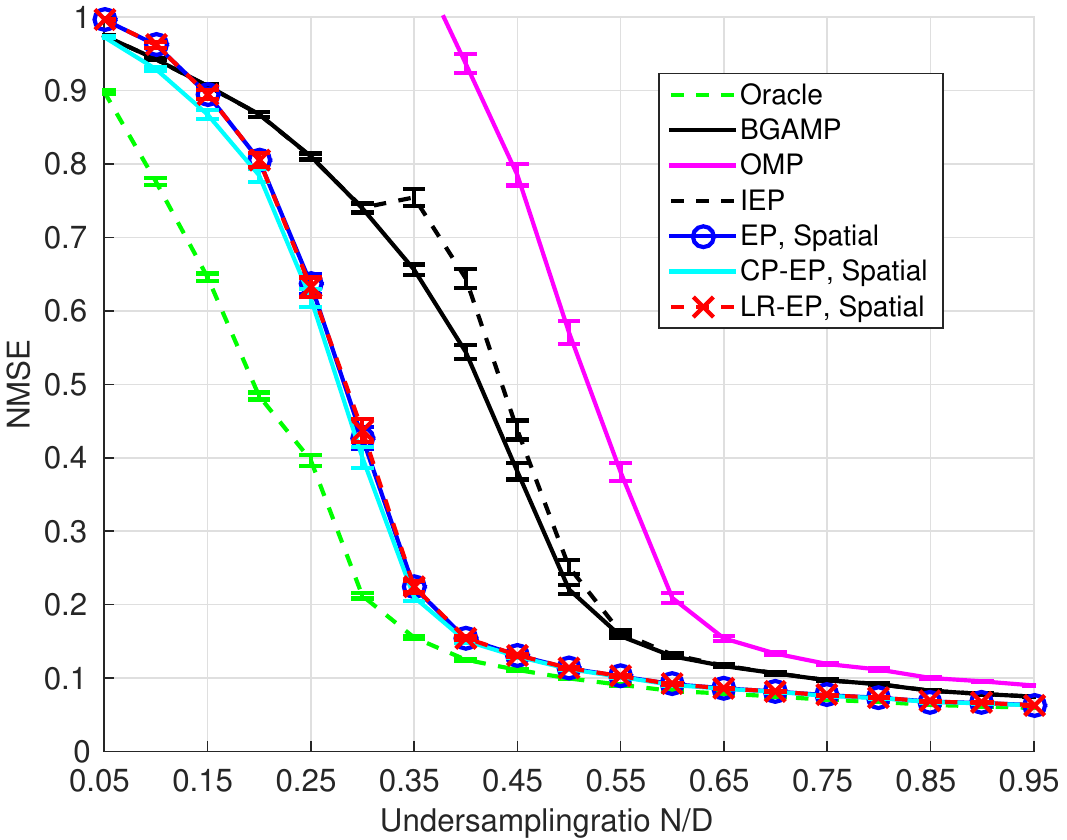}}
\subfigure[F-measure]{\includegraphics[height = \figheight]{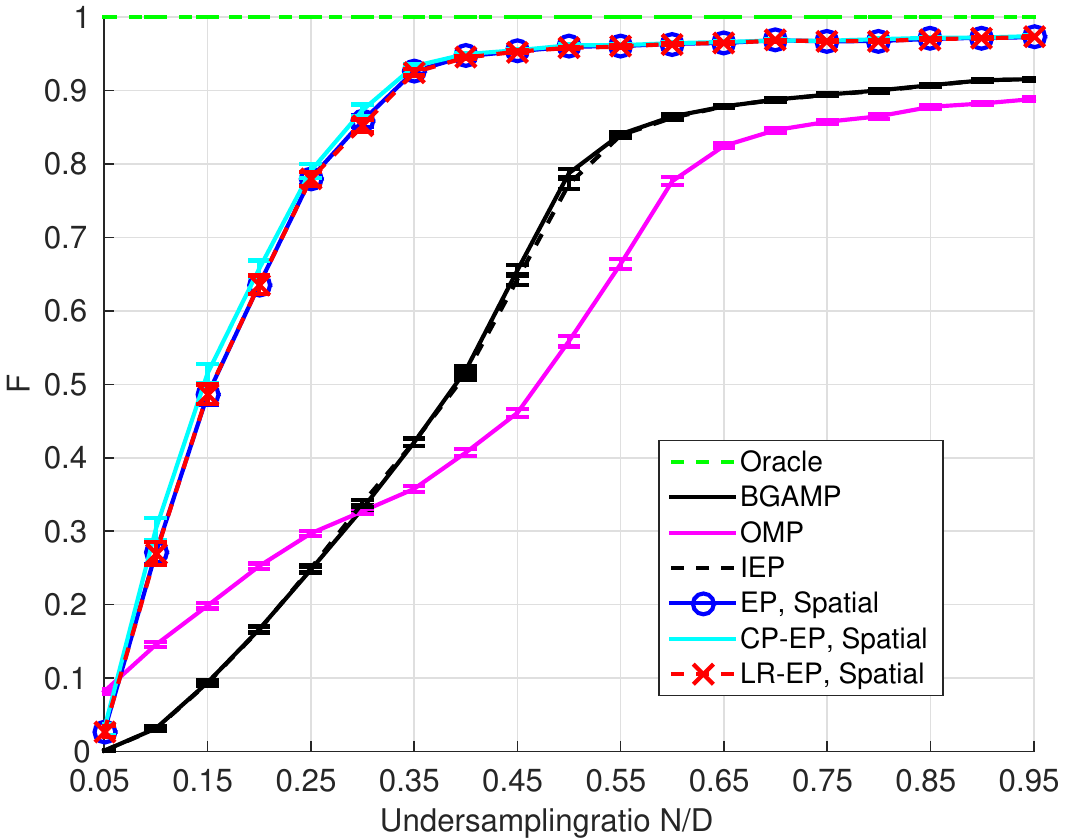}}
\subfigure[Iterations]{\includegraphics[height = \figheight]{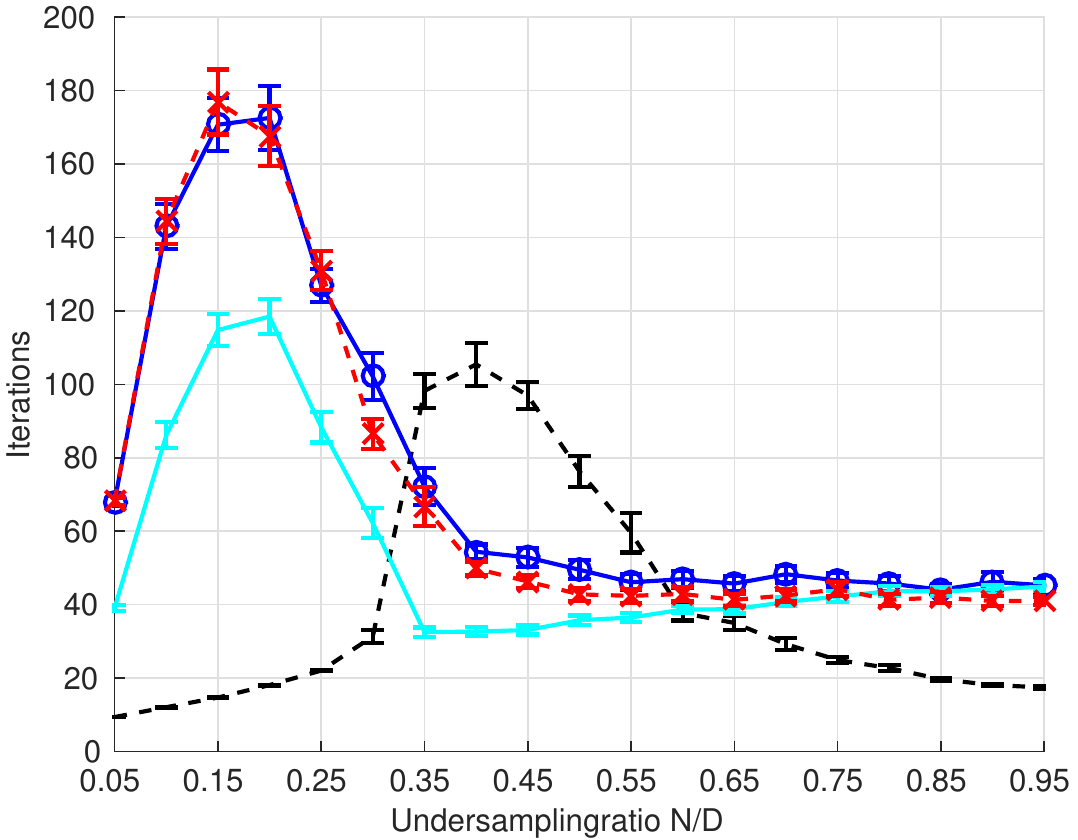}}
\subfigure[Time per iterations]{\includegraphics[height = \figheight]{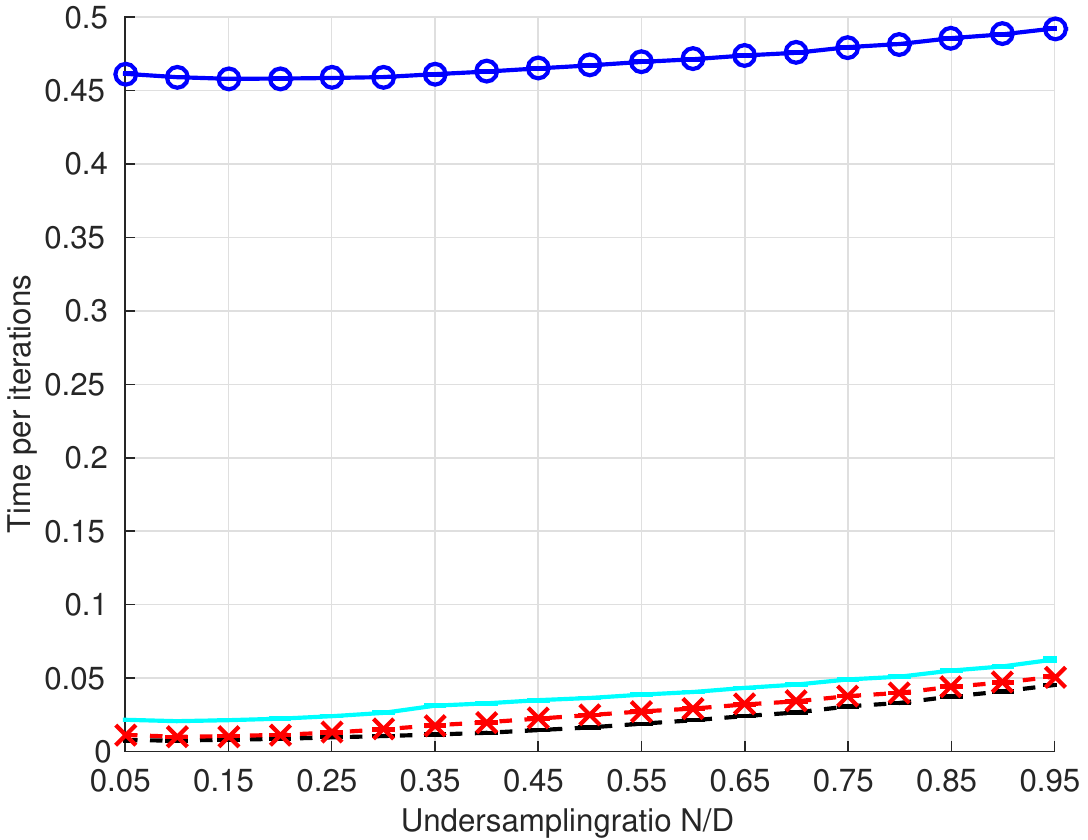}}
\caption{Performance of the methods as a function of undersampling ratio $N/D$ for $T = 1$. We compare full EP (EP, Spatial), EP with diagonal prior covariance (IEP), the common precision approximation (CP-EP) and the low rank approximation (LR-EP). The results are averaged over 100 realization. }
\label{fig:experiment 3}
\end{figure}
\\
\\
The two black curves in Figure \ref{fig:experiment 3} show the results for BG-AMP (black, solid) and EP with diagonal prior covariance (black, dashed). Both of these methods are based on conventional independent spike-and-slab priors. It is seen that the methods with prior correlation, that is EP (blue), CP-EP (cyan), and LR-EP (red, dashed), are uniformly better than the methods with independent priors both in terms of NMSE and F-measure. In fact, these methods achieve as good performance as the support-aware oracle estimator around $N/D = 0.6$ in terms of NMSE. Furthermore, it is also seen that the two approximations CP-EP and LR-EP are indistinguishable from the full EP algorithm in terms of accuracy. Panel (c) and (d) show the number iterations and the run time per iteration for the EP-based methods. Here it is seen that IEP has the lowest computational complexity per iteration, but the CP-EP and LR-EP are almost as fast. 

\subsection{Experiment 4: Compressed Sensing of Optical Characters}
\begin{figure}[tp]
\centering
\subfigure[NMSE]{\includegraphics[width = 0.32\textwidth]{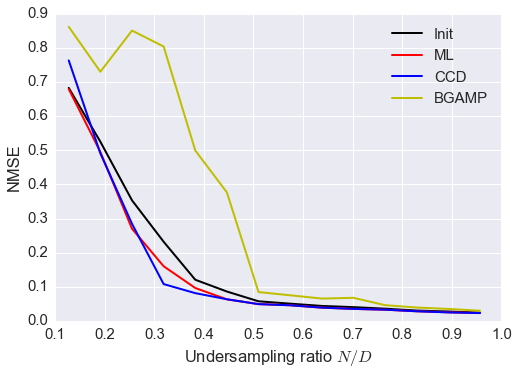}}
\subfigure[F-measure]{\includegraphics[width = 0.32\textwidth]{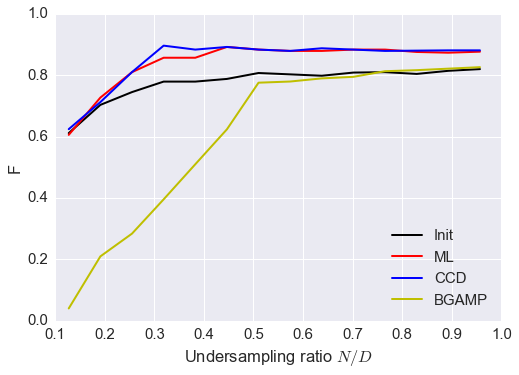}}\hfill\\
\subfigure[Bias $\kappa_1$]{\includegraphics[width = 0.32\textwidth]{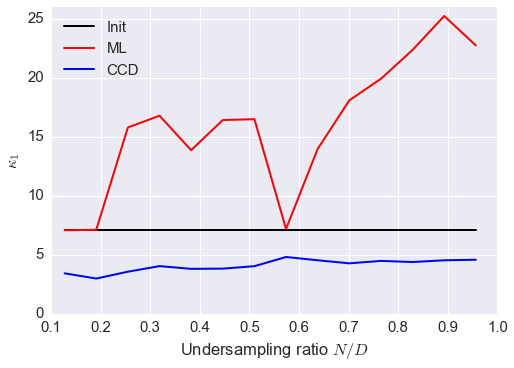}}
\subfigure[Magnitude $\kappa_2$]{\includegraphics[width = 0.32\textwidth]{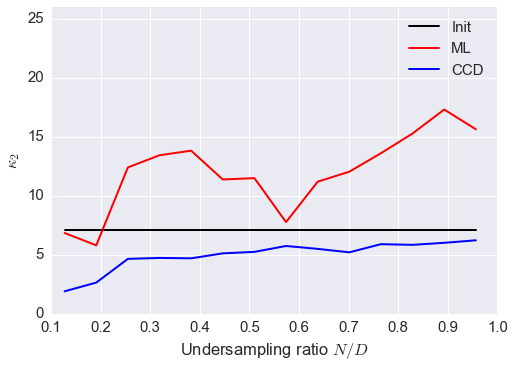}}
\subfigure[Lengthscale $\kappa_3$]{\includegraphics[width = 0.32\textwidth]{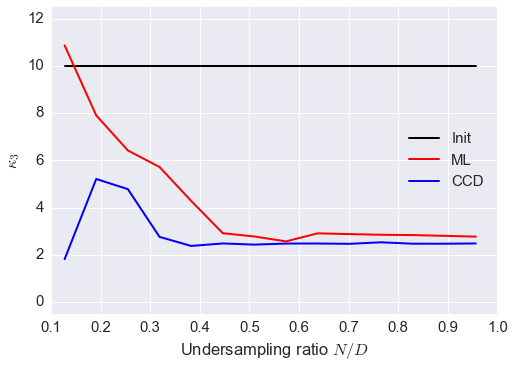}}
\caption{Performance of compressed sensing of numerical digits as a function of undersampling ratio. Panels (a) and (b) show the NMSE and F-measure, while panels (c)--(e) show the estimated values of the hyperparameters as a function of the undersampling ratio. For the CCD method, the panels (c)--(e) show the CCD-weighted average of the hyperparameters.}
\label{fig:experiments_mnist_learning}
\end{figure}

In this experiment, we apply the structured spike-and-slab model with Gaussian likelihood to an application of compressed sensing \citep{Donoho:2006:CS:2263438.2272089} of numerical characters from the MNIST data set \citep{LeCun1998-ui,JMLR:v14:hernandez-lobato13a}. The images of the numerical digits are $28 \text{ pixels} \times 28 \text{ pixels}$ and they are represented as vectors $\x_0 \in \mathbb{R}^{784}$. The objective is to reconstruct $\x_0$ from a small set of linear and noisy measurements $\y = \A\x_0 + \bm{\epsilon}$. The sensing matrix $\A$ is sampled independently from a standardized Gaussian distribution, that is $A_{ni} \sim \mathcal{N}(0, 1)$ and the noise variance is scaled such that the SNR is fixed $10\text{dB}$. 
\\
\\
We use a squared exponential kernel with a single lengthscale defined on the 2D image grid to encourage the neighbourhood structure expected in the images. We impose a Gaussian prior distribution on $\nu_0$ with zero mean and variance $\kappa_1^2$, that is $\nu_0 \sim \mathcal{N}(0, \kappa_1^2)$ and integrate over $\nu_0$ analytically to get the kernel function
\begin{align}
	k(i, j) = \kappa^2_1 + \kappa^2_2 \exp \left(  -\frac{\|\bm{d}_i - \bm{d}_j \|^2_2}{2\kappa_3^2}  \right) \label{eq:kernel_3params},
\end{align}
where $\bm{d}_i$ is the image grid coordinates of $\gamma_i$. We assume that the noise variance is known and we fix the prior mean and variance of the 'slab' component to a standardized Gaussian with $\rho_0 = 0$ and $\tau_0 = 1$. Thus, the hyperparameters to be learned are $\bm{\Omega} = \left\lbrace \kappa_1, \kappa_2, \kappa_3 \right\rbrace$. For the CCD procedure, we have to choose prior distributions for the hyperparameters. For the lengthscale parameter, we can use the fact that the 'pen' is roughly a few pixels wide on average and choose a log-normal prior with mean 4 and standard deviation 2, that is $\kappa_3 \sim \mathcal{LN}\left(4, 2^2\right)$. The $10$'th and $90$'th percentiles for this distribution are approximately 2 and 7, respectively. For the remaining two hyperparameters, we will use the same prior distribution, that is $\kappa_1, \kappa_2 \sim \mathcal{LN}\left(4, 2^2\right)$, but for a different reason than for the lengthscale parameter $\kappa_3$. The mode of the distribution $\mathcal{LN}\left(4, 2^2\right)$ is approximate $2.9$ and then the $10$'th and $90$'th percentiles of the distribution of $\phi(\gamma)$ for $\gamma \sim \mathcal{N}(0, 2.9^2)$ are approximately $0.0001$ and $0.9999$, respectively. Furthermore, the choice of lognormal priors generally works well for the CCD scheme, which can yield poor performance if the distributions have too heavy tails.
\\
\\
We use the low-rank approximation for all computations in this experiment. Figure \ref{fig:experiments_mnist_learning}(a)--(b) show the NMSE reconstruction error and F measure as a function of the undersampling ratio $\frac{N}{D}$. In this experiment, we also compare with the BGAMP method, which is informed about the noise level. We also use a standardized Gaussian as slab distribution for BGAMP. The black curves in panels (a) and (b) show the performance for the model when the hyperparameter are fixed to the initial values. It is seen that for small undersampling rates $N/D < 0.5$, we obtain slightly better results in terms of NMSE when adapting the hyperparameters, but we get a uniform improvement in terms of the F measure. Figure \ref{fig:experiments_mnist_learning}(c)--(e) show the estimated values for the hyperparameters as a function of the undersampling ratio. It is seen that the ML solution tends to overestimate the lengthscale for small sample sizes. In this case, only weak information are propagated from the observations to the prior of $\bm{\gamma}$ and thus the model becomes over-regularized. It is also seen that the bias and magnitude parameters are correlated as expected from the relationship in eq. \ref{eq:marginal_prior_prob}, (see Appendix \ref{appendix:prior} for more details).
\begin{figure}[tp]
\centering
\includegraphics[width = \textwidth]{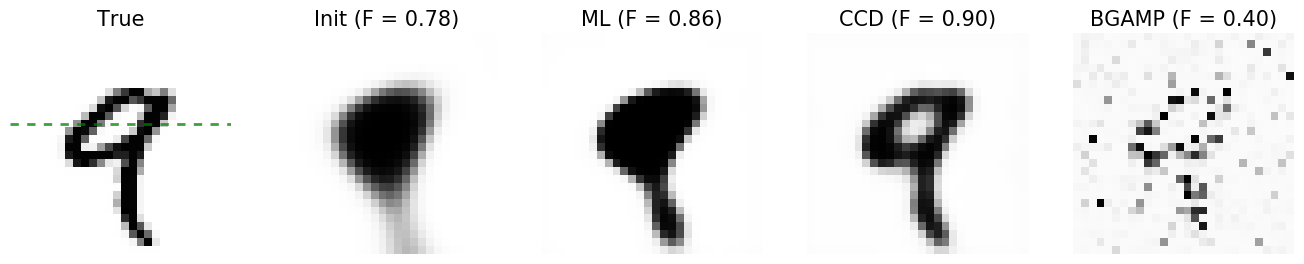}
\caption{Posterior mean of $\z$ for compressed sensing of numerical digits, where $N/D \approx 0.3$. The panels in the top row show the posterior mean of the support, while the panels in the bottom row show the posterior mean of signal. Figure \ref{fig:experiments_mnist_line} shows the posterior distributions of the row indicated the dashed line. }
\label{fig:experiments_mnist_examples}
\end{figure}
\begin{figure}[tp]
\centering
\subfigure[Posterior $\bm{\gamma}$]{\includegraphics[width=0.32\textwidth]{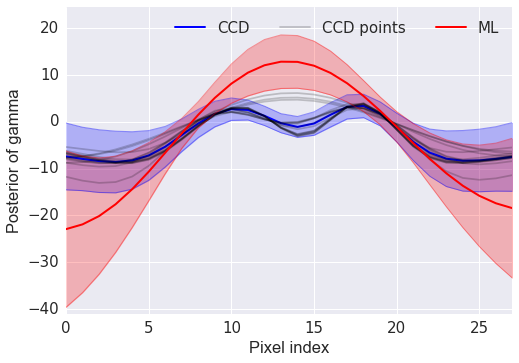}}
\subfigure[Posterior support]{\includegraphics[width=0.32\textwidth]{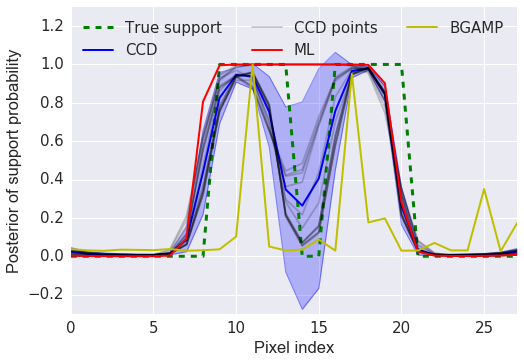}}
\subfigure[Posterior signal]{\includegraphics[width=0.32\textwidth]{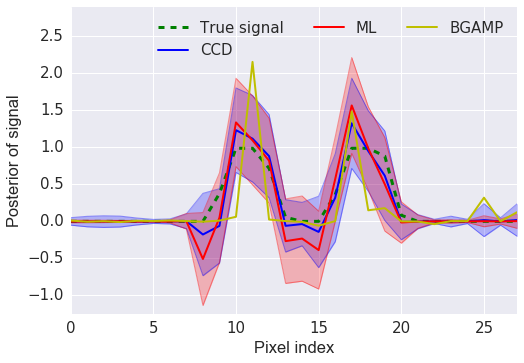}}
\caption{Comparison of the posterior distribution of $\bm{\gamma}$, $\bm{z}$ and $\bm{x}$ for the row shown by the green dashed line in Figure \ref{fig:experiments_mnist_examples} for $N/D \approx 0.3$.}
\label{fig:experiments_mnist_line}
\end{figure}
\\
\\
Figure \ref{fig:experiments_mnist_examples} shows the posterior mean of the support for each method for $N/D \approx 0.3$, where it is seen that we obtain a qualitative and quantitative improvement of the support estimate by taking a priori knowledge into account and integrating over the uncertainty. Figure \ref{fig:experiments_mnist_line} shows the posterior distribution for $\bm{\gamma}$, $\bm{z}$ and $\x$ for the line indicated by the green dashed line in Figure \ref{fig:experiments_mnist_examples}(a). Recall, that the posterior distributions obtained using CCD are finite mixture models. The thin gray lines in left and center columns show the posterior mean of the individual mixture components, while the solid colored lines and the shaded areas show the mean and variance of the mixture distributions, respectively. From the center panel, it is seen all methods fail to capture the true support perfectly, but the mean of the support of the CCD solution are significantly improved compared to the ML solution and more interestingly, the CCD solution also has high variance in the region, where it is wrong. These uncertainties are not properly reflected in the NMSE and F metrics, but the log posterior density of the true support of the ML solution is $-181.654$, while the same quantity for the CCD method and BG-AMP evaluate to $-74.181$ and $-339.065$, respectively.

\subsection{Experiment 5: Phoneme Recognition}
\begin{figure}[tp]
\centering
\subfigure[The two classes]{\includegraphics[height = 5cm]{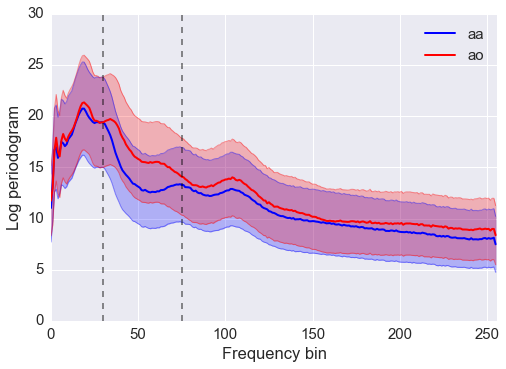}}
\subfigure[Difference in mean]{\includegraphics[height = 5cm]{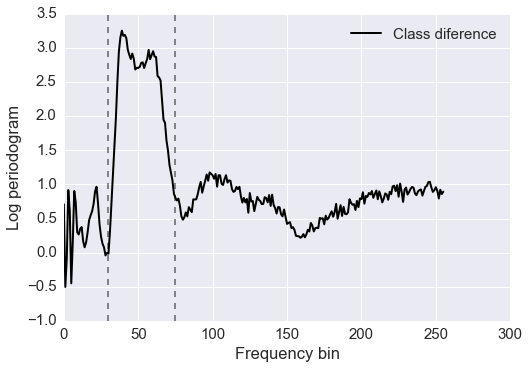}}
\caption{a) The frequency-wise mean and standard deviation of the log-periodogram of the two spoken vowels ''aa'' and ''ao''. b) The difference of the two mean signals. }
\end{figure}
\begin{figure}[tp]
\centering
\subfigure[Posterior $\bm{\gamma}$]{\includegraphics[width=0.32\textwidth]{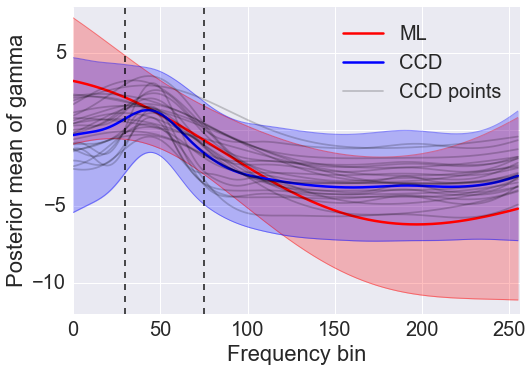}}
\subfigure[Posterior support]{\includegraphics[width=0.32\textwidth]{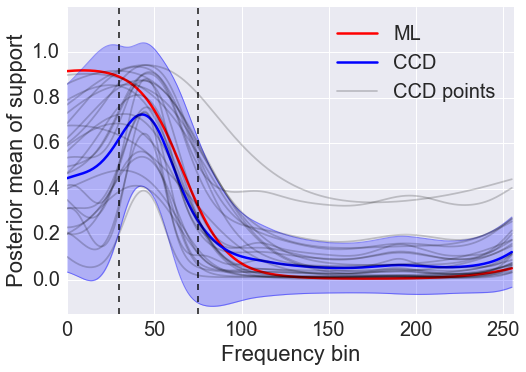}}
\subfigure[Posterior signal]{\includegraphics[width=0.32\textwidth]{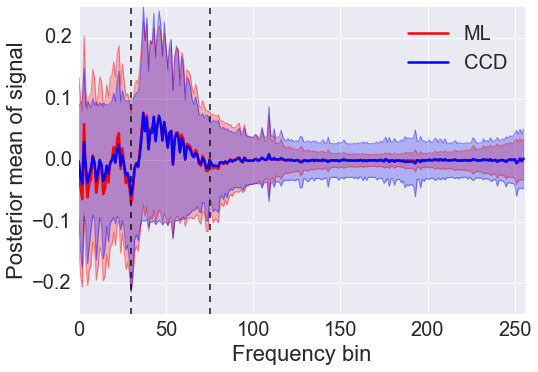}}
\caption{Comparison of the posterior distribution of $\bm{\gamma}$, $\bm{z}$ and $\bm{x}$ for ML and CCD hyperparameter inference. The posterior distribution for the CCD approximation is a mixture model and the thin black solid lines show the posterior mean of each individual component. These are omitted for the posterior of the signal to improve visual clarity. }
\label{fig:phoneme_comparison_of_prior}
\end{figure}

\setcounter{subfigure}{0}
In this experiment, we consider the task of phoneme recognition \citep{hastie01statisticallearning, miguel2011a}. In particular, we consider the problem of discriminating between the spoken vowels ''aa'' and ''ao'' using their log-periodograms as features. The data set consists of 695 and 1022 utterances of the vowels ''aa'' and ''ao'', respectively, along with their corresponding labels. The response variable in this experiment is binary and therefore we use the probit model rather than the Gaussian observation model.
\begin{table}[tp]
\centering
\begin{tabular}{lcccc}
\toprule
\textbf{Method} & \textbf{Training error} & \textbf{Test error} & \textbf{Train LPPD} & \textbf{Test LPPD}\\
\hline
NBSBC & $9.7$ ($\pm 0.3$) & $19.5$   ($\pm 0.1$) & $-42.7$ ($\pm 0.8$) & $-698.6$ ($\pm 3.3$) \\
LR-EP (ML) & $13.3$ ($\pm 0.3$) & $19.4$ ($\pm 0.1$) & $-50.6$ ($\pm 0.8$) & $-673.8$ ($\pm 2.2$) \\
LR-EP (CCD) & $13.4$ ($\pm 0.3$) &  $\mathbf{19.2}$ ($\pm 0.1$) & $-50.7$ $(\pm 0.8)$ & $\mathbf{-665.5}$ ($\pm 1.5$) \\
\bottomrule
\end{tabular}
\caption{Results for phoneme classification experiment}
\label{fig:phoneme_class}
\end{table}
\\
\\
Each log-periodogram has been sampled at 256 uniformly spaced frequencies. The left-most panel in Figure \ref{fig:phoneme_class} shows the frequency-wise mean and standard deviation of the two classes and the right-most panel shows the difference of the two mean signals. We choose a squared exponential kernel for $\bm{\gamma}$ since it is assumed that frequency bands rather than single frequencies are relevant for discriminating between the two classes. The total number of observations is $1717$ and we use $N = 150$ examples for training and the remaining $1567$ examples for testing. We repeat the experiment 100 times using different partitions into training and test sets. The training and test sets are generated such that the prior odds of the two classes are the same for both training and test. The number of input features is $D = 257$ (256 frequencies + bias). 
\\
\\
We use the low rank approximation for this experiment, and we choose the number of eigenvectors such that $99\%$ of the total variance in $\bm{\Sigma}_0$ is explained. We use the maximum likelihood method and the CCD marginalization method for the hyperparameter inference. We choose the prior mean of $\x$ as $\rho_0 = 0$ to reflect our ignorance on the sign of the active weights and we impose a half Student's $t$ distribution on the prior standard deviation of the $\x$, that is $\sqrt{\tau_0} \sim {t}^+(\text{df}=4)$, which is considered to be weakly informative.
\\
\\
As in the compressed sensing experiment, we impose a zero-mean normal distribution on the prior mean of $\bm{\gamma}$ and integrate it out analytically to obtain a kernel of the form given in eq. \eqref{eq:kernel_3params}. Compared to the compressed sensing example, our a priori knowledge of the structure of the support are more diffuse, but we expect that the lengthscale is significantly smaller than the number of frequency bins. Therefore, we choose a log-normal prior with mean 40 and standard deviation 30, that is $\kappa_3 \sim \mathcal{LN}\left(40, 30^2\right)$. The $5$'th and $95$'th percentiles for this distribution is approximately $10$ and $100$, respectively. For the remaining two hyperparameters, we use the same two prior distributions as in the previous experiment, that is $\kappa_1, \kappa_2 \sim \mathcal{LN}\left(4, 2^2\right)$. To predict the label of a new observation, we compute the predictive distribution by integrating the probit likelihood with respect to the approximate posterior distribution of the weights.
\\
\\
We compare our method against the network-based sparse Bayesian classification (NBSBC) method, which also uses EP to approximate the posterior distribution of linear model with coupled spike-and-slab priors. Instead of using a Gaussian process to encode the structure of the support, the NBSBC model encodes the structure in a network using a Markov random field prior. This method has been shown to outperform competing method on this specific problems \citep{miguel2011a}. 
\\
\\
Table \ref{fig:phoneme_class} summarizes the performance of the methods based on the average number of misclassifications and average log posterior predictive density (LPPD). It is seen that the LR-EP (ML) method achieves similar performance in terms test error as the NBSBC method, but it performs marginally better in terms of test LPPD. On the other hand, it is seen that the LR-EP (CCD) method outperforms both other methods. The panels in Figure \ref{fig:phoneme_comparison_of_prior} shows the posterior distributions of $\bm{\gamma}$, $\bm{z}$ and $\bm{x}$. The posterior distribution for the CCD approximation is a finite mixture model and each of the thin black lines shows the posterior mean for each individual mixture component. However, these are omitted for the posterior of $\x$ to improve the visual clarity of the figure. Based on Figure \ref{fig:phoneme_class}(b), we expect the weights for the frequencies between bin 35 and bin 70 to most discriminative of the two classes and it is seen that both the ML method and the CCD method have high posterior probabilities for the support in the region.

\setcounter{subfigure}{0}

\subsection{Experiment 6: Spatio-temporal Example}
\begin{figure}[tp] 
\centering
\subfigure[True $\z_0$]{\includegraphics[height = 3cm]{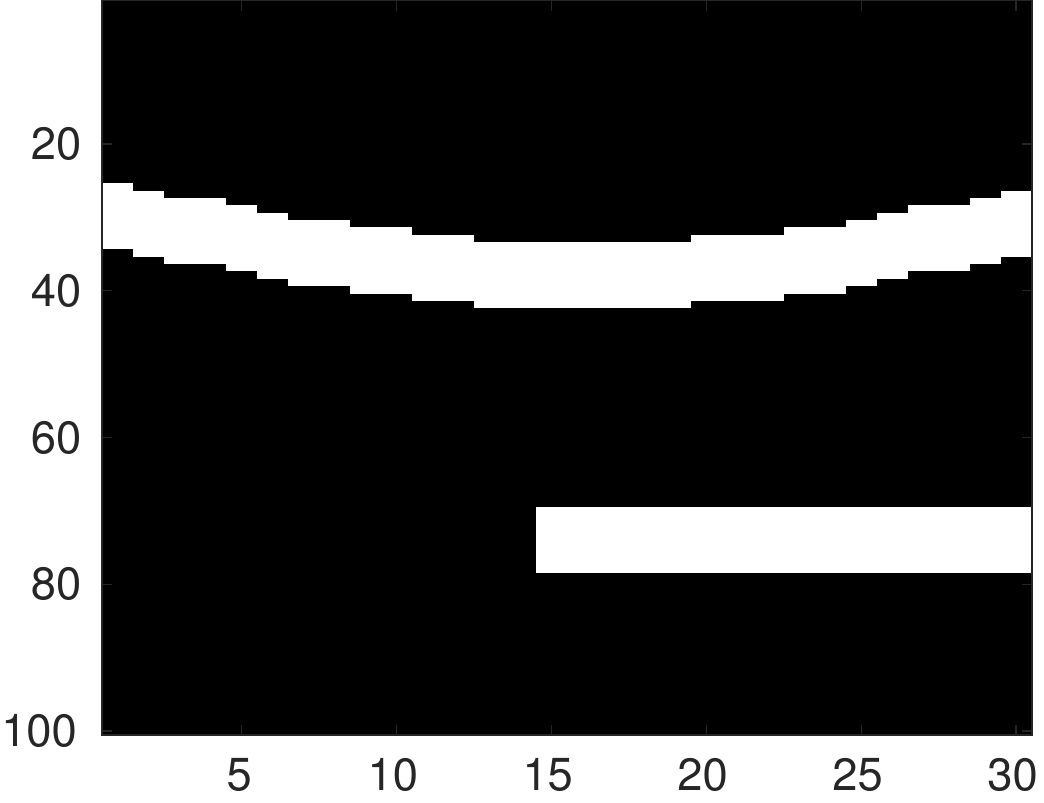}}
\subfigure[$\hat{\z}_{IEP}$]{\includegraphics[height = 3cm]{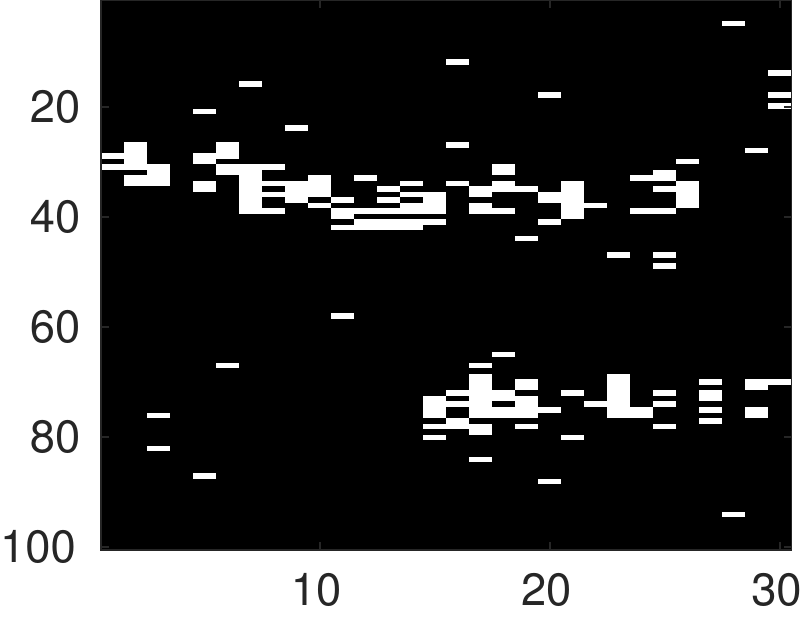}}
\subfigure[$\hat{\z}_{EP}$]{\includegraphics[height = 3cm]{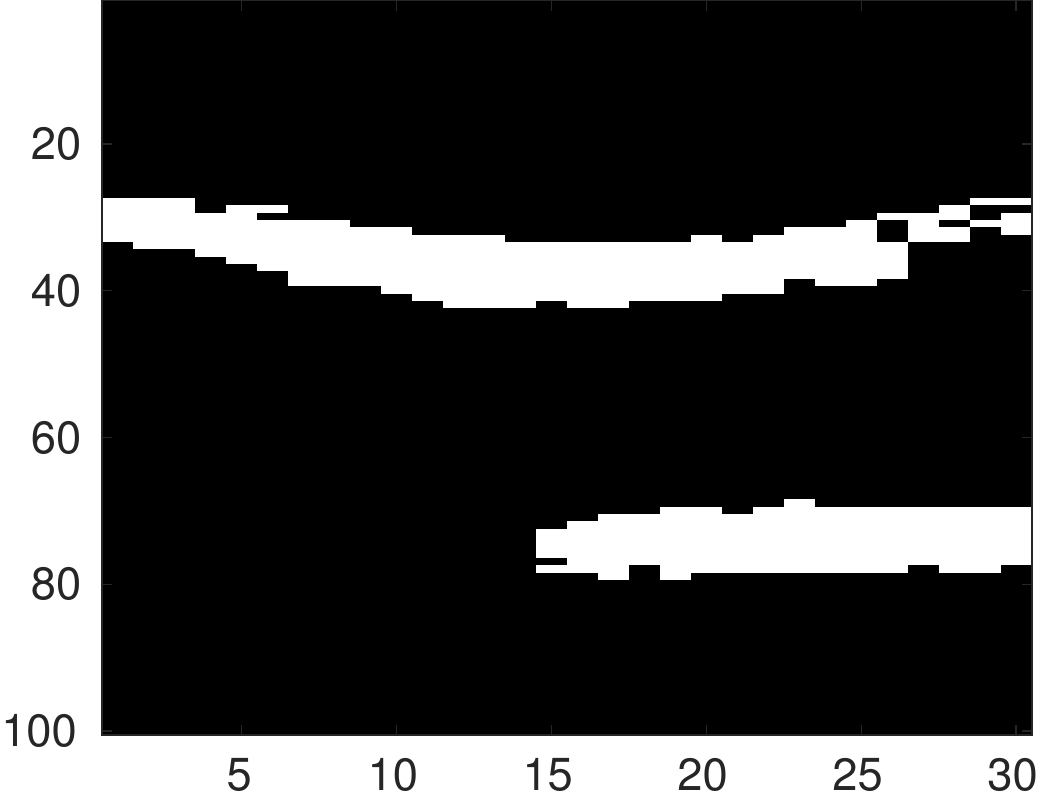}}\\
\subfigure[$\hat{\z}_{LR}$]{\includegraphics[height = 3cm]{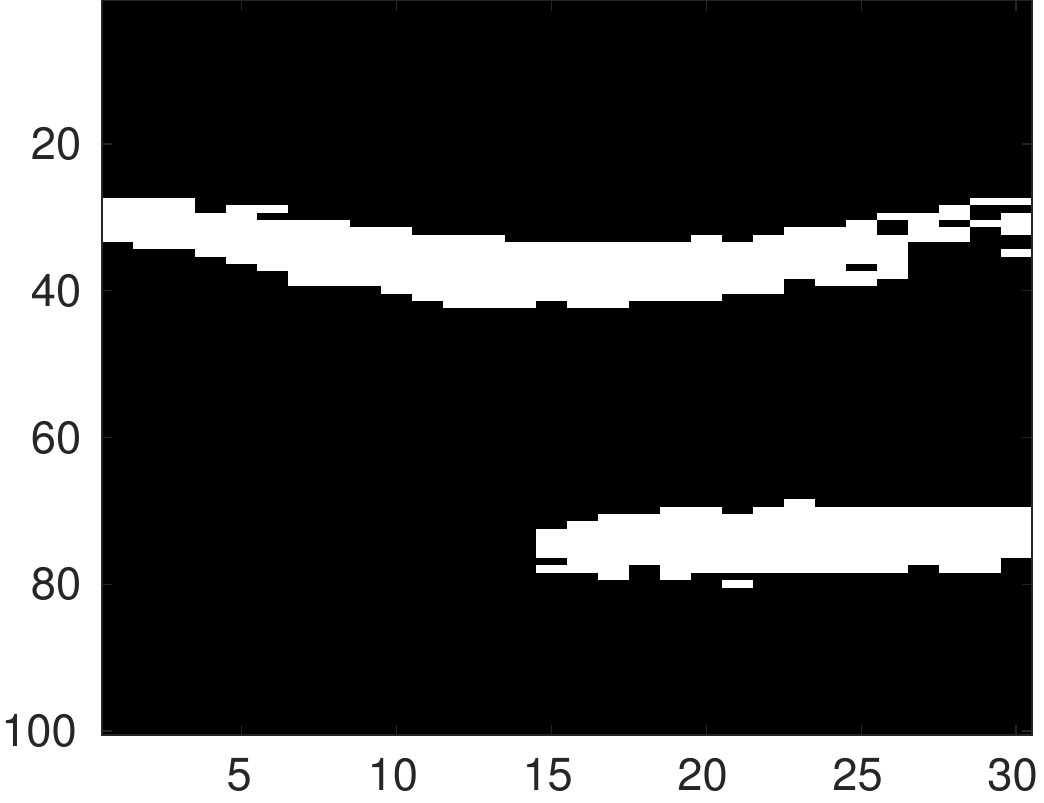}}
\subfigure[$\hat{\z}_{CP}$]{\includegraphics[height = 3cm]{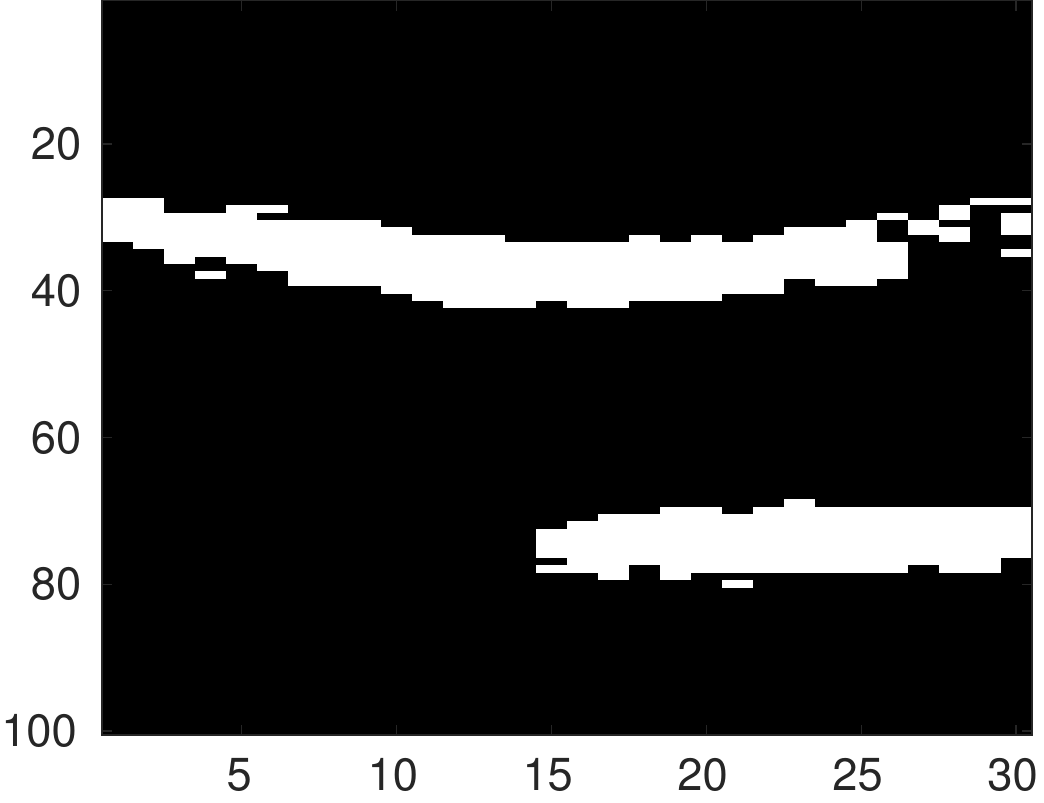}}
\subfigure[$\hat{\z}_{G}$]{\includegraphics[height = 3cm]{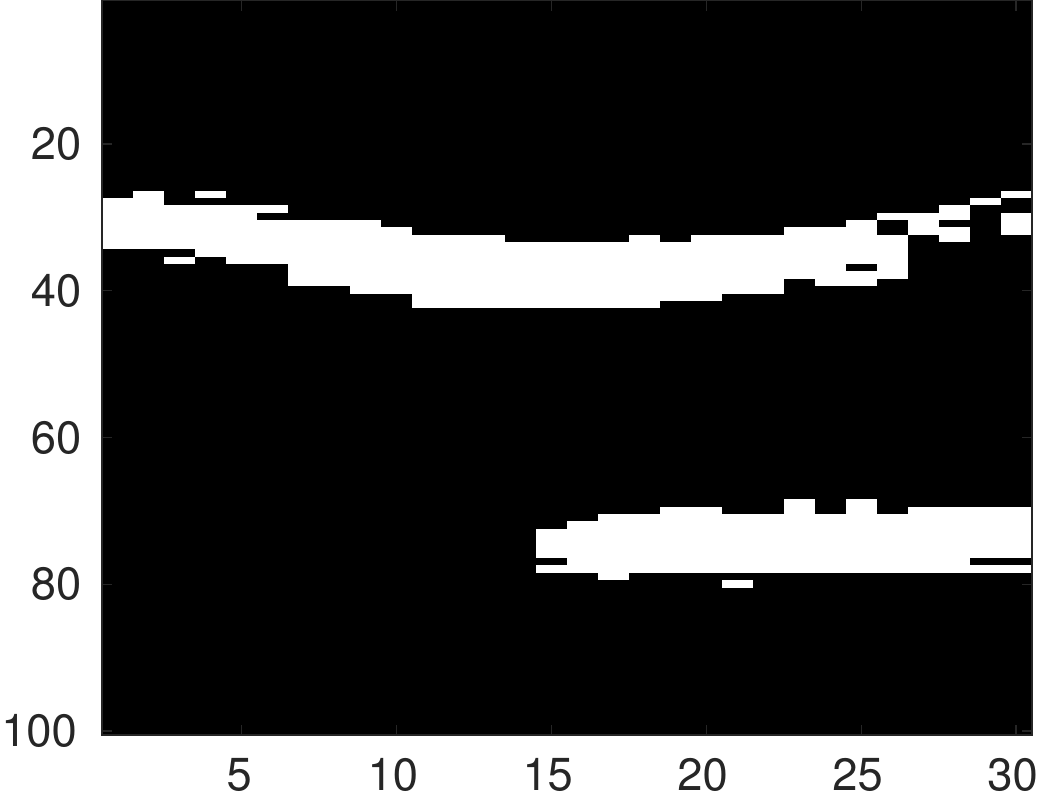}}\\
\subfigure[$EP$]{\includegraphics[width = 3cm]{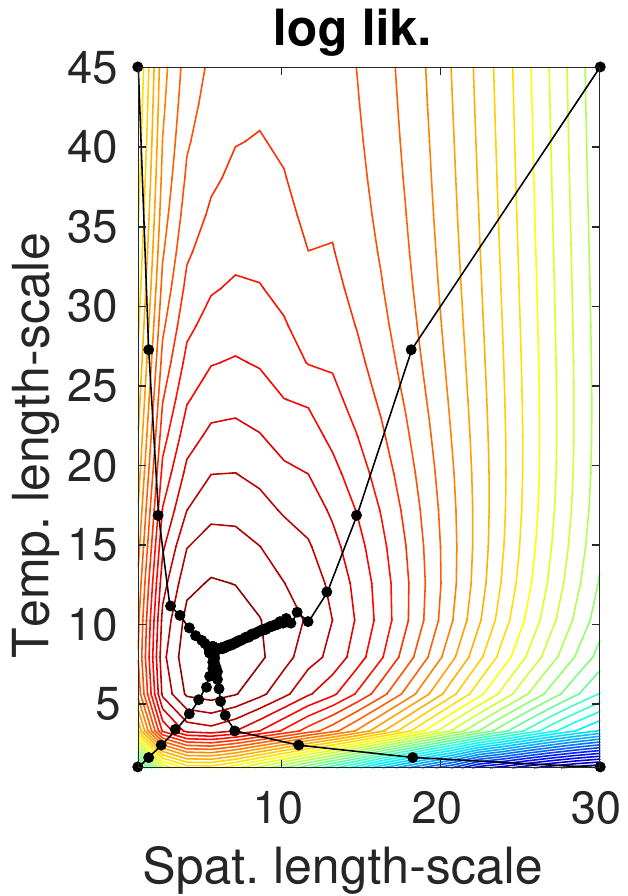}}
\subfigure[$LR$]{\includegraphics[width = 3cm]{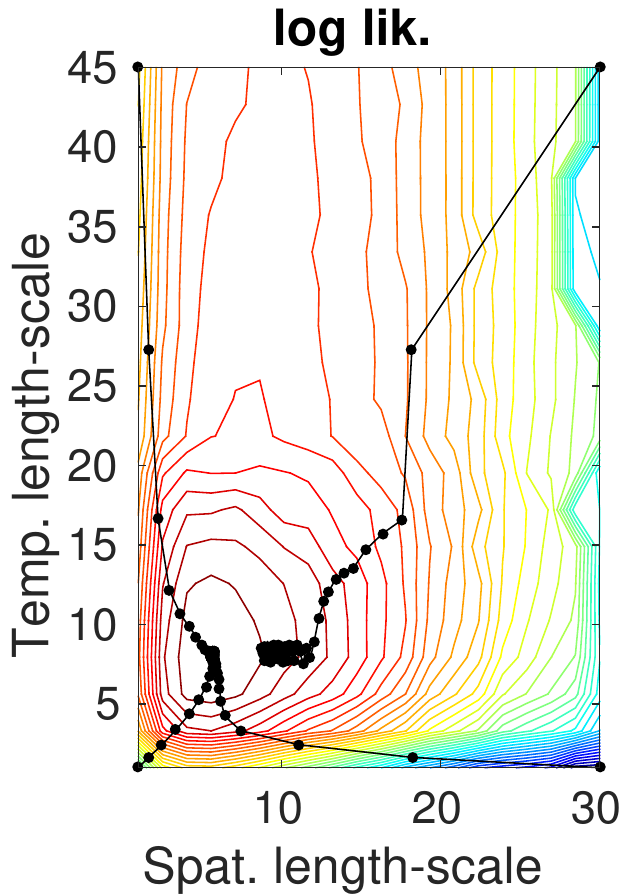}}
\subfigure[$CP$]{\includegraphics[width = 3cm]{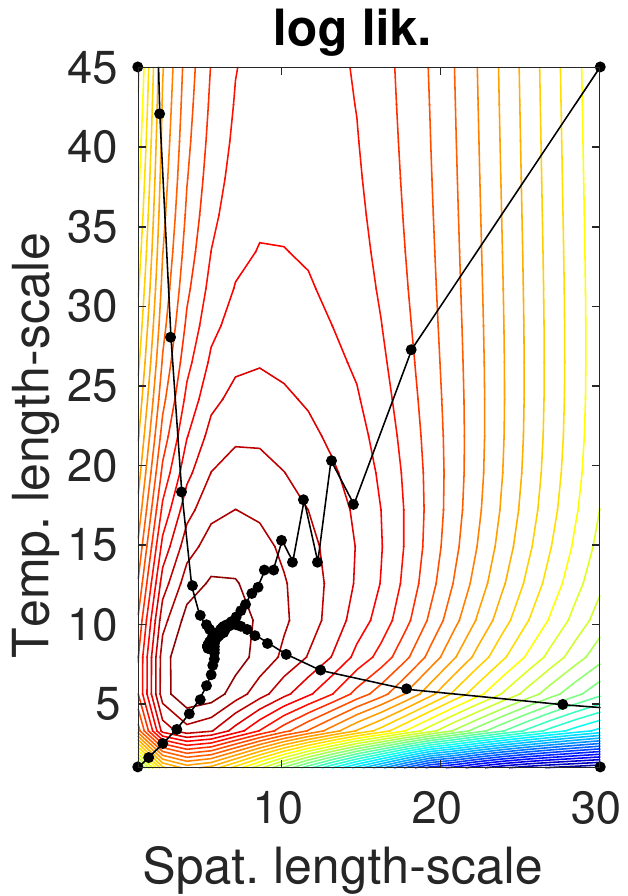}}
\subfigure[$G$]{\includegraphics[width = 3cm]{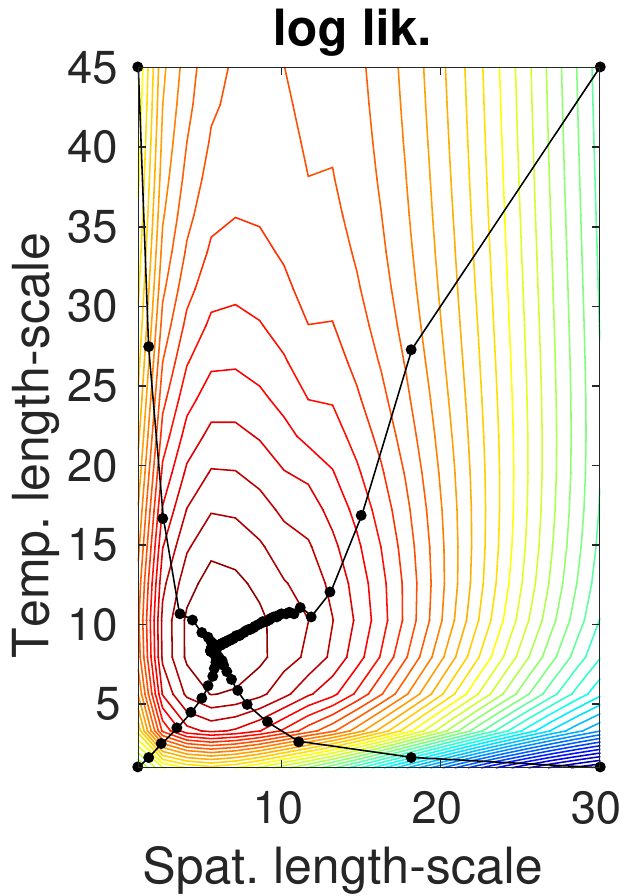}}
\subfigure[$EP$]{\includegraphics[width = 3cm]{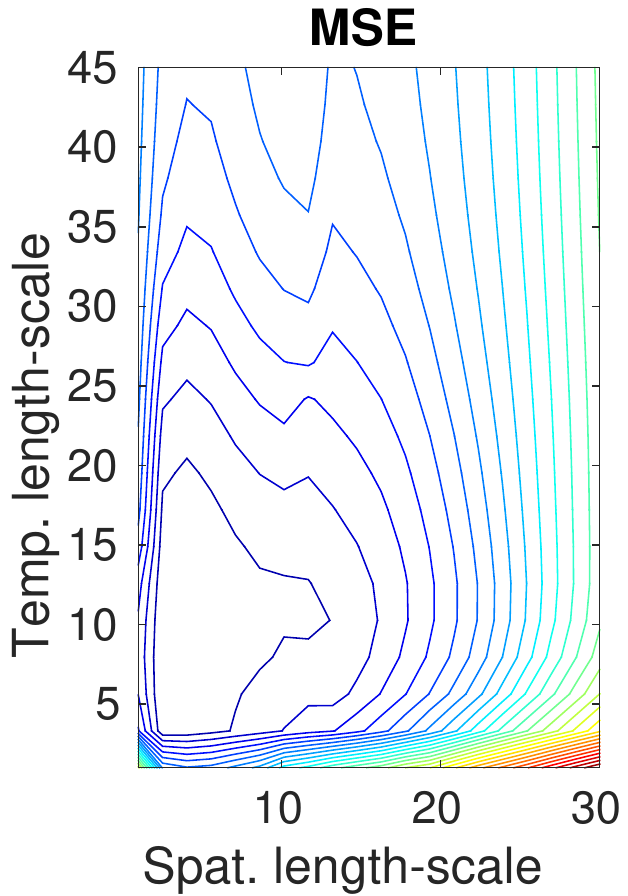}}
\subfigure[$LR$]{\includegraphics[width = 3cm]{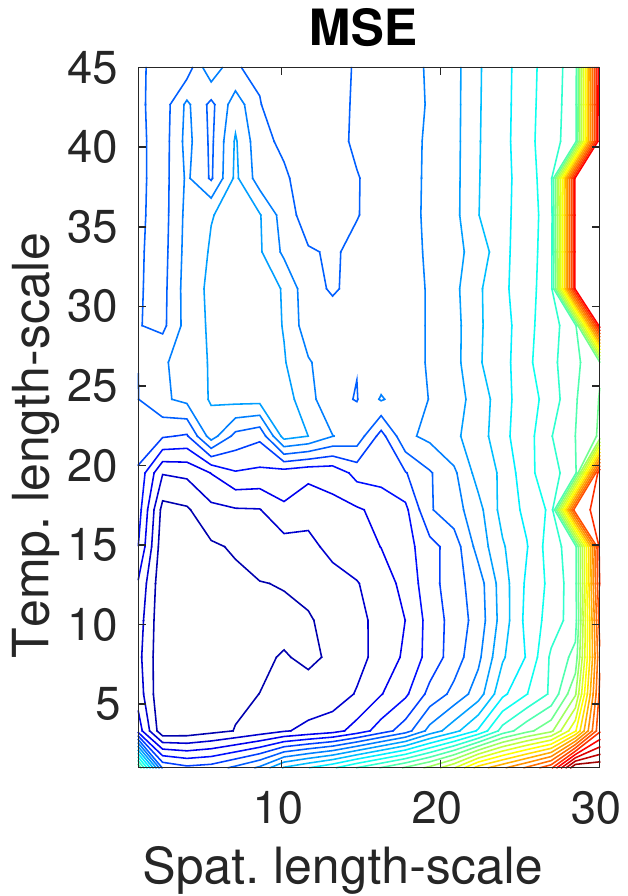}}
\subfigure[$CP$]{\includegraphics[width = 3cm]{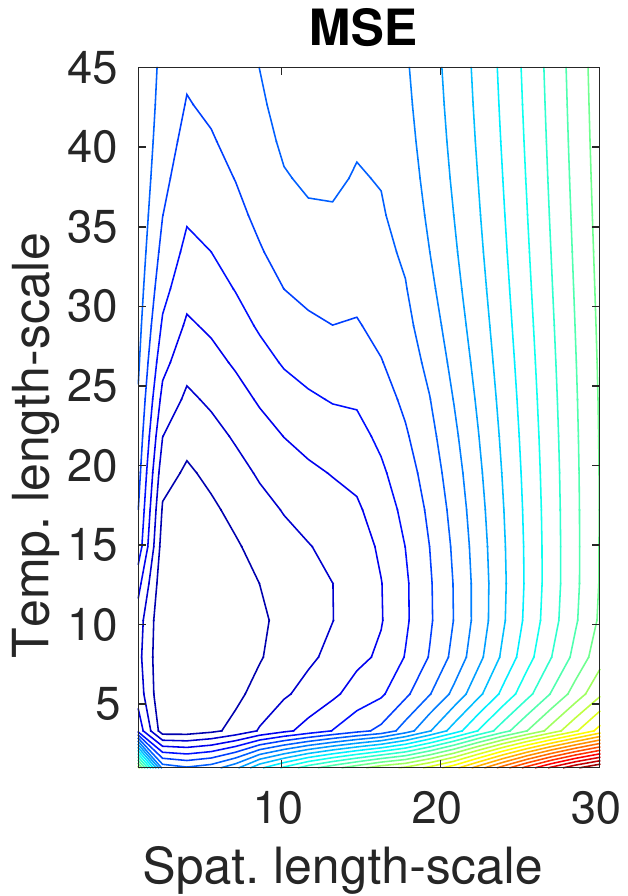}}
\subfigure[$G$]{\includegraphics[width = 3cm]{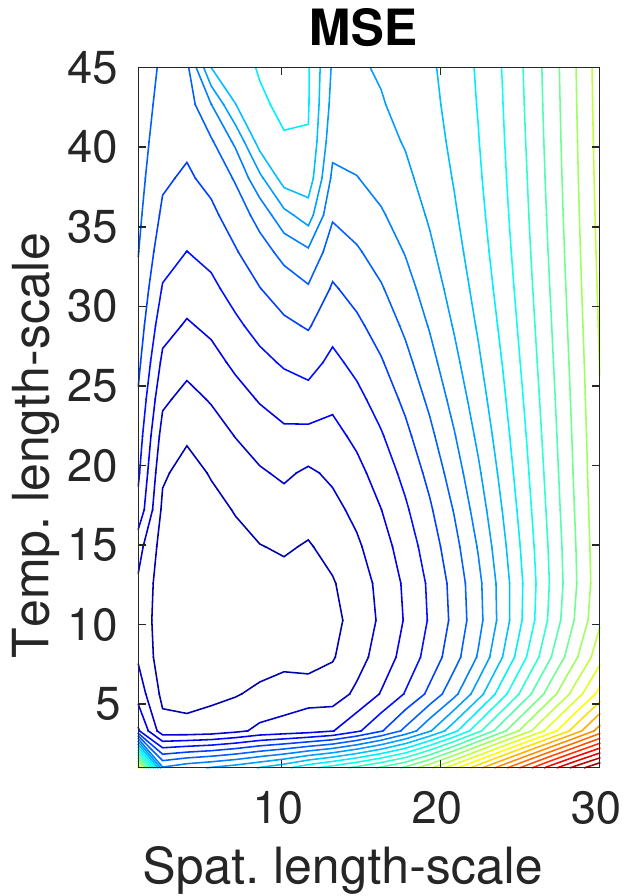}}
\caption{Results for a simulated spatio-temporal example with $D = 100$, $T = 30$, $N = 33$ and SNR $= 5dB$. Panels (a)--(f) compare the true sparsity pattern and the reconstructed sparsity patterns and panels (g)--(n) show the approximate marginal likelihood and the MSE error metric as a function of the spatial and temporal length-scale. For the low rank approximation (LR-EP), the number of eigenvectors is chosen to explain 99\% of the variance and the group size for group approximation (G-EP) is chosen to 5 and 10 in the spatial and temporal dimensions, respectively.}
\label{fig:exp5}
\end{figure}

In the previous experiments the focus was on problems with only one measurement vector, whereas in this and the following experiments we consider problems with multiple measurement vectors. Specifically, in this experiment we qualitatively study the properties of the proposed algorithm in the spatio-temporal setting using simulated data. We have synthesized a signal, where the support set satisfies the following three properties: 1) non-stationarity, 2) spatiotemporal correlation, and 3) the number of active coefficients change over time. The support of the signal is shown in panel (a) in Figure \ref{fig:exp5}. Based on the support set, we sample the active coefficients from a zero-mean isotropic Gaussian distribution. We then observe the desired signal $\X$ through linear measurements $\Y = \A\X + \bm{E}$, where both the forward model $\A$ and the noise $\bm{E}$ is sampled from a zero-mean isotropic Gaussian distribution. The noise variance is scaled such that the SNR is $5$dB. We apply our proposed method to estimate $\X$ given the forward model $\A$ and the observations $\Y$.
\\
\\
Panel (b) in Figure \ref{fig:exp5} shows the reconstructed support $\Z$ using the proposed EP algorithm with a diagonal prior covariance matrix on $\bm{\Gamma}$, which implies no prior correlation in the support. The panels (c)--(f) shows the reconstructed support for full EP, low rank EP, common precision EP and group EP, which all assumes that the prior covariance matrix for $\bm{\Gamma}$ is a Kronecker product of two squared exponential components. For the group approximation the group size is chosen to 5 and 10 in the spatial and temporal dimension, respectively, and for the low-rank approximation the rank is chosen such that the minimum number of eigenvectors explain $99\%$ of the variance in the prior. All hyperparameters are chosen by maximizing the approximate marginal likelihood. By inspecting the panels (a)--(f) it is seen that the reconstructed support is qualitatively improved by modeling the additional structure. Furthermore, the reconstructions using the approximation schemes do not differ significantly from the result using full EP. 
\\
\\
Panels (g)--(j) shows the marginal likelihood approximation as a function of the spatial and temporal length scale of the prior covariance matrix for the proposed methods, while the panels (k)--(g) show the corresponding NMSE between the reconstructed coefficients $\hat{\X}$ and the true coefficient $\X$. The black curves superimposed on the marginal likelihood plots show the trajectories of the optimization path for the length-scales of the prior distribution starting from four different initial values. It is seen that the marginal likelihood approximation is unimodal and correlates strongly with the NMSE surface, which suggests that it is reasonable to tune the length-scales of the prior covariance using the marginal likelihood approximation. However, we emphasize that this is not always the case and for some problems this indeed leads to suboptimal results. 


\subsection{Experiment 7: Phase Transitions for Multiple Measurement Vectors}
\begin{figure}[t]
\centering
\subfigure[Sample of $\bm{\Gamma}$]{\includegraphics[height = 3cm]{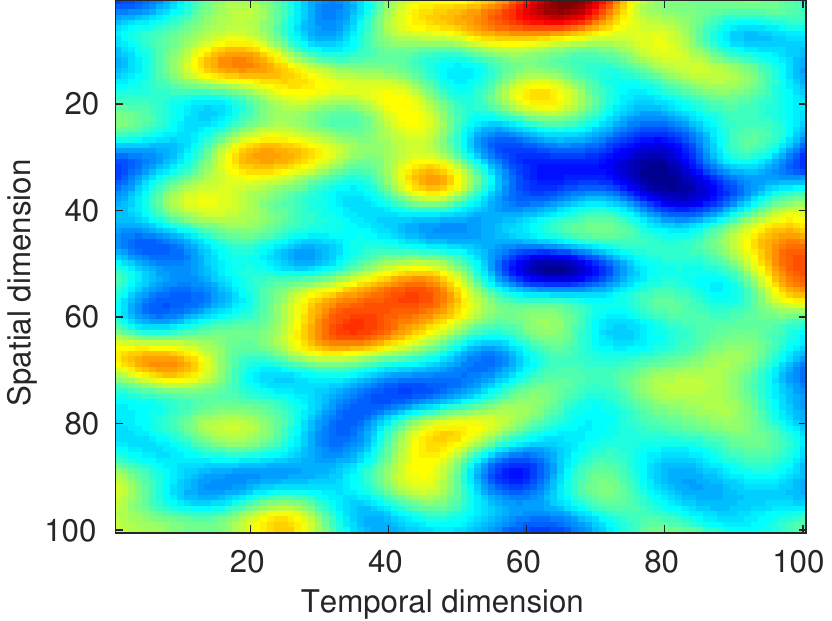}}
\subfigure[Sample of $\bm{Z}$]{\includegraphics[height = 3cm]{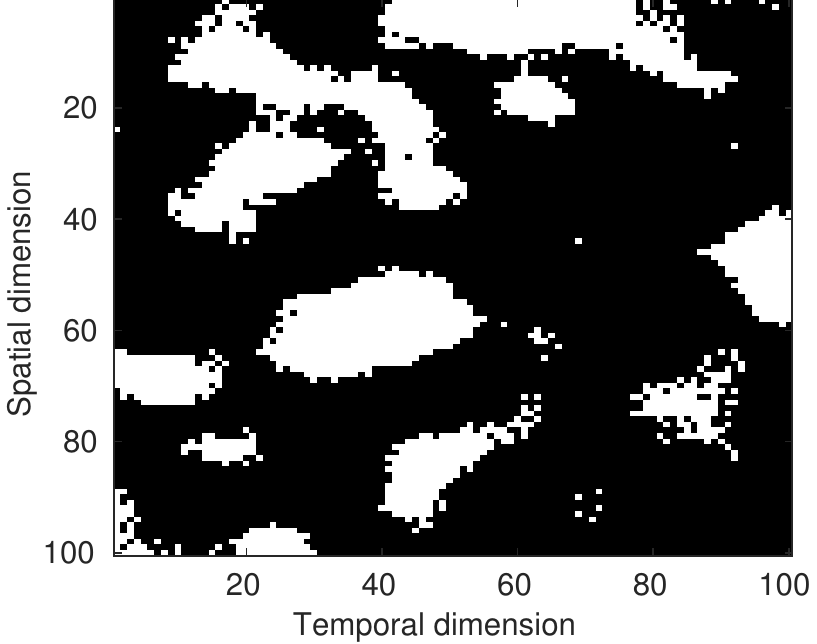}}
\subfigure[Sample of $\X$]{\includegraphics[height = 3cm]{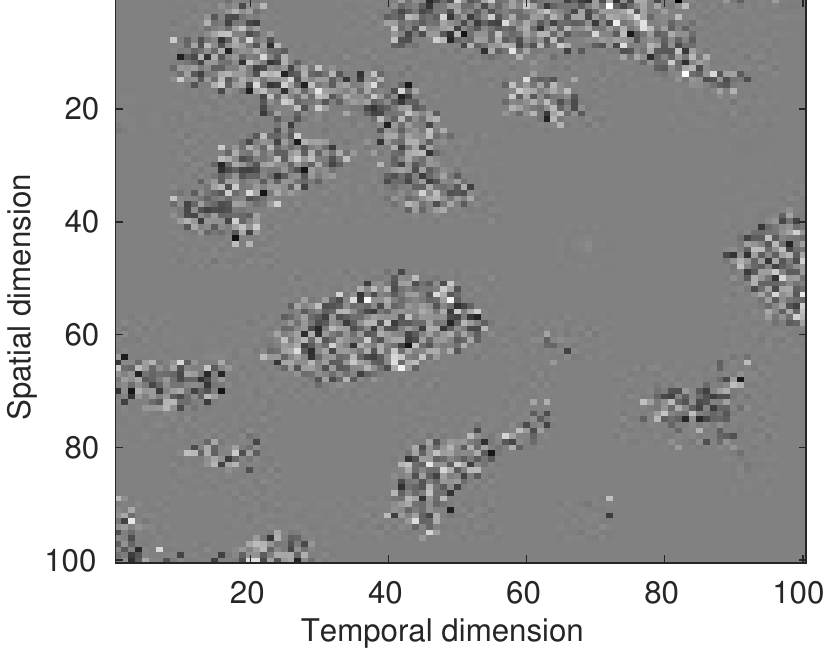}}
\caption{Example of a realization of the synthetic signals in Experiment 6. }
\label{fig:sample_experiment 6}
\end{figure}

\begin{figure}[thp]
\centering
\subfigure[NMSE]{\includegraphics[height = \figheight]{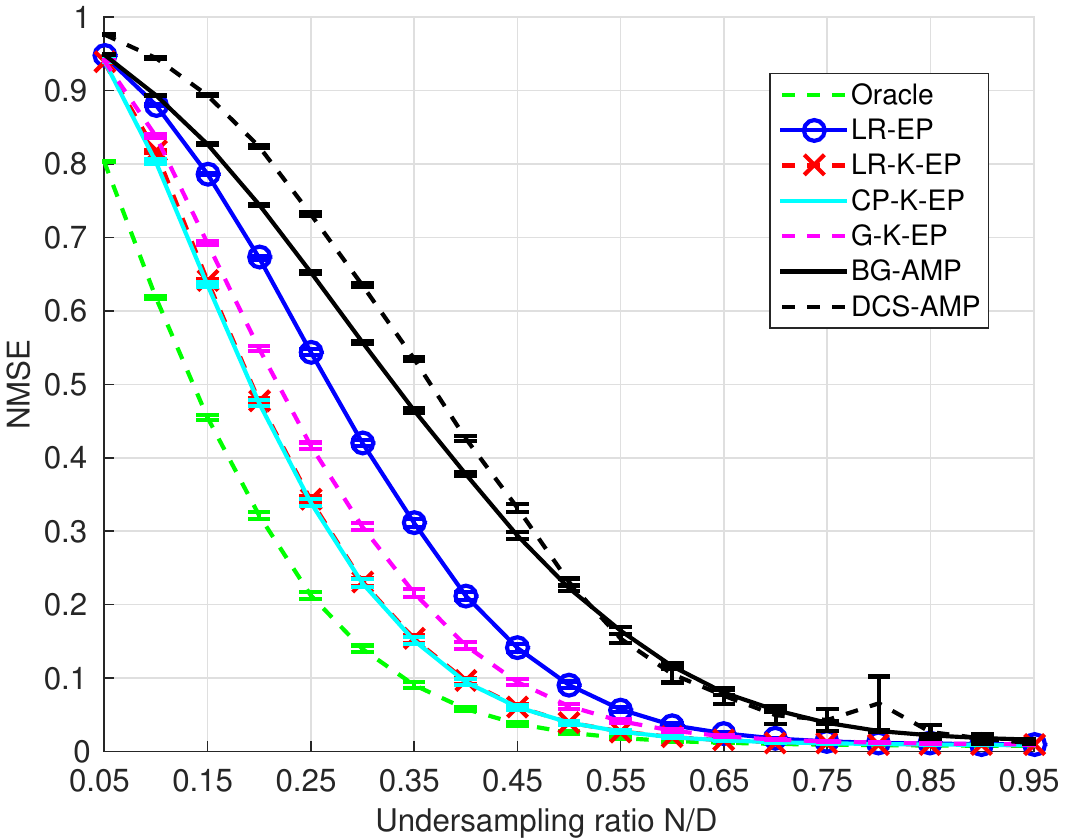}}
\subfigure[F-measure score]{\includegraphics[height = \figheight]{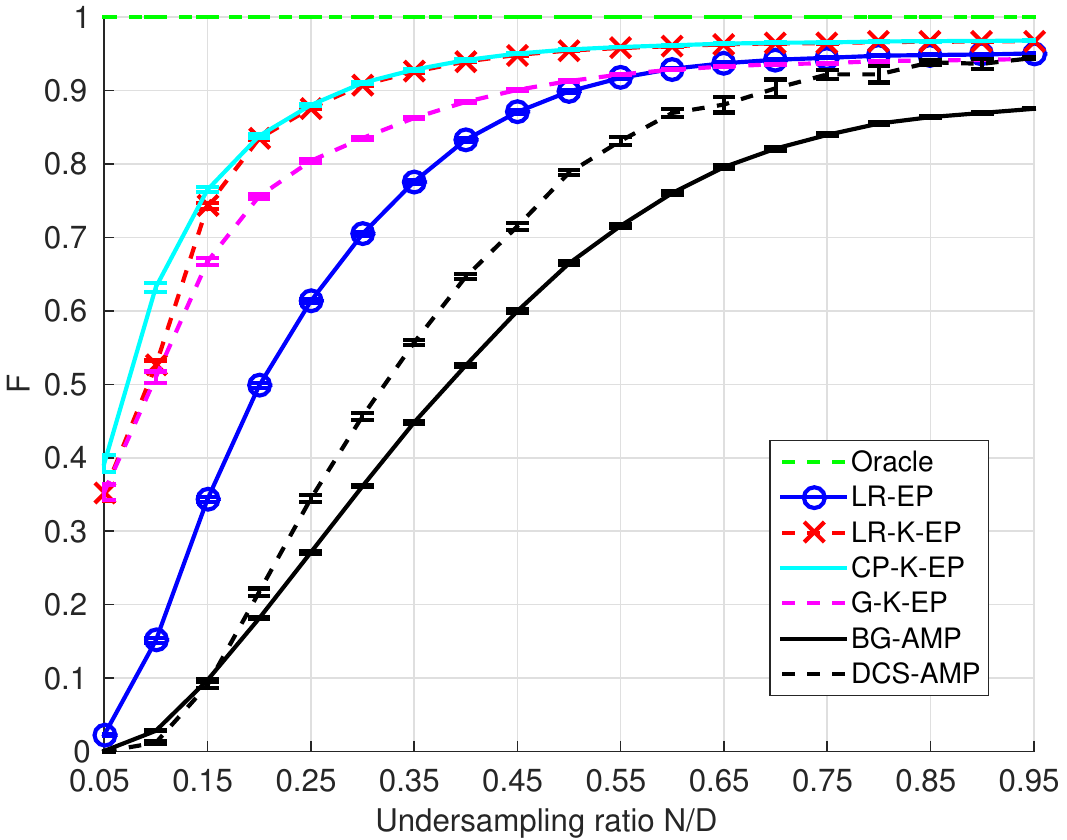}}
\subfigure[Iterations]{\includegraphics[height = \figheight]{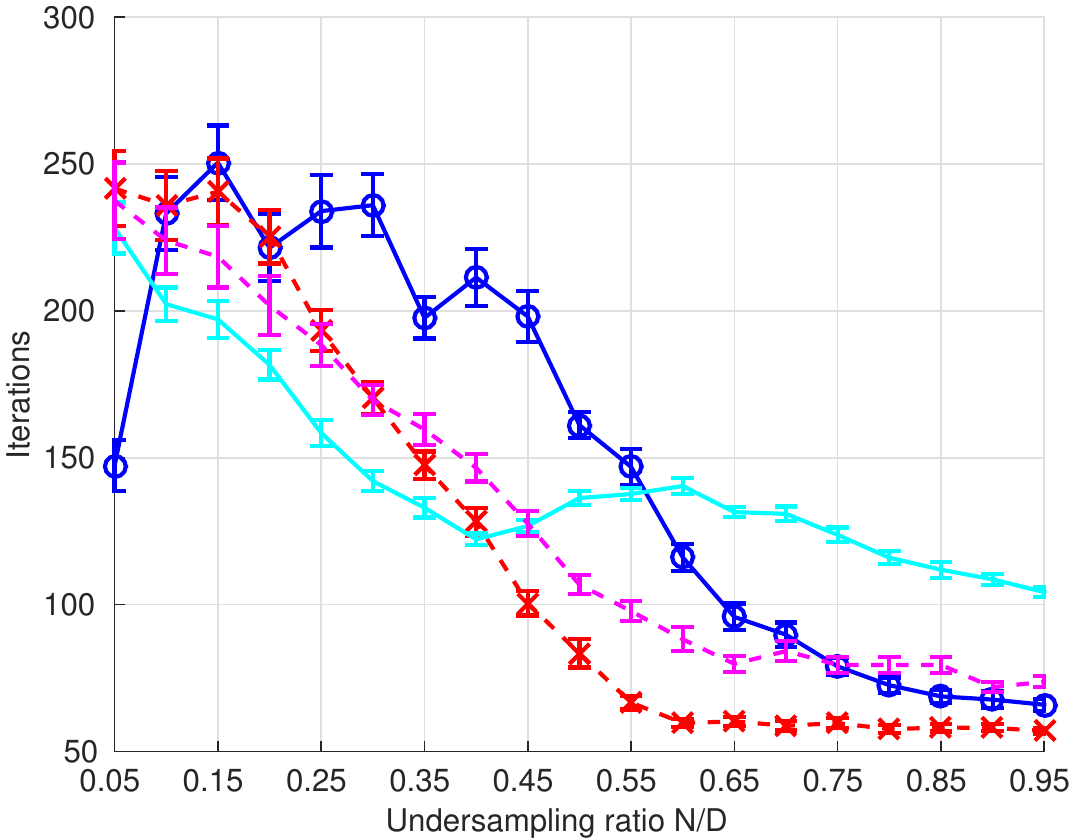}}
\subfigure[Time per iterations]{\includegraphics[height = \figheight]{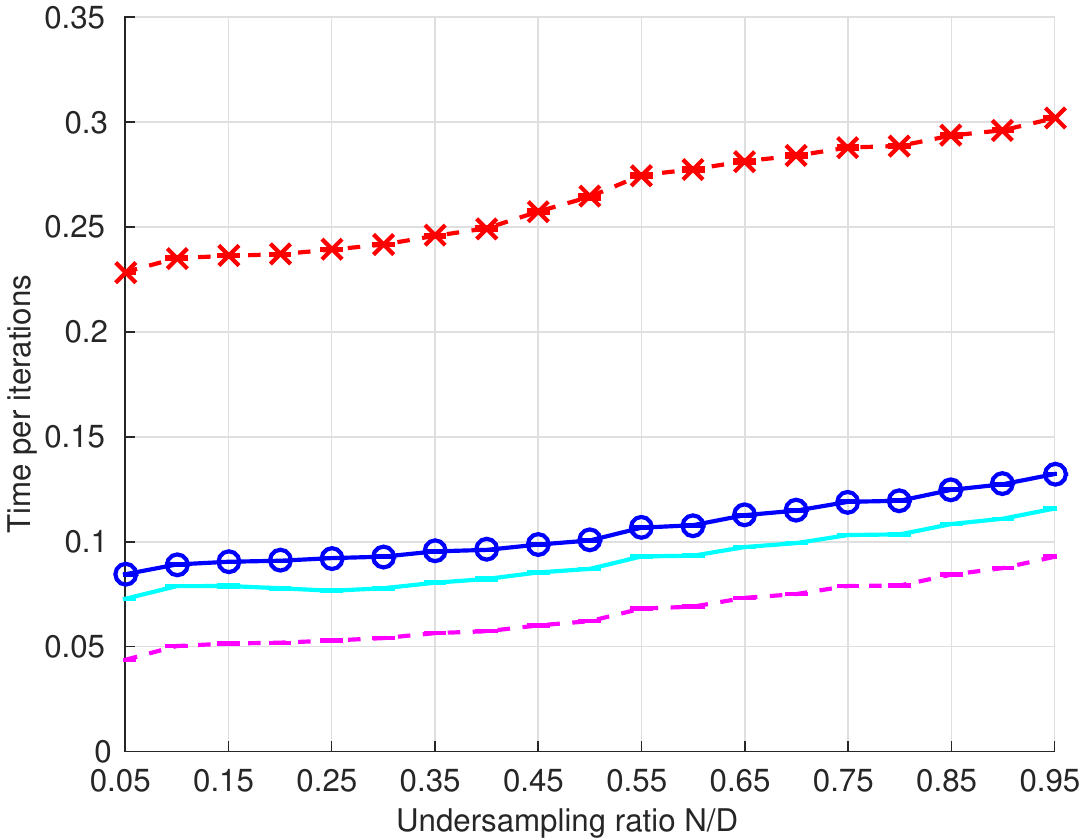}}
\caption{Performance of the methods as a function of undersampling ratio $N/D$ for $T = 100$. We compare low-rank EP with spatial structure only (LR-EP, Spatial), low rank EP spatio-temporal kronecker structure  (LR-L-EP), the common precision EP with spatio-temporal kronecker structure (CP-K-EP), group EP with spatio-temporal kronecker structure (G-K-EP), the low rank approximation (LR-EP) with BGAMP and DCSAMP. The results are averaged over 100 realization. }
\label{fig:experiment 6}
\end{figure}

The multiple measurement vector problem also exhibits a phase transition analogously to the single measurement vector problem described in Experiment 3 \citep{cotter2005sparse,ziniel2013a,Andersen2015}. In this experiment, we investigate how the location of the phase transition of the EP algorithms improves when the sparsity pattern of the underlying signal is smooth both in space and time and multiple measurement vectors are available. Using a similar setup as in Experiment 3, we generate 100 realizations of $\X$ from the prior specified in eq. \eqref{eq:kronecker_prior_start}--\eqref{eq:kronecker_prior_end} such that the total number of active components is fixed to $K = \frac{1}{4}DT = 2500$. The covariance structure is of the form $\bm{\Sigma}_0 = \bm{\Sigma}_{\text{temporal}} \otimes \bm{\Sigma}_{\text{spatial}}$, where both the temporal and spatial components are chosen to be squared exponential kernels. Figure \ref{fig:sample_experiment 6} shows an example of a sample realization of $\bm{\Gamma}, \bm{\Z}$ and $\X$ from the prior distribution. For each of the realizations of $\X$, we generate a set of linear observations $Y = \A\X + \bm{E}$, where the forward model $\A$ is Gaussian i.i.d. and $E_{nt} \sim \N\left(0, \sigma^2\right)$ is zero-mean Gaussian scaled such that the SNR is fixed to $20$dB. For reference we compare our methods against BG-AMP \citep{vila2013expectation} and DCS-AMP \citep{ziniel2013a}. The DCS-AMP method is a temporal extension to the BG-AMP method (see Experiment 3 for a brief description), and it uses approximate message passing inference based on spatially i.i.d. spike-and-slab priors, but assumes that the binary support variables evolve in time according to a first order Markov process. Both BG-AMP and DCS-AMP methods are informed about the true number of active coefficients and the true noise level. The results are shown in Figure \ref{fig:experiment 6}.
\\
\\
The method LR-EP (blue) assumes that the sparsity pattern is  spatially correlated, but independent in time. The method LR-K-EP (red, dashed) applies the low rank approximation to EP and assumes that the sparsity pattern is spatio-temporally correlated and that the prior covariance for $\bm{\Gamma}$ is described by a Kronecker product (hence the prefix ``-K''). Similarly, the methods CP-K-EP (cyan) and G-K-EP (magenta) have the same assumptions about the sparsity pattern, but use the common precision approximation and the group approximation, respectively. For G-K-EP we use groups of 5 in both the spatial dimension and temporal dimension. In this experiment, we do not run full EP with the spatiotemporal prior because it would be prohibitively slow. 
\\
\\
On panel (a) in Figure \ref{fig:experiment 6} it is seen that as the number of measurements increase, all methods eventually reach the NMSE level of the support-aware oracle estimator, but the general picture is that the more structure a method takes into account (i.i.d. sparsity vs. spatial sparsity vs. spatio-temporal sparsity), the better it performs in terms of NMSE. In particular, at $N/D \approx 0.3$ BG-AMP achieves NMSE $\approx 0.63$ and LR-EP achieves NMSE $\approx 0.44$ while LR-K-EP and CP-K-EP achieve NMSE $\approx 0.24$. Panel (b) shows a similar picture for F-measure. Furthermore, it is seen that the performance of LR-K-EP and CP-K-EP are similar and slightly better than the performance of G-K-EP both in terms of NMSE and F-measure. However, the G-K-EP approximation has the lowest computational complexity per iteration as seen in panel (d). In terms of run time the EP-methods are slower compared to the AMP-based methods, which have linear time complexity in all dimensions. However, the EP methods are not limited to Gaussian i.i.d. ensembles as the AMP-based methods are.

\subsection{Experiment 8: EEG Source Localization}
\begin{figure}[tpb]
\centering
\includegraphics[width = 0.5\textwidth]{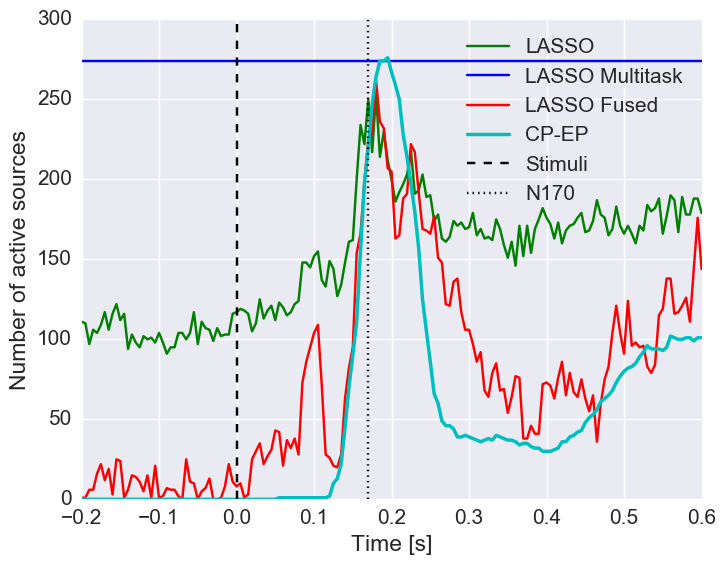}
\caption{The number of active dipoles sources as a function of the time for the common precision approximation (CP-EP) and the LASSO, the Multi-task LASSO and the fused LASSO for a face perception experiment. The stimuli were presented at time $t = 0$. }
\label{fig:eeg_1}
\end{figure}
In the final experiment, we apply the proposed method to an EEG source localization problem \citep{baillet2001a}, where the objective is to infer the locations of the active sources on the cortical surface of the human brain based on electroencephalogram (EEG) measurements. The brain is modelled using a discrete set of current dipoles distributed on the cortical surface and Maxwell's equations then describe how the magnitudes of the dipole sources relate to the EEG signals measured at the scalp.  We apply the proposed method to an EEG data set, where the subjects are presented with pictures of faces and scrambled faces. The data set is publicly available and the experimental paradigm is described in \citep{citeulike:6559653}. The data set has $N = 128$ electrodes and contains a total of 304 epochs evenly distributed between the two conditions: face or scrambled face. Each epoch has a duration of $800ms$ corresponding to $T = 161$ samples in time and the stimuli are presented at $t = 0s$. We generated a forward model\footnote{We used the SPM8 software \citep{SPM8(2010)}.} with 5124 dipole sources, that is $\A \in \mathbb{R}^{128 \times 5124}$. To encourage spatio-temporal coherence of the sources, we choose the covariance matrix for $\bm{\Gamma}$ to be of the form $\bm{\Sigma}_0 = \kappa^2\cdot \bm{\Sigma}_{\text{temporal}} \otimes \bm{\Sigma}_{\text{spatial}}$, where both the temporal component and spatial component are squared exponential kernels with individual length-scales. For simplicity, we use the Euclidean distance to compute the pairwise distances among the dipole sources as opposed to the more advances approach, where the distances are computed within the manifold defined by the cortex. 
\\
\\
The resulting inverse problem has $N = 128$ measurements, $T = 161$ measurement vectors and $D = 5124$ unknowns per time point and a total of $DT = 5124\cdot 151 = 824964$ unknowns. The forward model has a condition number of $\text{cond}\left(\A\right) = 3.1099\cdot10^{15}$. Thus, the problem instance is both heavily ill-posed and ill-conditioned. Because of the dimensions of this problem we use the common precision approximation for this data set. In fact, a low rank approximation of the prior covariance matrix $\bm{\Sigma}_0$ will require 3961 eigenvectors to explain $90\%$ of the variance and the matrix low rank eigenvector matrix $\bm{U}\in \mathbb{R}^{824964 \times 3691}$ would then require more than $20GB$ of memory to store in 64 bit double precision. 
\\
\\
\newcommand{\footnoteref}[1]{\textsuperscript{\ref{#1}}}
Tuning the hyperparameters using the approximate marginal likelihood estimate leads to poor solutions for this data set. In particular, the length-scales were significantly overestimated, which is consistent with what we observed in the compressed sensing experiment for very small samples sizes (see Figure \ref{fig:experiments_mnist_learning}(e)). However, manually specifying the hyperparameters using prior knowledge (spatial lengthscale $10mm$, temporal lengthscale $50ms$ and magnitude $\kappa^2 = 10$, prior mean 0), yields a posterior approximation with several interesting aspects. Ideally, we would compare the findings with the same posterior quantities for the BG-AMP and DCS-AMP methods as discussed earlier, but the highly correlated columns of the forward model make the AMP-approximations break down as they assume that the entries of the forward model are sampled from an Gaussian i.i.d. distribution. Instead, we compare with the LASSO\footnote{\label{footnote:sklearn}We used the implementation in scikit-learn toolbox \citep{scikit-learn}.} \citep{Tibshirani94regressionshrinkage}, the multi-task LASSO
\footnoteref{footnote:sklearn}
 \citep{Obozinski2006-dz} and the fused LASSO\footnote{We used the implementation in SPAMS toolbox \citep{Jenatton2010-gj,Mairal2010-kv}. } \citep{Tibshirani2005-wf}. The LASSO, the multi-task LASSO and the fused LASSO all minimize a quadratic reconstruction error subject to an $\ell_1$ constraint, but the multi-task LASSO also assumes that the sparsity pattern is constant in time (joint sparsity) and the fused LASSO has an additional constraint on the temporal first-order difference of the solution $\sum_{i,t}|x_{i,t} - x_{i,t-1}|$.

\begin{figure}[tpb]
\centering
\subfigure[CP-EP]{\includegraphics[width = 0.24\textwidth, trim = 4.5cm 2cm 4.5cm 1.5cm, clip = true]{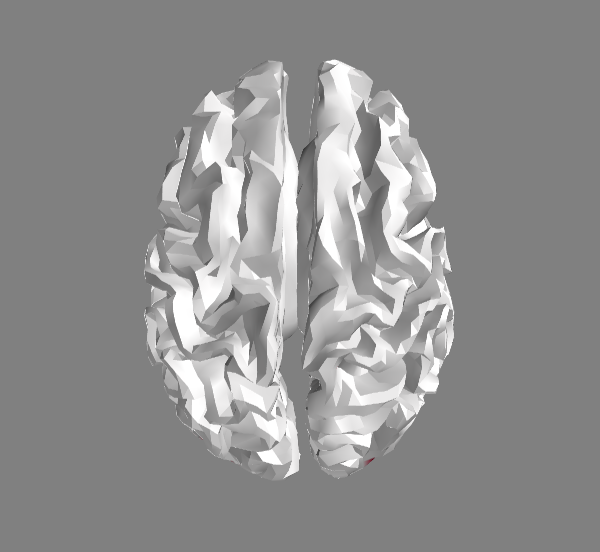}}
\subfigure[LASSO]{\includegraphics[width = 0.24\textwidth, trim = 4.5cm 2cm 4.5cm 1.5cm, clip = true]{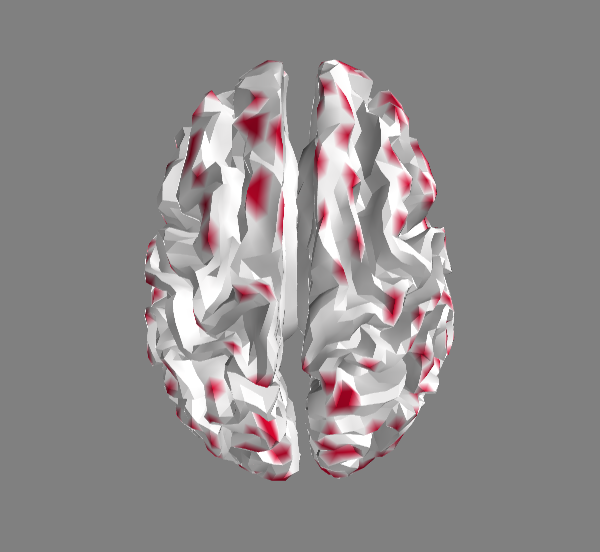}}
\subfigure[LASSO Multitask]{\includegraphics[width = 0.24\textwidth, trim = 4.5cm 2cm 4.5cm 1.5cm, clip = true]{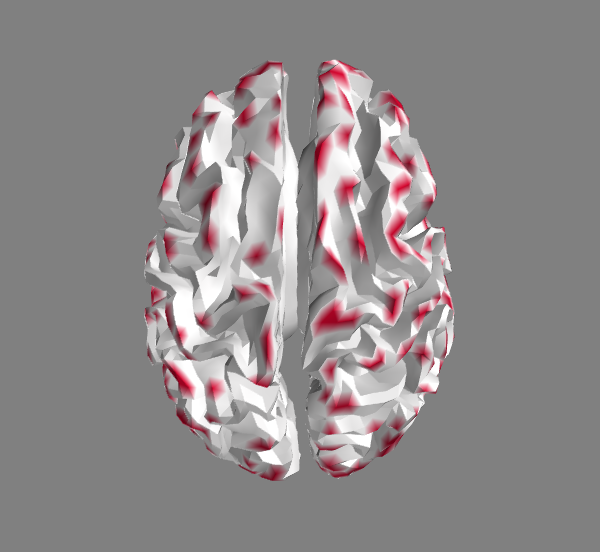}}
\subfigure[LASSO Fused]{\includegraphics[width = 0.24\textwidth, trim = 4.5cm 2cm 4.5cm 1.5cm, clip = true]{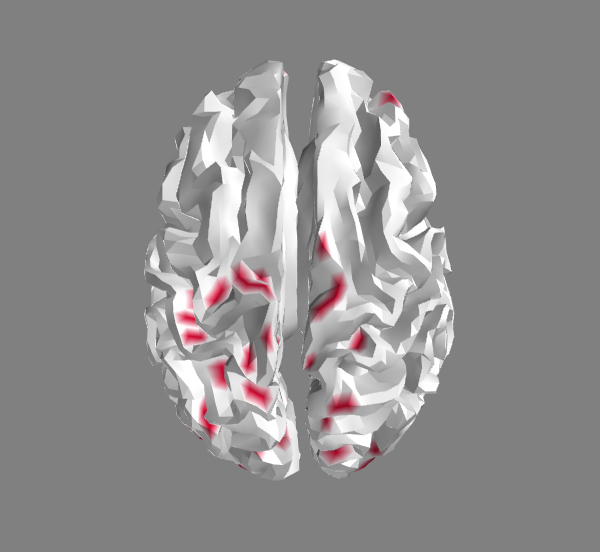}}\hfill\\
\subfigure[CP-EP]{\includegraphics[width = 0.24\textwidth, trim = 2.5cm 3cm 2.5cm 4cm, clip = true]{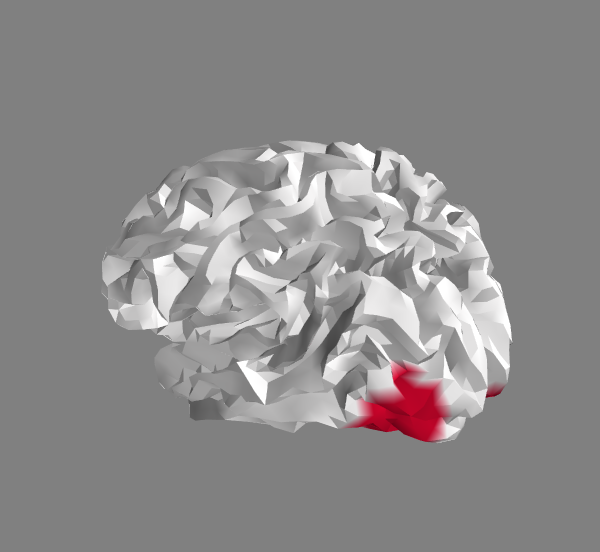}}
\subfigure[LASSO]{\includegraphics[width = 0.24\textwidth, trim = 2.5cm 3cm 2.5cm 4cm, clip = true]{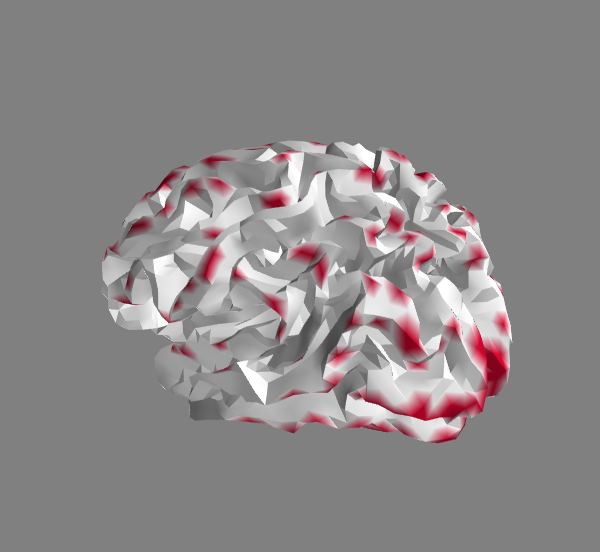}}
\subfigure[LASSO Multitask]{\includegraphics[width = 0.24\textwidth, trim = 2.5cm 3cm 2.5cm 4cm, clip = true]{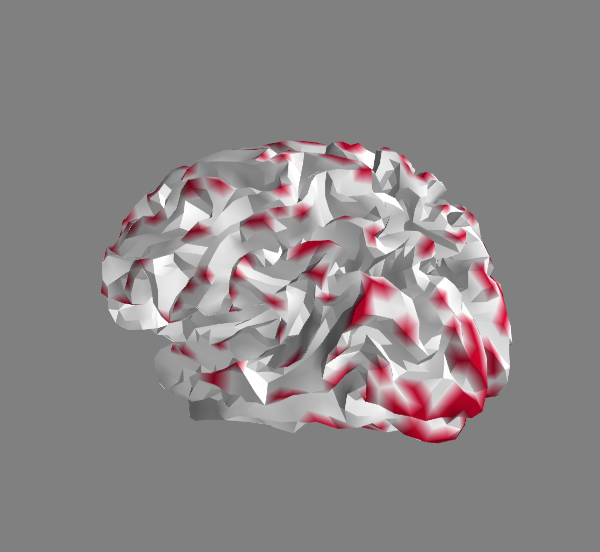}}
\subfigure[LASSO Fused]{\includegraphics[width = 0.24\textwidth, trim = 2.5cm 3cm 2.5cm 4cm, clip = true]{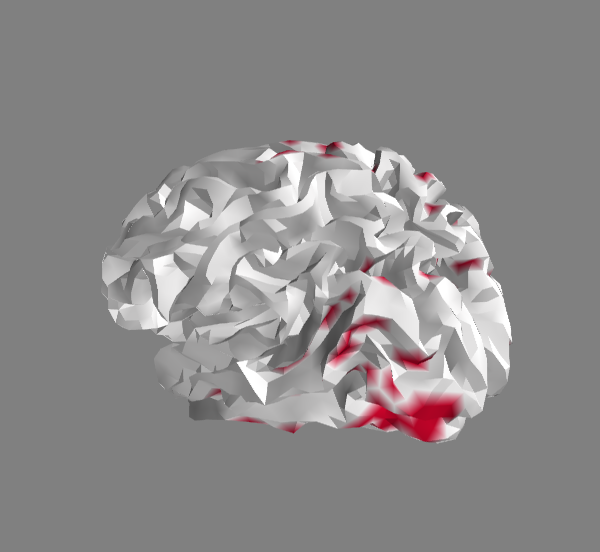}}\hfill\\
\subfigure[CP-EP]{\includegraphics[width = 0.24\textwidth, trim = 4cm 3cm 4cm 3cm, clip = true]{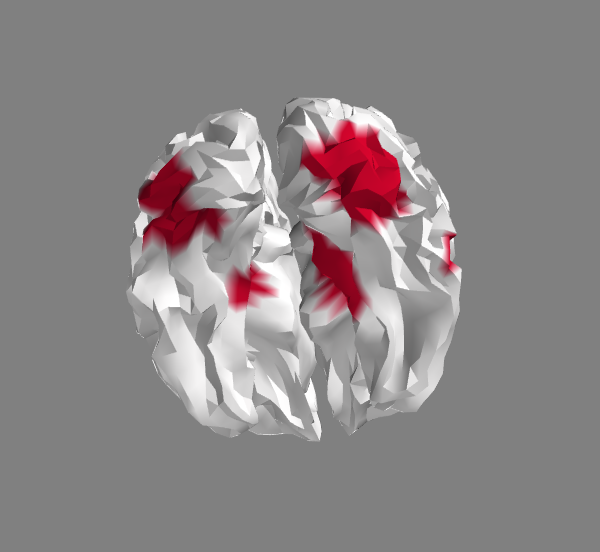}}
\subfigure[LASSO]{\includegraphics[width = 0.24\textwidth, trim = 4cm 3cm 4cm 3cm, clip = true]{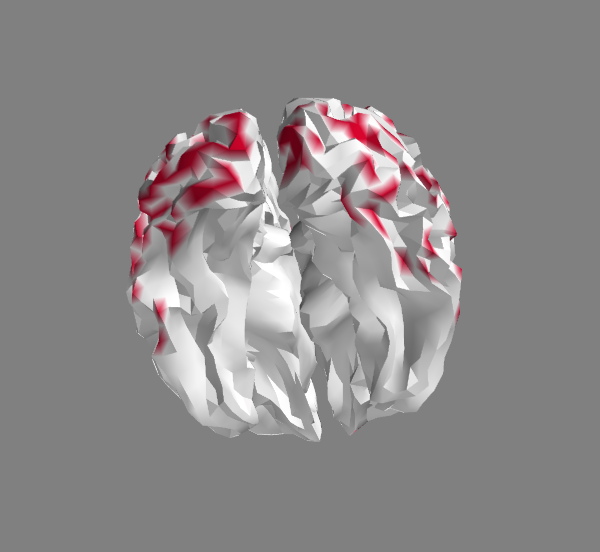}}
\subfigure[LASSO Multitask]{\includegraphics[width = 0.24\textwidth, trim = 4cm 3cm 4cm 3cm, clip = true]{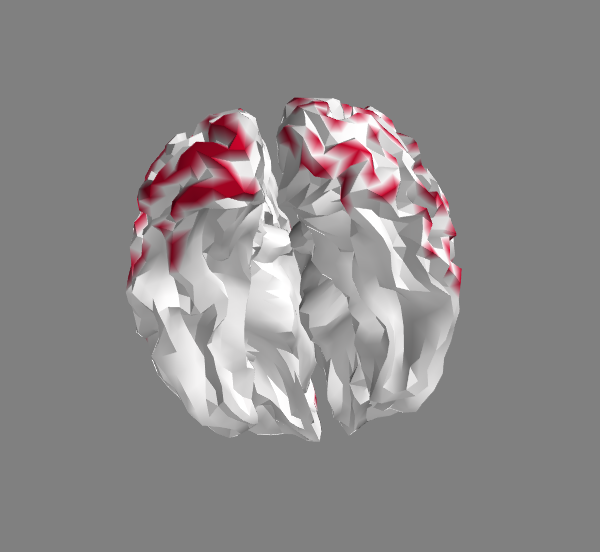}}
\subfigure[LASSO Fused]{\includegraphics[width = 0.24\textwidth, trim = 4cm 3cm 4cm 3cm, clip = true]{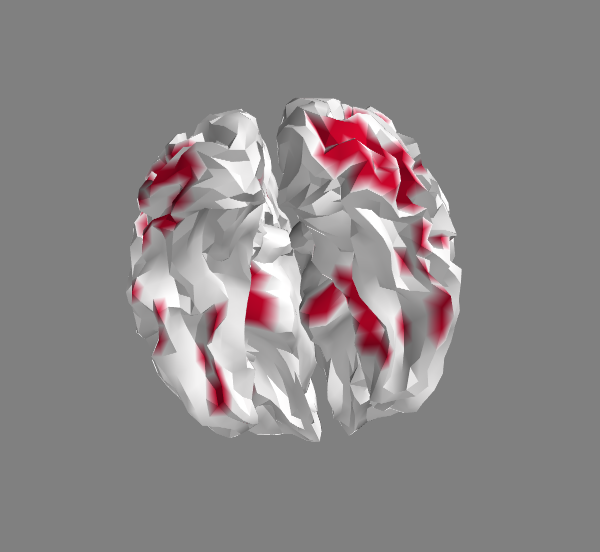}}\hfill\\
\caption{Estimate support sets for each method at time $t \approx 180ms$ for the face perception experiment. The top, middle and bottom rows show the brain from the top, side and bottom respectively. }
\label{fig:sources}
\end{figure}

The CP-EP method has been informed with appropriate values of the kernel hyperparameters. This is possible because the structure of the sparsity pattern is encoded using generic covariance functions that are easily interpretable. However, this is not an option for the LASSSO methods as the regularization parameters of these methods are more difficult to interpret in terms of the geometry of the problem. To compensate, we used the estimated solution obtained using the CP-EP method to inform the LASSO methods as follows. For the regular LASSO and the multi-task LASSO, we chose the value of the regularization parameter such that the number of active sources matches the CP-EP solution at the time point with the largest number of active sources (see Figure \ref{fig:eeg_1}). For the fused LASSO, we also matched the average autocorrelation across all sources.
\\
\\
Figure \ref{fig:eeg_1} shows the number of active dipole sources as a function of time for each method. The reconstructed support for both CP-EP and the fused LASSO are well-localized in time, whereas the distribution of active sources for LASSO are very diffuse in time. For the CP-EP method, it is seen that the number of active sources is zero until roughly time $t \approx 150ms$, where the number of active sources increase and peaks at $t \approx 180ms$, which is consistent with the known time delay of approximately 170ms for the face perception, that is the so-called N170 ERP component \citep{Itier2004-bc}. Figure \ref{fig:sources} shows a visualization of the estimated sets of active sources for time $t = 180ms$ from a top view, a side view and a bottom view, respectively. Interestingly, CP-EP detects four spatially coherent areas: left and right occipital and fusiform face areas that are associated with the face perception \citep{journals/neuroimage/HensonMF09}. The LASSO, the Multi-task LASSO and the fused LASSO also detect several active dipoles in the left and right occipital areas, but they also detect active sources distributed over the entire cortex as seen in the top row. 
\\
\\
Thus, from this experiment we conclude that this problem is too ill-posed for learning the hyperparameters of the model, but we can still extract meaningful information from the data using the model if we have access to additional a priori information. We note that learning hyperparameters is often difficult in neuroimaging due to high-dimensional signals and poor signal to noise conditions \citep{Varoquaux2017-ax}.

\FloatBarrier

\section{Summary and Outlook}

In this work, we have addressed the problem of solving multiple underdetermined linear inverse problems subject to a sparsity constraint. We have proposed a new generalization of the spike-and-slab prior distribution to encode a priori correlation of the support of the solution in both space and time by imposing a transformed Gaussian process on the spike-and-slab probabilities. An expectation propagation (EP) algorithm for posterior inference under the proposed model has been derived. Computations involved in EP updates scale like $\mathcal{O}\left(D^3T^3\right)$ where $D$ is the number of features and $T$ is number of inverse problems, hence for large scale problems, the standard EP algorithm can be prohibitively slow. We therefore introduced three different approximation schemes for the covariance structure to reduce the computational complexity. First, assuming that the prior has a Kronecker decomposition brings complexity $\mathcal{O}\left(D^3+T^3\right)$, based on this decomposition, a further $K$-rank approximation brings a reduction of complexity to $\mathcal{O}\left(K^2DT\right)$, we also proposed a common precision approximation of complexity $\mathcal{O}\left(D^2T + T^2D\right)$, and finally a scheme based on spatio-temporal grouping of variables effectively reducing $D$ and $T$ by the grouping factor.  We also discussed several ways to handle unknown hyperparameters, including maximum likelihood estimates, maximum a posterior estimates and efficient numerical integration using central composite design (CCD) approach.
\\
\\
We investigated the role of the spatio-temporal prior and the approximation schemes in a series of experiments. First we studied a simple 1D problem with spatial, translational invariant smoothness of the support (single measurement case, $T=1$). For a signal with two small connected components in the support, we illustrate the solutions for variable smoothness of the prior. For a wide range of prior parameters the correct form of the support is recovered, while the two support regions were found to merge in a single region as the smoothness length scale approaches the distance between the two regions. In the second experiment we investigated the role of the three covariance function approximations, also in a single measurement setup ($T=1$). We found that all approaches accurately reconstruct the true simulated support and inverse problem solutions despite the approximation of the Gaussian process posterior. It is well-known that the quality of the inferred solutions strongly depends on both the undersampling ratio and the sparsity level of the true solution. We investigated how the location of the phase transition is improved by invoking the smoothness prior. We found that the methods based on assumed prior correlation, were uniformly better than the methods with independent priors both in terms of the quality of normalized mean square error and in terms of their accuracy of support recovery (F-measure). The covariance approximation schemes are almost as fast as the scheme without smoothness, while yielding greatly improved performance. 
\\
\\
In the experiments 4 and 5, we investigated two applications: compressed sensing of numerical characters and phoneme recognition, respectively. In the former, we demonstrated how the quality of the reconstructed digits was improved using the structured prior. We also found that for the severely undersampled problems, maximum likelihood learning tends to overestimate the lengthscale of the kernel, which in turn lead to poor estimates of the support of and the weight. However, we also demonstrated how this could be alleviated by imposing a proper prior distribution to the lengthscale parameter and integrating over the uncertainty using CCD. In the latter experiment, we demonstrated how the probit likelihood can be used to extend to model binary sparse classification problems and we found that our algorithm compare well with published benchmarks.
\\
\\
In a sixth experiment we studied the properties of the proposed algorithms in the spatio-temporal setting using simulated data. Signals were synthesized so that the support set showed non-stationarity, spatio-temporal correlation and so that the cardinality of the support set changed over time. We estimated prior hyperparameters by optimizing the approximate marginal likelihood and found they converged to optimal settings in all cases. We found that there was a good correspondence between the approximate marginal likelihood and the solution's quantitative performance measure (NMSE).
\\
\\
Also for the multiple measurement vector problem it is known that there is a phase transition-like dependence of the solution quality on undersampling ratio. In the sixth experiment we investigated how the location of the phase transition of the EP algorithms improved when the sparsity pattern of the underlying signal is smooth in both space and time for the multiple measurements case. We compare our various approximate solvers with the state-of-the-art tools based on approximate message passing: BG-AMP, DCS-AMP, both of which were informed about the true number of active coefficients and the true noise level. The full EP was too demanding to run for this problem. Significant improvement were found for the methods that exploited sparsity structure. Comparing performance with AMP methods, the EP methods performed best both in terms of identifying the support (F measure) and in terms of NMSE. Run times for the EP-methods were longer compared to the AMP-based methods, which have linear complexity in all dimensions. We noted importantly that the EP methods also can be used for more general forward model ensembles (A), while the AMP-based methods assume a Gaussian i.i.d. ensemble.
\\
\\
In the final experiment, we applied the proposed methods to the hard problem of EEG source localization; data for this experiment was derived from a publicly available brain imaging data set designed to detect brain areas involved in face perception \citep{citeulike:6559653}. This was a larger scale application with $N = 128$ measurements and a total number of $824964$ unknowns, hence, only the common precision approximation was feasible. Furthermore, the forward model was very ill-conditioned in contrast to the well-conditioned i.i.d. ensembles considered in the simulations. For this data set, the hyperparameters of the kernel, for example, spatial and temporal lengthscale, could not be estimated from the data and thus, additional prior knowledge was required to perform inference.  In spite of these challenges highly interesting results were obtained: All four main foci of activation as earlier detected by fMRI, but not in these EEG data by other inference schemes, were here found to have well-defined and spatially extended support by the new approximate inference scheme. In contrast to fMRI EEG allowed us to monitor the dynamics in these areas in high temporal resolution.
\\
\\
This work has led to several interesting lines of research. First of all, from the the compressed sensing experiment as well as the source localization experiment, we concluded that the lengthscale parameter of the kernel cannot be learned from the data if the problem is too ill-posed. Thus, in future work we will extend the model to handle EEG data for multiple subjects simultaneously in a hierarchical manner, which allows us to use much more data to estimate the hyperparameters. Future studies also include an analysis of the phase transitions of the approximate  log marginal likelihood in the hyperparameter space of the spatiotemporal prior as discussed in Experiment 1. Furthermore, we also plan to apply the proposed algorithms to brain decoding problems, for example, in classification of fMRI task pattern data sets. Finally, we also plan to investigate the use of spatio-temporal sparsity priors for factor models like PCA and ICA. 
\FloatBarrier



\appendix
\section{Moments Computations for $f_{2,t,j}$}
\label{app:moments_for_f2tj}

In this section, we consider the update for the terms $\tilde{f}_{2,t,j}\left(x_{t,j}, z_{t,j}\right)$. First we compute the so-called cavity distribution $Q^{\backslash 2,t,j}\left(x_{t,j}, z_{t,j}\right)$ by removing the contribution of $f_{2,t,j}(x_{t,j}, {t,j})$ from the marginals of the joint approximation $Q\left(\x, \z, \bm{\gamma}\right)$
\begin{align}
Q^{\backslash 2,t,j}(x_{t,j}, z_{t,j}) &= \frac{Q\left(x_{t,j}, z_{t,j}\right)}{\tilde{f}_{2,t,j}\left(x_{t,j}, z_{t,j}\right)} = \frac{\N\left(x_{t,j}\big|\hat{m}_{t,j}, \hat{V}_{t,j}\right)\Ber\left(z_{t,j}\big|\phi\left(\hat{\gamma}_{t,j}\right)\right)}{\N\left(x_{t,j}\big|\hat{m}_{2,t,j}, \hat{V}_{2,t,j}\right)\Ber\left(z_{t,j}\big|\phi\left(\hat{\gamma}_{2,t,j}\right)\right)}\nonumber\\
&= K^{\backslash 2,t,j}\cdot \N\left(x_{t,j}\big|\hat{m}^{\backslash 2,t,j}, \hat{V}^{\backslash 2,t,j}\right)\Ber\left(z_{t,j}\big|\phi\left(\hat{\gamma}^{\backslash 2,{t,j}}\right)\right),
\end{align}
where
\begin{align}
\hat{v}^{\backslash 2, t, j} &= \left[ \hat{V}^{-1}_{jj} - \hat{v}^{-1}_{2,t,j}  \right]^{-1},\\
\hat{m}^{\backslash 2, t, j} &= \hat{v}^{\backslash 2, t, j} \left[ \hat{V}^{-1}_{t,jj}\hat{m}_{t,j} - \hat{v}^{-1}_{2,t,j}\hat{m}_{2,t,j}  \right],\\
\hat{\gamma}^{\backslash 2, t, j}  &= \hat{\gamma}_{3,t,j}.
\end{align}
Note that the cavity parameter for $\gamma$ for $f_{2,t,j}$ is simply equal to $\hat{\gamma}_{3,t,j}$ (and vice versa) since $\hat{\gamma}_{2,t,j}$ and $\hat{\gamma}_{3,t,j}$ are the only two terms contributing to $\gamma_{t,j}$.
\\
\\
Next, we minimize  the KL-divergence between $f_{2,t,j}Q^{\backslash 2,t,j}$ and $q$ or equivalently matching the moments between the two distributions. 
Following the latter approach we first compute the (unnormalized) moment w.r.t. $z_{t,j}$ 
\begin{align}
Z_1 &= \sum_{z_{t,j}}\int z_{t,j} f_{2,t,j}\left(x_{t,j}, z_{t,j}\right)Q^{\backslash 2, t,j}\left(x_{t,j}, z_{t,j}\right)\text{d} x_{t,j}\nonumber\\
&= \phi\left(\hat{\gamma}^{\backslash 2,t,j}\right)\N\left(0\big|\hat{m}^{\backslash 2,i} - \rho_0, \hat{V}^{\backslash 2, t,j}+\tau_0\right).
\end{align}
Next, the zeroth moment w.r.t $x_{t,i}$ or the normalization constant of $f_{2,t,j}Q^{\backslash 2,t,j}$
\begin{align}
X_0 &= \sum_{z_{t,j}}\int f_{2,t,j}\left(x_{t,j}, z_{t,j}\right)Q^{\backslash 2, t,j}\left(x_{t,j}, z_{t,j}\right)\text{d} x_{t,j}\\
&= \sum_{z_{t,j}}\int \left[\left(1-z_{t,j}\right)\delta(x_{t,j}) + z_{t,j}\N\left(x_{t,j}\big|\rho_0, \tau_0\right)\right]\nonumber\\
 &\quad\N\left(x_{t,j}\big|\hat{m}^{\backslash 2,{t,j}}, \hat{V}^{\backslash 2, t,j}\right)\Ber\left(z_{t,j}\big|\phi\left(\hat{\gamma}^{\backslash 2,{t,j}}\right)\right)\text{d} x_{t,j}\nonumber\\
&= \left(1-\phi\left(\hat{\gamma}^{\backslash 2,t,j}\right)\right) \N\left(0\big|\hat{m}^{\backslash 2,i}, \hat{V}^{\backslash 2, i}\right)+ \phi\left(\hat{\gamma}^{\backslash 2,t,j}\right)\N\left(0\big|\hat{m}^{\backslash 2,t,j} - \rho_0, \hat{V}^{\backslash 2, t,j}+\tau_0\right)\nonumber\\
&= \left(1-\phi\left(\hat{\gamma}^{\backslash 2,t,j}\right)\right) \N\left(0\big|\hat{m}^{\backslash 2,i}, \hat{V}^{\backslash 2, i}\right)+ Z_1
\end{align}
We now compute the (unnormalized) first moment w.r.t. $x_{t,j}$
\begin{align}
X_1 &= \sum_{z_{t,j}}\int x_{t,j} f_{2,t,j}\left(x_{t,j}, z_{t,j}\right)Q^{\backslash 2, i}\left(x_{t,j}, z_{t,j}\right)\text{d} x_{t,j}\nonumber\\
&= \phi\left(\hat{\gamma}^{\backslash 2,t,j}\right)\N\left(0\big|\hat{m}^{\backslash 2,t,j} - \rho_0, \hat{V}^{\backslash 2, t,j}+\tau_0\right)\frac{\frac{\hat{m}^{\backslash 2, t,j}}{\hat{V}^{\backslash 2, t,j}}+\frac{\rho_0}{\tau_0}}{\frac{1}{\tau_0} + \frac{1}{\hat{V}^{\backslash 2, t,j}}}\nonumber\\
&= Z_1\frac{\hat{m}^{\backslash 2, t,j}\tau_0+\rho_0\hat{V}^{\backslash 2, t,j}}{\tau_0 + \hat{V}^{\backslash 2, t,j}}
\end{align}
and the second (unnormalized) moment w.r.t. $x_{t,j}$
\begin{align}
X_2 &= \sum_{z_{t,j}}\int x_i^2 f_{2,i}\left(x_i, z_i\right)Q^{\backslash 2, i}\left(x_i, z_i\right)\text{d} x_i\nonumber\\
&= \phi\left(\hat{\gamma}^{\backslash 2,i}\right)\N\left(0\big|\hat{m}^{\backslash 2,i} - \rho_0, \hat{V}^{\backslash 2, i}+\tau_0\right)\left[\left(\frac{\frac{\hat{m}^{\backslash 2, i}}{\hat{V}^{\backslash 2, i}}+\frac{\rho_0}{\tau_0}}{\frac{1}{\tau_0} + \frac{1}{\hat{V}^{\backslash 2, i}}}\right)^2 + \frac{1}{\frac{1}{\tau_0} + \frac{1}{\hat{V}^{\backslash 2, i}}}\right]\nonumber\\
&= Z_1\left[\left(\frac{\hat{m}^{\backslash 2, t,j}\tau_0+\rho_0\hat{V}^{\backslash 2, t,j}}{\tau_0 + \hat{V}^{\backslash 2, t,j}}\right)^2 + \frac{\tau_0\hat{V}^{\backslash 2, i}}{\hat{V}^{\backslash 2, i} + \tau_0}\right]
\end{align}
The central moments for $Q^*$ in eq. \eqref{eq:kl_optim} are given by
\begin{align}
E\left[x_{t,j}\right] = \frac{X_1}{X_0}, && V\left[x_{t,j}\right] = \frac{X_2}{X_0} - \frac{X_1^2}{X_0^2}, &&
E\left[z_{t,j}\right] = \frac{Z}{X_0}.
\end{align}

\section{Moment Computations for $\tilde{f}_{3,t,j}$}
\label{app:moments_for_f3tj}
The moments matching for $\tilde{f}_{3,t,j}$ is derived in a similar manner as for $\tilde{f}_{2,t,j}$ (see appendix \ref{app:moments_for_f2tj} for details).  First we compute the cavity distribution $Q^{\backslash 3,t,j}\left(z_{t,j}, \gamma_{t,j}\right)$ by removing the contribution of $f_{3_t,j}(z_{t,j}, \gamma_{t,j})$ from the marginals of the joint approximation $Q$
\begin{align}
Q^{\backslash 3,t,j}(z_{t,j}, \gamma_{t,j}) &= \frac{Q\left(z_{t,j}, \gamma_{t,j}\right)}{\tilde{f}_{3,t,j}\left(z_{t,j}, \gamma_{t,j}\right)} = \frac{\Ber\left(z_{t,j}\big|\phi\left(\hat{\gamma}_{t,j}\right)\right)\N\left(\gamma_{t,j}, \hat{\mu}_{t,j}, \hat{\Sigma}_{t,jj}\right)}{\Ber\left(z_{t,j}\big|\phi\left(\hat{\gamma}_{3,t,j}\right)\right)\N\left(\gamma_{t,j}, \hat{\mu}_{3,t,j}, \hat{\Sigma}_{3,t,j}\right)}\nonumber\\
&= K^{\backslash 3,t,j}\cdot \Ber\left(z_{t,j}\big|\phi\left(\hat{\gamma}^{\backslash 3,t,j}\right)\right)\N\left(\gamma_{t,j}\big|\hat{\mu}^{\backslash 3,t,j}, \hat{\Sigma}^{\backslash 3,t,j}\right),
\end{align}
where
\begin{align}
\hat{\Sigma}^{\backslash 3,t,j} &= \left(\hat{\Sigma}_{t,jj}^{-1} - \Sigma_{3,t,j}^{-1}\right)^{-1},\\
\hat{\mu}^{\backslash 3,t,j} &=  \hat{\Sigma}^{\backslash 3,t,j}\left(\hat{\Sigma}_{t,jj}^{-1}\hat{\mu}_{t,j} -\hat{\Sigma}_{3,t,j}^{-1}\hat{\mu}_{3,t,j} \right),\\
\hat{\gamma}^{\backslash 3,t,j} &= \hat{\gamma}_{2,t,j}.
\end{align}
Once again we minimize the KL-divergence between $f_{3,t, j}Q^{\backslash 3,t,j}$ and $Q$ or equivalently matching the moments between the two distributions.  We now compute the moments w.r.t. $\gamma_{j,t}$ and $z_{j,t}$ of the (unnormalized) tilted distribution
\begin{align} \label{eq:inf_f3_int1_app}
G_m &= \sum_{z_{j,t}} \int \gamma^m_{j,t}\cdot f_{3,t,j}\left(z_{j,t}, \gamma_{j,t}\right) Q^{\backslash 3,t,j}\left(z_{j,t}, \gamma_{j,t}\right) \dd \gamma_{j,t} \quad\text{for}\quad m = 0,1,2,\\
Z_1 &= \sum_{z_{j,t}} \int z_{j,t}\cdot f_{3,t,j}\left(z_{j,t}, \gamma_{j,t}\right) Q^{\backslash 3,t,j}\left(z_{j,t}, \gamma_{j,t}\right) \dd \gamma_{j,t}\label{eq:inf_f3_int2_app}
\end{align}
We first compute the normalization constant of $f_{3,t,j}Q^{\backslash 3,t,j}$
\begin{align}
G_0 &= \sum_{z_{t,j}}\int f_{3,t,j}\left(z_{t,j}, \gamma_{t,j}\right)Q^{\backslash 3,t,j}\left(z_{t,j}, \gamma_i\right) \text{d} \gamma_{t,j}\nonumber\\
&= \sum_{z_{t,j}}\int \Ber\left(z_{t,j}\big|\phi\left(\gamma_{t,j}\right)\right)\Ber\left(z_{t,j}\big|\phi\left(\hat{\gamma}^{\backslash 3,t,j}\right)\right)\N\left(\gamma_{t,j}\big|\hat{\mu}^{\backslash 3,t,j}, \hat{\Sigma}^{\backslash 3,t,j}\right) \text{d} \gamma_{t,j}\nonumber\\
&= \sum_{z_i}\int \left[\left(1-z_i\right)\left(1 - \phi\left(\gamma_i\right)\right)\left(1 - \phi\left(\hat{\gamma}^{\backslash 3,i}\right)\right) + z_i\phi\left(\gamma_i\right)\phi\left(\hat{\gamma}^{\backslash 3,i}\right)\right]\N\left(\gamma_i\big|\hat{\mu}^{\backslash 3,i}, \hat{\Sigma}^{\backslash 3,i}\right) \text{d} \gamma_i\nonumber\\
&= \left(1 - \phi\left(\hat{\gamma}^{\backslash 3,i}\right)\right)\int \left(1 - \phi\left(\gamma_i\right)\right)\N\left(\gamma_i\big|\hat{\mu}^{\backslash 3,i}, \hat{\Sigma}^{\backslash 3,i}\right) \text{d} \gamma_i \nonumber\\
&\quad+ \phi\left(\hat{\gamma}^{\backslash 3,i}\right)\int \phi\left(\gamma_i\right)\N\left(\gamma_i\big|\hat{\mu}^{\backslash 3,i}, \hat{\Sigma}^{\backslash 3,i}\right) \text{d} \gamma_i\nonumber
\end{align}
Integrals of the form $\int \phi\left(\gamma_i\right)\N\left(\gamma_i\big|\hat{\mu}^{\backslash 3,i}, \hat{\Sigma}^{\backslash 3,i}\right) \text{d} \gamma_i$ can be solved analytically \citep{rasmussen2006a},
\begin{align}
\int \phi\left(\gamma_i\right)\N\left(\gamma_i\big|\hat{\mu}^{\backslash 3,i}, \hat{\Sigma}^{\backslash 3,i}\right) \text{d} \gamma_i &= \phi\left(c_{3,i}\right), \quad c_{3,i} \triangleq \frac{\hat{\mu}^{\backslash 3,i}}{\sqrt{1 + \hat{\Sigma}^{\backslash 3,i}}}.
\end{align}
Inserting this result back into the expression for $G_0$ yields
\begin{align}
G_0 &= \left(1 - \phi\left(\hat{\gamma}^{\backslash 3,i}\right)\right)\left(1 - \phi\left(c_{3,i}\right)\right) + \phi\left(\hat{\gamma}^{\backslash 3,i}\right)\phi\left(c_{3,i}\right).
\end{align}
We can now compute the moments of the unnormalized distribution
\begin{align}
Z_1 &= \sum_{z_i}\int z_i f_{3,i}\left(z_i, \gamma_i\right)Q^{\backslash 3,i}\left(z_i, \gamma_i\right) \text{d} \gamma_i\nonumber\\
&=\phi\left(\hat{\gamma}^{\backslash 3,i}\right)\phi\left(c_{3,i}\right),
\end{align}
Then the first moment w.r.t. to $z_{i,t}$ is obtained as $E\left[z_{i,t}\right] = Z_1/G_0$.
\\
\\
For the moments w.r.t. $\gamma_i$, we get
\begin{align}
G_1 &= \sum_{z_i}\int \gamma_i f_{3,i}\left(z_i, \gamma_i\right)Q^{\backslash 3,i}\left(z_i, \gamma_i\right) \text{d} \gamma_i\nonumber\\
&= \sum_{z_i}\int \gamma_i\left[\left(1-z_i\right)\left(1 - \phi\left(\gamma_i\right)\right)\left(1 - \phi\left(\hat{\gamma}^{\backslash 3,i}\right)\right) + z_i\phi\left(\gamma_i\right)\phi\left(\hat{\gamma}^{\backslash 3,i}\right)\right]\nonumber\\
&\quad\N\left(\gamma_i\big|\hat{\mu}^{\backslash 3,i}, \hat{\Sigma}^{\backslash 3,i}\right) \text{d} \gamma_i\nonumber\\
&= \left(1 - \phi\left(\hat{\gamma}^{\backslash 3,i}\right)\right)\int \gamma_i\left(1 - \phi\left(\gamma_i\right)\right)\N\left(\gamma_i\big|\hat{\mu}^{\backslash 3,i}, \hat{\Sigma}^{\backslash 3,i}\right) \text{d} \gamma_i\nonumber\\
&\quad + \phi\left(\hat{\gamma}^{\backslash 3,i}\right)\int \gamma_i\phi\left(\gamma_i\right)\N\left(\gamma_i\big|\hat{\mu}^{\backslash 3,i}, \hat{\Sigma}^{\backslash 3,i}\right) \text{d} \gamma_i\nonumber\\
&= \left(1 - \phi\left(\hat{\gamma}^{\backslash 3,i}\right)\right)\left[\hat{\mu}^{\backslash 3,i} - \int\gamma_i\phi\left(\gamma_i\right)\N\left(\gamma_i\big|\hat{\mu}^{\backslash 3,i}, \hat{\Sigma}^{\backslash 3,i}\right) \text{d} \gamma_i\right]\nonumber\\
&\quad + \phi\left(\hat{\gamma}^{\backslash 3,i}\right)\int \gamma_i\phi\left(\gamma_i\right)\N\left(\gamma_i\big|\hat{\mu}^{\backslash 3,i}, \hat{\Sigma}^{\backslash 3,i}\right) \text{d} \gamma_i \label{eq:update_prior_pz_first_moment_gamma}
\end{align}
Again we turn to \citep{rasmussen2006a} for the analytical solution of the above integrals
\begin{align}
\int \gamma_i\phi\left(\gamma_i\right)\N\left(\gamma_i\big|\hat{\mu}^{\backslash 3,i}, \hat{\Sigma}^{\backslash 3,i}\right) \text{d} \gamma_i &= \phi\left(c_{3,i}\right)\hat{\mu}^{\backslash 3,i} + \phi\left(c_{3,i}\right)\frac{\hat{\Sigma}^{\backslash 3,i}\N\left(c_{3,i}\big|0, 1\right)}{\phi\left(c_{3,i}\right)\sqrt{1 + \hat{\Sigma}^{\backslash 3,i}}}\nonumber\\
&= \phi\left(c_{3,i}\right)\hat{\mu}^{\backslash 3,i} + \phi\left(c_{3,i}\right)d_{3,i}, \label{eq:first_moment_phi_gauss}
\end{align}
where we have defined
\begin{align}
d_{3,i} \triangleq \frac{\hat{\Sigma}^{\backslash 3,i}\N\left(c_{3,i}\big|0, 1\right)}{\phi\left(c_{3,i}\right)\sqrt{1 + \hat{\Sigma}^{\backslash 3,i}}}.
\end{align}
Plugging eq. \eqref{eq:first_moment_phi_gauss} back into eq. \eqref{eq:update_prior_pz_first_moment_gamma} and simplifying yields
\begin{align}
G_1 &= \left(1 - \phi\left(\hat{\gamma}^{\backslash 3,i}\right)\right)\left[\left(1 - \phi\left(c_{3,i}\right)\right)\hat{\mu}^{\backslash 3,i}  - \phi\left(c_{3,i}\right)d_{3,i}\right]+ \phi\left(\hat{\gamma}^{\backslash 3,i}\right)\phi\left(c_{3,i}\right)\left[\hat{\mu}^{\backslash 3,i} + d_{3,i}\right]\nonumber\\
%
%
&= \left(1 - \phi\left(\hat{\gamma}^{\backslash 3,i}\right)\right)\left(1 - \phi\left(c_{3,i}\right)\right)\hat{\mu}^{\backslash 3,i}  - \left(1 - \phi\left(\hat{\gamma}^{\backslash 3,i}\right)\right)\phi\left(c_{3,i}\right)d_{3,i}+ Z_1\left[\hat{\mu}^{\backslash 3,i} + d_{3,i}\right]\nonumber\\
&= \left(G_0 - Z_1\right)\hat{\mu}^{\backslash 3,i}  - \left(1 - \phi\left(\hat{\gamma}^{\backslash 3,i}\right)\right)\phi\left(c_{3,i}\right)d_{3,i}+ Z_1\left[\hat{\mu}^{\backslash 3,i} + d_{3,i}\right]\nonumber\\
%
%
%
%
&= G_0\hat{\mu}^{\backslash 3,i}   +\left(2 Z_1 - \phi\left(c_{3,i}\right)\right)d_{3,i}\nonumber\\
\end{align}
Thus, the first moment w.r.t. $\gamma_{i,t}$ is given by $\E\left[\gamma_{i,t}\right] = G1/G0$.
\\
\\
Similarly, we compute the second moment w.r.t. $\gamma_i$
\begin{align}
G_2 &= \sum_{z_i}\int \gamma_i^2 f_{3,i}\left(z_i, \gamma_i\right)Q^{\backslash 3,i}\left(z_i, \gamma_i\right) \text{d} \gamma_i\nonumber\\
&= \left(1 - \phi\left(\hat{\gamma}^{\backslash 3,i}\right)\right)\int \gamma_i^2\left(1 - \phi\left(\gamma_i\right)\right)\N\left(\gamma_i\big|\hat{\mu}^{\backslash 3,i}, \hat{\Sigma}^{\backslash 3,i}\right) \text{d} \gamma_i\nonumber\\
&\quad + \phi\left(\hat{\gamma}^{\backslash 3,i}\right)\int \gamma_i^2\phi\left(\gamma_i\right)\N\left(\gamma_i\big|\hat{\mu}^{\backslash 3,i}, \hat{\Sigma}^{\backslash 3,i}\right) \text{d} \gamma_i \label{eq:update_prior_pz_second_moment_gamma}
\end{align}
The solution to the above integrals are given by \citep{rasmussen2006a}
\begin{align}
&\int \gamma_i^2\phi\left(\gamma_i\right)\N\left(\gamma_i\big|\hat{\mu}^{\backslash 3,i}, \hat{\Sigma}^{\backslash 3,i}\right) \text{d} \gamma_i \nonumber\\
&=  \phi\left(c_{3,i}\right)\left[2\hat{\mu}^{\backslash 3,i}\left(\hat{\mu}^{\backslash 3,i} + d_{3,i}\right) + \left(\hat{\Sigma}^{\backslash 3,i} - \left(\hat{\mu}^{\backslash 3,i}\right)^2\right) - b_{3,i}\right]
\end{align}
where
\begin{align}
	b_{3,i} \triangleq \frac{ \left(\hat{\Sigma}^{\backslash 3,i}\right)^2 c_{3,i}\N\left(c_{3,i}\big|0, 1\right)}{\phi\left(c_{3,i}\right)\left(1 +  \hat{\Sigma}^{\backslash 3,i}\right)}
\end{align}
Furthermore, we define
\begin{align}
	w_{3,i} \triangleq 2\hat{\mu}^{\backslash 3,i}\left(\hat{\mu}^{\backslash 3,i} + d_{3,i}\right) + \left(\hat{\Sigma}^{\backslash 3,i} - \left(\hat{\mu}^{\backslash 3,i}\right)^2\right) - b_{3,i}
\end{align}
Substituting the above result back into eq. \eqref{eq:update_prior_pz_second_moment_gamma} and rearranging yields
\begin{align}
G_2 &= \left(1 - \phi\left(\hat{\gamma}^{\backslash 3,i}\right)\right)\left[\left(\hat{\mu}^{\backslash 3,i}\right)^2 + \hat{\Sigma}^{\backslash 3,i} - \phi\left(c_{3,i}\right)w_{3,i}\right] + \phi\left(\hat{\gamma}^{\backslash 3,i}\right)\phi\left(c_{3,i}\right)w_{3,i}\nonumber\\
&= \left(1 - \phi\left(\hat{\gamma}^{\backslash 3,i}\right)\right)\left[\left(\hat{\mu}^{\backslash 3,i}\right)^2 + \hat{\Sigma}^{\backslash 3,i} - \phi\left(c_{3,i}\right)w_{3,i}\right] + Z_1w_{3,i}\nonumber\\
\end{align}
Thus, the second moment is given by $\E\left[\gamma_{i,t}^2\right] = G2/G0$. Finally, the central moments of $Q^*$ then becomes
\begin{align}
\E\left[\gamma_{j,t}\right] = \frac{G_1}{G_0}, && \V\left[\gamma_{j,t}\right] = \frac{G_2}{G_0} - \frac{G_1^2}{G_0^2}, &&
\E\left[z_{j,t}\right] = \frac{Z_1}{G_0}.
\end{align}
These moments completely determine the distribution $Q^{3, \text{new}}$ and thus, we compute the updates for $f_{3,i}$ as follows
\begin{align}
\hat{\Sigma}^{\text{new}}_{3,i}&= \left[\V\left[\gamma_i\right]^{-1} - \left(\hat{\Sigma}^{\backslash 3,i}\right)^{-1}\right]^{-1},\\
\hat{\mu}^{\text{new}}_{3,i}&= \hat{\Sigma}^{\text{new}}_{3,i}\left[\V\left[\gamma_i\right]^{-1}\E\left[\gamma_i\right] - \left(\hat{\Sigma}^{\backslash 3,i}\right)^{-1}\hat{\mu}^{\backslash 3,i}\right],\\
\hat{\gamma}^{\text{new}}_{3,i} &= d\left(\phi\left(\E\left[z_i\right]\right), \hat{\gamma}^{\backslash 3,i}\right),
\end{align}

\section{Moments Computations for Probit Likelihood} \label{app:probit}
The purpose of this section is to describe the details of the EP approximation of the structured spike-and-slab prior with a probit likelihood. Using the notation described in Section \ref{sec:inference}, the probit likelihood term is given by
\begin{align}
f_{1,t}\left(\x_t\right) = p(\y_t\big|\x_t) = \prod_{n=1}^N \phi\left(y_{n,t}\A_{n,\cdot} \x_t\right).
\end{align}
First we compute the cavity distribution $Q^{\backslash 1,t,n}\left(\x\right)$ by removing the contribution of $\tilde{f}_{1_t,n}(\x)$ from the marginals of the joint approximation $Q$
\begin{align}
	Q^{\backslash 1, t, n}(\x_t) &= \frac{\mathcal{N}\left(\x_t\big|\m_t, \bm{V}_t\right)}{\tilde{f}_{1, t, n}(\x_t)} = K^{\backslash 1, t, n} \mathcal{N}\left(\x_t\big| \m^{\backslash 1, t, n}, \bm{V}^{\backslash 1, t, n}\right),
\end{align}
where
\begin{align}
\bm{V}^{\backslash 1, t, n} &= \left(\hat{\bm{V}}_{t}^{-1} - \hat{\bm{V}}_{1,t,n}^{-1}\right)^{-1},\\
\m^{\backslash 1, t, n} &=  \bm{V}^{\backslash 1, t, n}\left(\hat{\bm{V}}_{t}^{-1}\hat{\m}_{t} -\hat{\bm{V}}_{1,t}^{-1}\hat{\m}_{1,t,n} \right).
\end{align}
for diagonal matrices $\bm{V}_t$ and $\bm{V}^{\backslash 1, t, n}$.
The tilted distribution then becomes
\begin{align*}
	\hat{q}_{1, t, n}\left(\x_t\right) &= \frac{1}{z_{1, t, n}}\phi\left(y_{n,t}\A_{n,\cdot} \x_t\right) \mathcal{N}\left(\x_t\big| \m^{\backslash 1, t, n}, \bm{V}^{\backslash 1, t, n}\right),
\end{align*}
First we compute the normalization constant, which is given by
\begin{align}
	z_{1, t, n} &= \int \phi\left(y_{n,t}\A_{n,\cdot} \x_t\right) \mathcal{N}\left(\x_t\big| \m^{\backslash 1, t, n}, \bm{V}^{\backslash 1, t, n}\right) \dd \x_t\\ 
	&=  \int \phi\left(u\right) \mathcal{N}\left(u\big| a_{1, t, n}, b_{1, t, n}\right) \dd u\\
	&= \phi\left(c_{1, t, n}\right),
\end{align}
where $a_{1, t, n} = y_{n,t}\A_{n,\cdot}\m^{\backslash 1, t, n}$, $b_{1, t, n} = \A_{n,\cdot}\bm{V}^{\backslash 1, t, n}\A_{n,\cdot}^T$ and $c_{1, t, n} = \frac{a_{1, t, n}}{\sqrt{1 + b_{1, t, n}}}$. Since $y_{n,t} \in \left\lbrace -1, 1 \right\rbrace$, $y_{n,t}$ does not appear in the expression for $b_{1, t, n}$ due to square form. 
Define the row-vector $\tilde{\A}_n = y_{n,t} \A_{n, \cdot} \in \mathbb{R}^{1 \times D}$, then the first moment w.r.t. $x_{t,j}$ is given by

\begin{align}
	\mathbb{E}\left[x_{t,j}\right] &= \frac{1}{z_{1, t, n}} \int x_{t,j} \phi\left(y_{n,t}\A_{n,\cdot} \x_t\right) \mathcal{N}\left(\x_t\big| \m^{\backslash 1, t, n}, \bm{V}^{\backslash 1, t, n}\right) \dd \x_t\\
	&= \frac{1}{z_{1, t, n}} \int x_{t,j} \phi\left(\tilde{\A}_{n,\cdot} \x_t\right) \mathcal{N}\left(\x_t\big| \m^{\backslash 1, t, n}, \bm{V}^{\backslash 1, t, n}\right) \dd \x_t\\
	&= \frac{1}{z_{1, t, n}} \int x_{t,j} \int\phi\left(\tilde{\A}_{n,-j} \x_{-j} + \tilde{a}_{n,j} x_{t,j}\right) \mathcal{N}\left(\x_{t, -j}\big| \m^{\backslash 1, t, n}_{-j}, \bm{V}^{\backslash 1, t, n}_{-j}\right) \dd \x_{t,-j}\nonumber\\
	&\quad  \mathcal{N}\left(x_{t,j}\big| \m^{\backslash 1, t, n}_j, \bm{V}^{\backslash 1, t, n}_{jj}\right)\dd x_{t, j}
\end{align}
Performing a change of variable, $z = \tilde{A}_{n,-j} \x_{t, -j}$, reduces the inner integral to a one-dimensional integral and thus, the resulting two nested one-dimensional integrals can be solved using standard results for Gaussian integrals \cite{rasmussen2006a}. The resulting moment becomes:
\begin{align}
	\mathbb{E}\left[x_{t,j}\right]  &=  m^{\backslash 1, t, n}_j  +    \alpha_{1,t,n} \tilde{a}_{n,j} V^{\backslash 1, t, n}_{jj},
\end{align}
where we have defined $\alpha_{1,t,n} = \frac{ \mathcal{N}(z) }{ \sqrt{1 + b_{1,t,n}}\phi(z) }$. Therefore,
\begin{align}
	\mathbb{E}\left[\x_{t}\right]  &=  \m^{\backslash 1, t, n}  +    \alpha_{1,t,n} \cdot  \left(\tilde{\A}_{n,\cdot} \circ \text{diag}\left(\bm{V}^{\backslash 1, t, n}\right)\right).
\end{align}
Carrying out similar calculations for $\x\x^T$ yields
\begin{align}
	\mathbb{V}\left[\x_{t}\right]  &=  \text{diag}\left(\bm{V}^{\backslash 1, t, n}\right) \nonumber\\
	&\quad  -    \alpha_{1,t,n} \cdot \frac{\left(\tilde{\A}_{n, \cdot}\mathbb{E}\left[\x_{t}\right] + \alpha_{1,t,n}\right)}{1 + b_{1,t,n}} \left(\tilde{\A}_{n,\cdot} \circ \text{diag}\left(\bm{V}^{\backslash 1, t, n}\right)\right)\circ  \left(\tilde{\A}_{n,\cdot} \circ \text{diag}\left(\bm{V}^{\backslash 1, t, n}\right)\right).
\end{align}
Using these moments, we compute the updates for $\tilde{f}_{1, t, n }$ as follows
\begin{align}
\hat{\bm{V}}^{\text{new}}_{1, t, n} &= \left[\bm{V}\left[\x_t\right]^{-1} - \left(\bm{V}^{\backslash 1, t, n}\right)^{-1}\right]^{-1},\\
\hat{\m}^{\text{new}}_{1, t, n}&= \hat{\bm{V}}^{\text{new}}_{1, t, n} \left[\bm{V}\left[\x_t\right]^{-1}\E\left[\x_t\right] - \left(\bm{V}^{\backslash 1, t, n}\right)^{-1}\m^{\backslash 1,t,n}\right].
\end{align}

\section{On the Prior Mean and Variance of $\Gamma$} \label{appendix:prior}
\begin{figure}[tpb]
\subfigure[Expected sparsity vs. prior mean and variance]{\includegraphics[width = 0.5\textwidth]{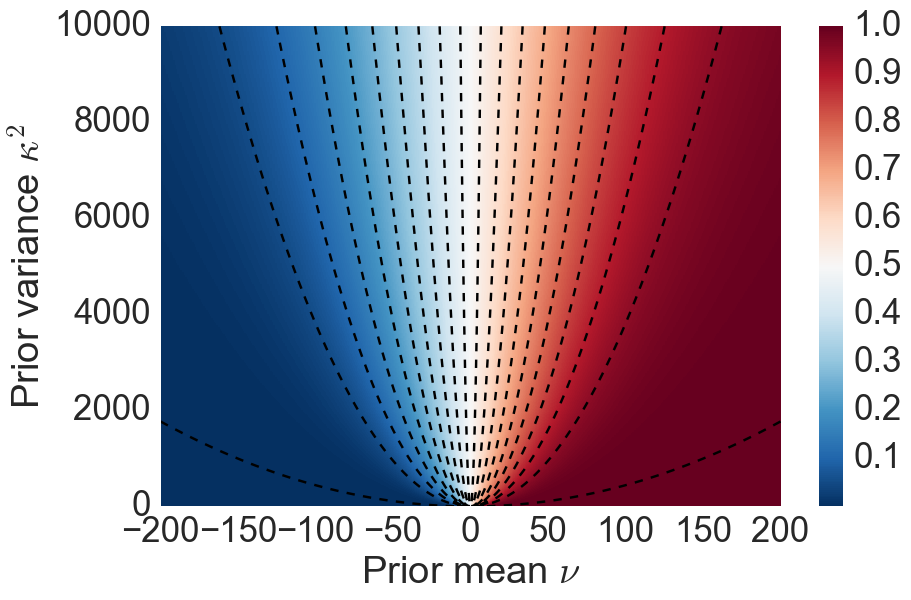}}
\subfigure[Distribution of $\phi(\gamma_i)$ for $p(z_i = 1) = \frac{1}{4}$]{\includegraphics[width = 0.5\textwidth]{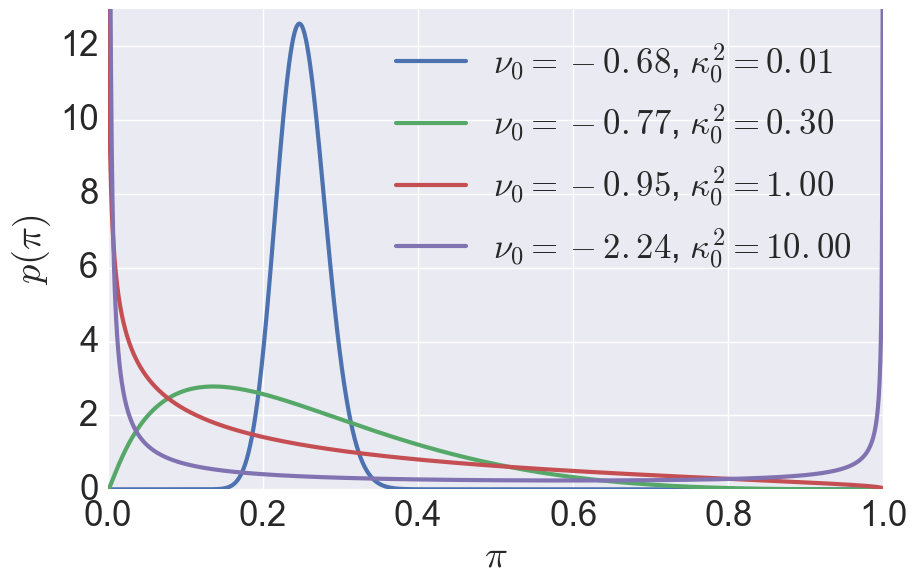}}
\caption{Properties of the prior distribution. (a) Marginal prior probability $p(z_i = 1)$ as a function of $(\nu_0, \kappa^2_0)$. The black dashed lines are isocontours. (b) Distribution of $\pi = \phi(\gamma_i)$ for 4 different pairs of $(\nu_0, \kappa^2_0)$, but for fixed value of $p(z_i = 1)$.}
\label{fig:app_kappa_prior}
\end{figure}

\begin{figure}[tpb]
\centering
\subfigure[$\x_1$]{\includegraphics[height = 2.5cm]{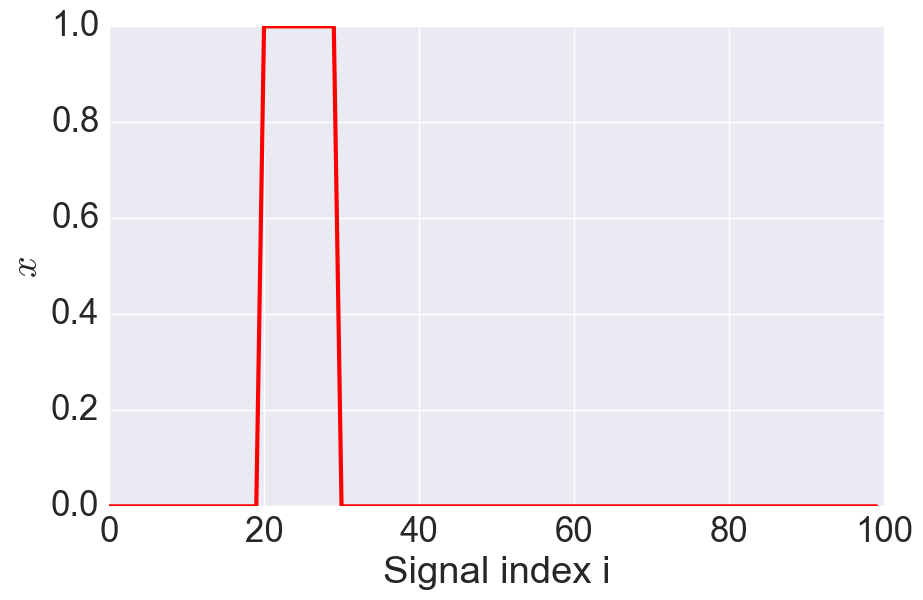}}
\subfigure[LPD for $\y_1$ (flat prior)]{\includegraphics[height = 2.5cm]{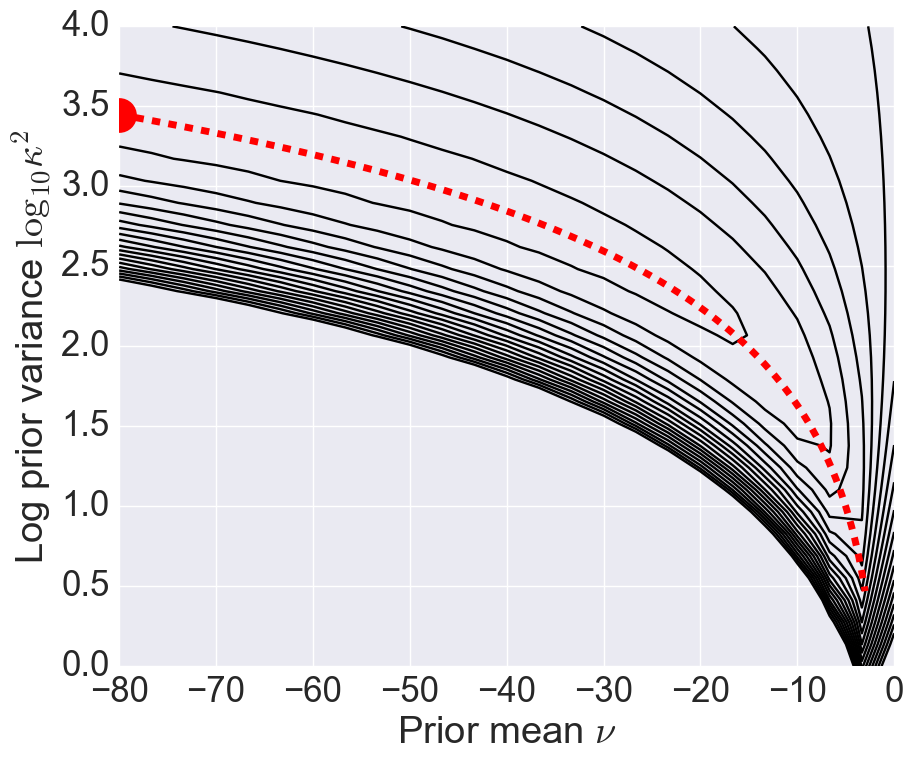}}
\subfigure[LPD for $\y_1$ (student t prior)]{\includegraphics[height = 2.5cm]{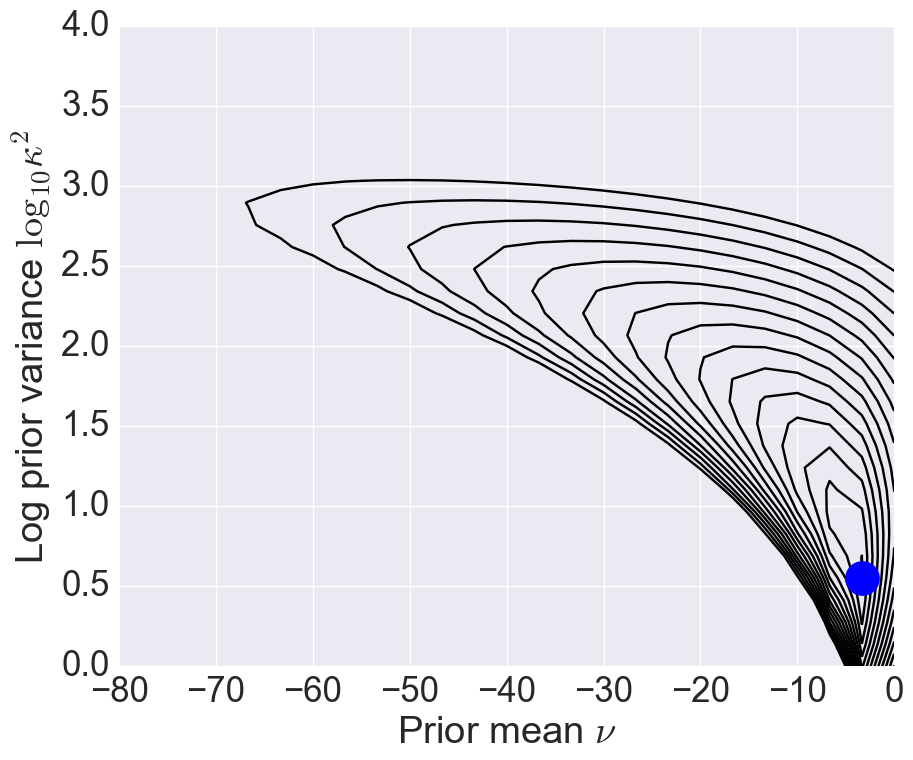}}
\subfigure[LPD for $\y_1$ (log-normal prior)]{\includegraphics[height = 2.5cm]{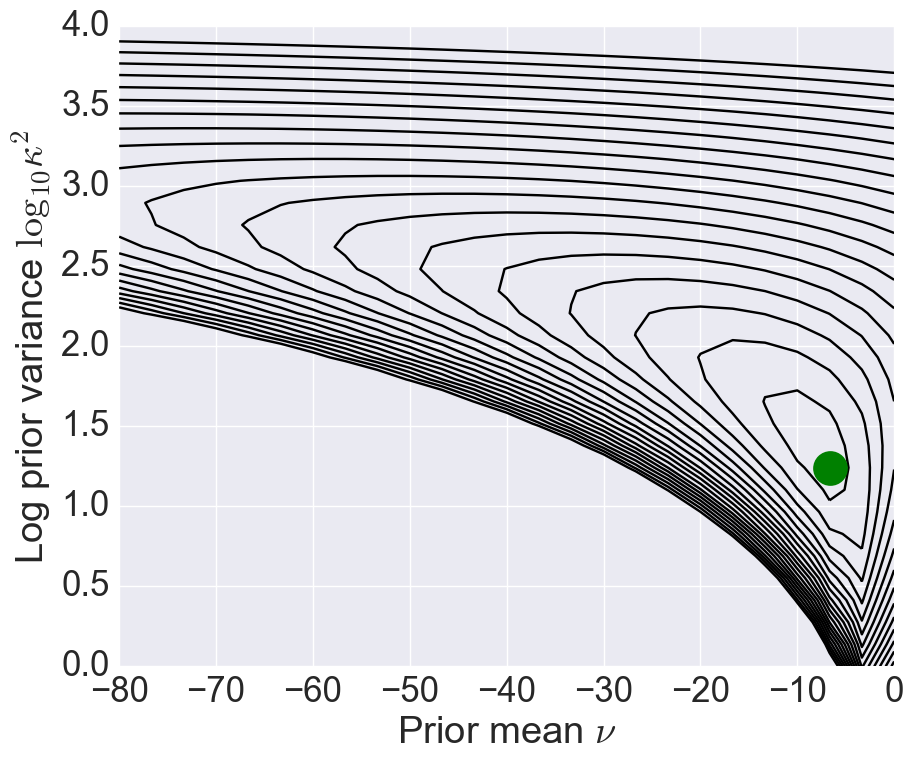}}\hfill\\
\subfigure[$\x_2$]{\includegraphics[height = 2.5cm]{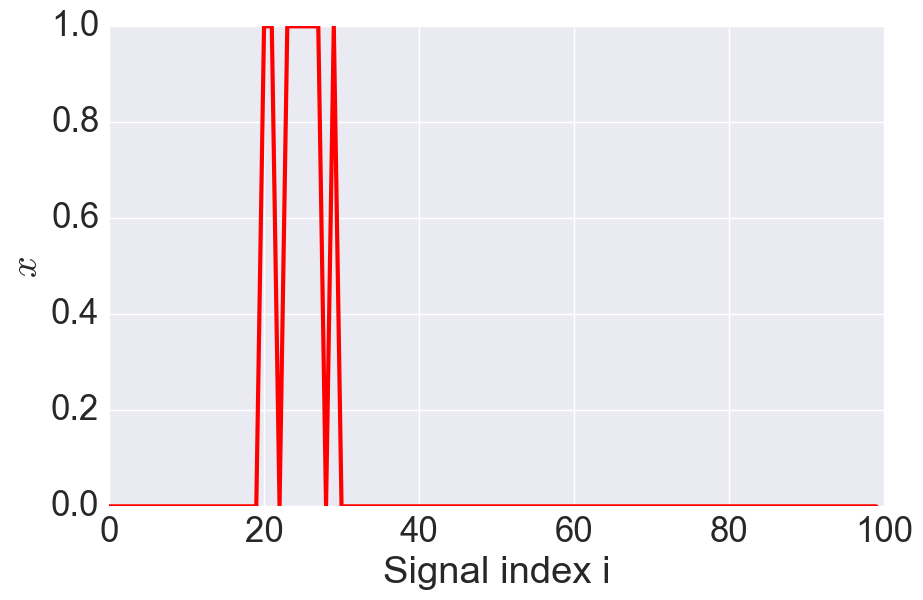}}
\subfigure[LPD for $\y_2$ (flat prior)]{\includegraphics[height = 2.5cm]{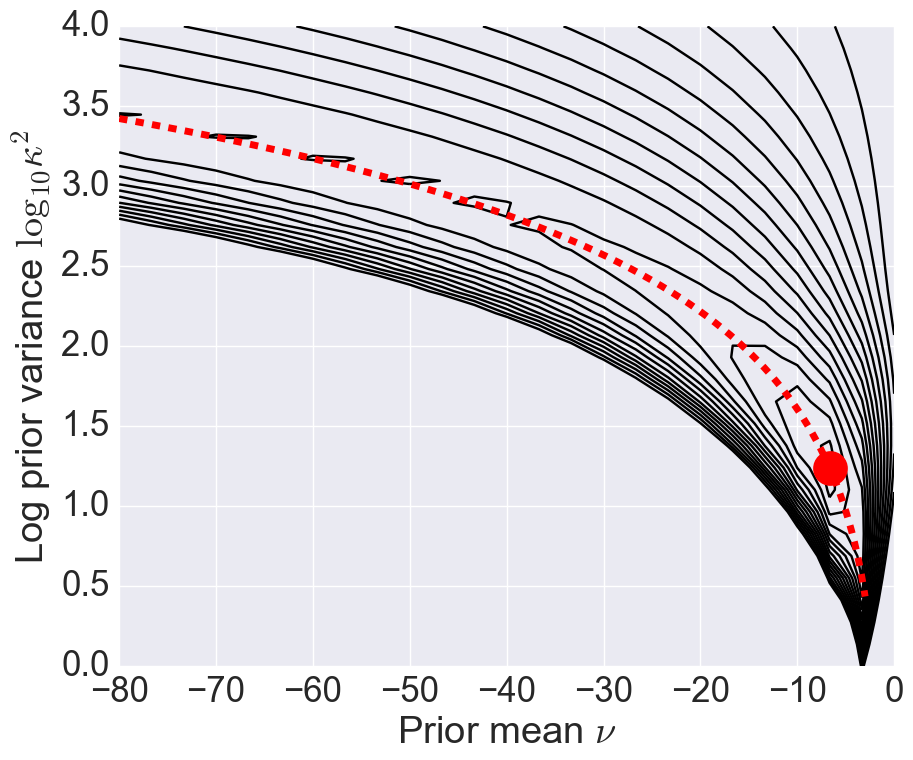}}
\subfigure[LPD for $\y_2$ (student t prior)]{\includegraphics[height = 2.5cm]{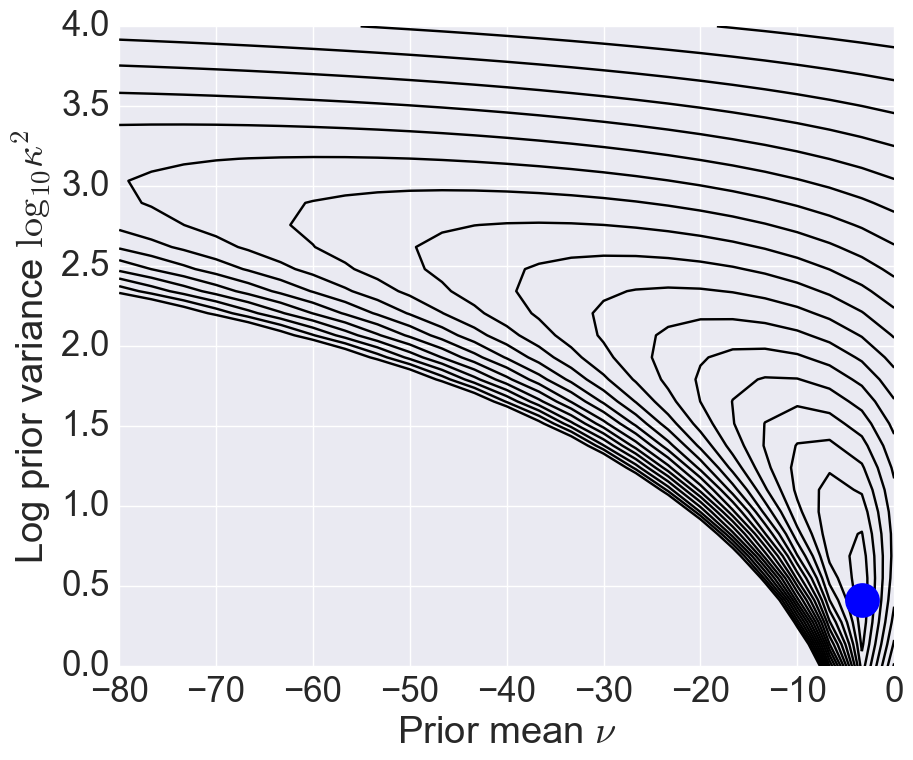}}
\subfigure[LPD for $\y_2$ (log-normal prior)]{\includegraphics[height = 2.5cm]{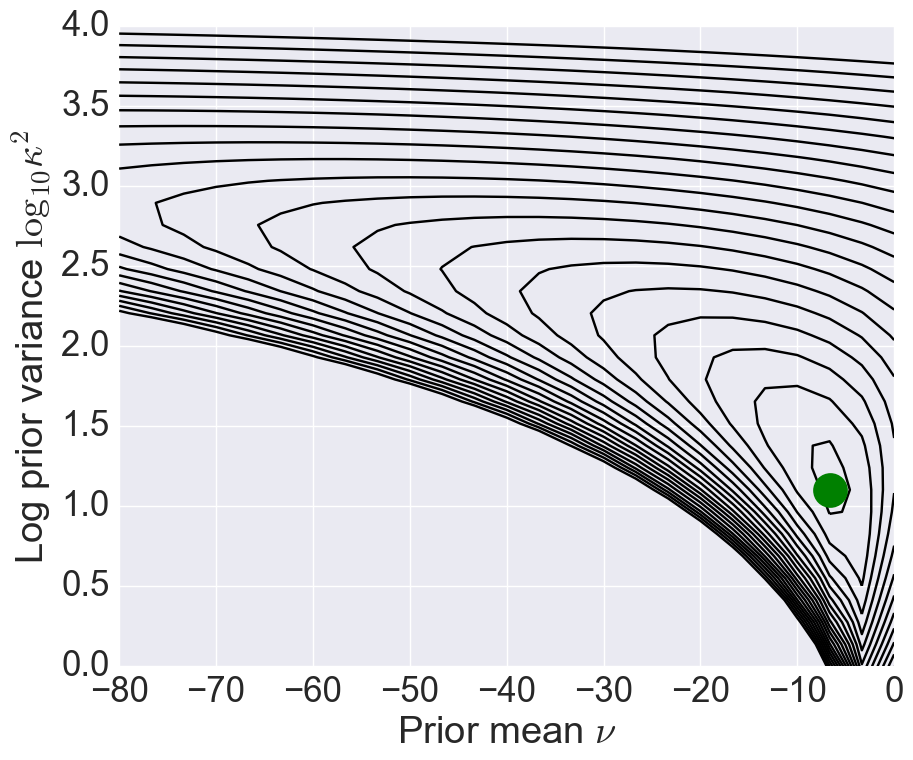}}
\caption{(a) Signal, where the support is contiguous. (b), (c), (d): Log posterior density for $\y_1 = \A\x_1 + \bm{\epsilon}$ with a flat prior, half student t prior (df = 4) and a log normal prior (mean 6, std. dev 3) for $\kappa_0$, respectively. (e ) Signal, where the support is not contiguous.(f), (g), (h): Log posterior density (LPD) for $\y_2 = \A\x_2 + \bm{\epsilon}$ with a flat prior, half student t prior (df = 4) and a log normal prior (mean 6, std. dev 3) for $\kappa_0$, respectively. The red dashed line shows a plot of the implicit function $\hat{p}_{ML} = p(z_i = 1) = \phi\left(\nu_0\, (1+\kappa^2_0)^{-\frac{1}{2}}\right)$ that intersects the maximum likelihood solution.}
\label{fig:app_kappa_marginal}
\end{figure}

\begin{figure}[tpb]
\centering
\subfigure[Posterior of $\bm{\Gamma}$ for $\y_1$]{\includegraphics[width = 0.24\textwidth]{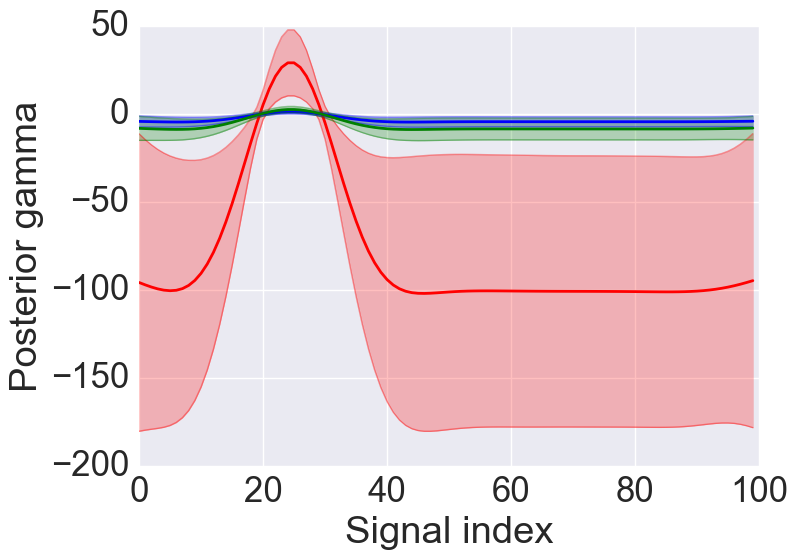}}
\subfigure[Posterior mean support for $\y_1$]{\includegraphics[width = 0.24\textwidth]{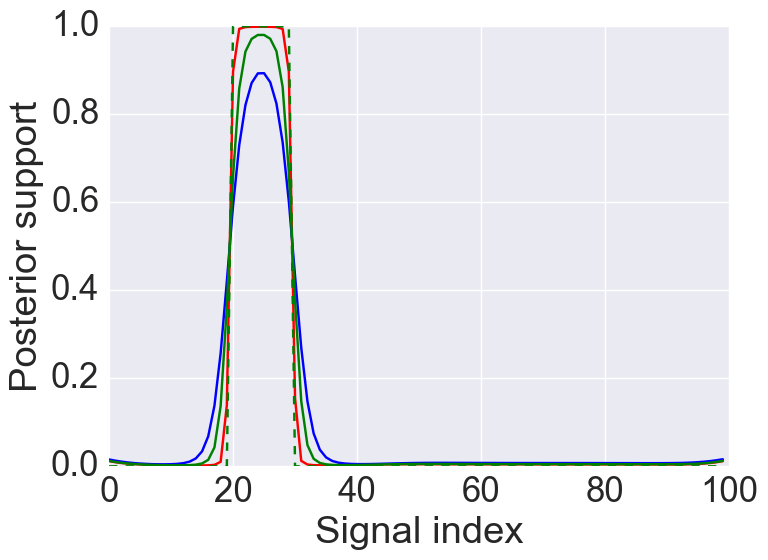}}
\subfigure[Posterior of $\bm{\Gamma}$ for $\y_2$]{\includegraphics[width = 0.24\textwidth]{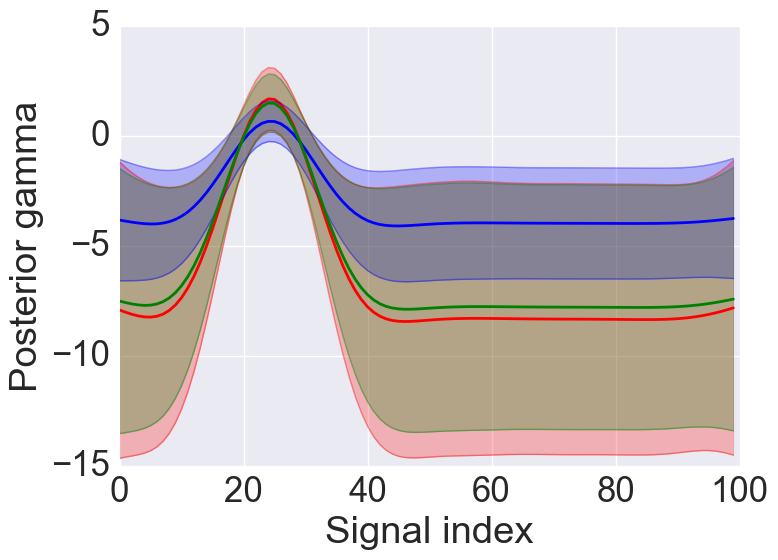}}
\subfigure[Posterior mean support for $\y_2$]{\includegraphics[width = 0.24\textwidth]{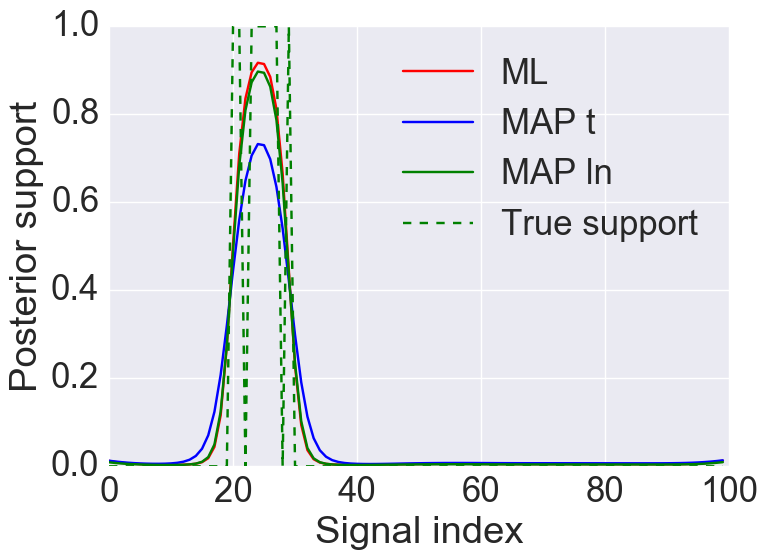}}
\caption{Posterior distributions for $\y_1$ and $\y_2$ for hyperparameter values indicated by the colors dots in Figure \ref{fig:app_kappa_marginal}. }
\label{fig:app_kappa_post}
\end{figure}

The purpose of this appendix is to elaborate on the interplay between the prior mean and the prior variance of $\bm{\Gamma}$. For this analysis we will assume that the $\bm{\Gamma}$ has constant mean $\bm{\mu}_0 = \nu_0 \, \bm{1}$ for $\nu_0 \in \mathbb{R}$, and covariance $\bm{\Sigma}_0 = \kappa^2_0 \bm{R}_0$, where $\bm{1} \in \mathbb{R}^D$ is a column vector of ones and $\bm{R}_0 \in \mathbb{R}^{D \times D}$ is a correlation matrix. Recall from eq. \eqref{eq:marginal_prior_prob} that the marginal prior probability of $z_i = 1$ is given by
\begin{align}
\hat{p} = p(z_i = 1) = \int p(z_i = 1\big|\gamma_i)p(\gamma_i) \text{d} \gamma_i = \int \phi(\gamma_i) \N\left(\gamma_i\big|\mu_i, \Sigma_{0, ii}\right) \text{d} \gamma_i = \phi\left(\frac{\nu_0}{\sqrt{1+\kappa^2_0}}\right). \label{eq:app_marginal_prior_prob}
\end{align}
It is seen from the above expression that the marginal expected sparsity level is controlled by $\nu_0$ and $\kappa^2_0$. Figure \ref{fig:app_kappa_prior}(a) shows the surface of $p(z_i = 1)$ as a function of $\nu_0$ and $\kappa^2_0$, where the black dashed isocontours confirm that the same level of marginal expected sparsity can be obtained for any combination of $(\nu_0, \kappa^2_0)$ that satisfies the relationship in eq. \eqref{eq:app_marginal_prior_prob} for some $\hat{p} \in \left(0, 1\right)$.
Also, note that the prior probability $\hat{p}$ is by definition equal to the expectation of $\phi(\gamma_i)$, that is $\hat{p} = \mathbb{E}_{p(\gamma_i)}\left[\phi(\gamma_i)\right]$. However, as $\phi$ is a monotonic function, we can derive the full distribution of $\pi = \phi(\gamma)$ through a change of variable as follows
\begin{align}
	p(\pi) = p_{\gamma}(\phi^{-1}(\pi))\big|\frac{\text{d} \phi^{-1}(\pi)}{\text{d} \pi}\big| = \mathcal{N}\left(  \phi^{-1}(\pi) \big| \nu_0, \kappa^2_0\right) \big|\frac{\text{d} \phi^{-1}(\pi)}{\text{d} \pi}\big| = \frac{ \mathcal{N}\left(  \phi^{-1}(\pi) \big| \nu_0, \kappa^2_0\right)}{\mathcal{N}\left( \phi^{-1}(\pi) \big | 0, 1\right)}.
\end{align}
Figure \ref{fig:app_kappa_prior}(b) shows a plot of the density of $\pi$ for 4 pairs of $(\nu_0, \kappa^2_0)$ that all satisfy $\hat{p} = \mathbb{E}\left[\pi\right] = \frac{1}{4}$. Thus, increasing $\kappa^2_0$ while keeping $\mathbb{E}\left[\pi\right]$ fixed pushes the mass of $p(\pi)$ to the boundary values. Informally, the distribution of $p(\pi)$ will approach a mixture of two Dirac distributions at 0 and 1 with weights $1 - \mathbb{E}\left[\pi\right]$ and $\mathbb{E}\left[\pi\right]$, respectively, for very large values of $\kappa^2_0$ relative to $\nu_0 \neq 0$.
In Section \ref{sec:Hyperparameter}, we discussed maximum likelihood among other methods for learning the hyperparameters of the structured spike-and-slab model. However, maximum likelihood learning of $\nu$ and $\kappa$ can in some instances give rise to the similar problems as encountered in maximum likelihood learning of logistic regression models on data sets, that are completely separated in one or more dimensions \citep{Gelman2008-wa}. The following small example illustrates the problem.
Consider an instance of $\y_1 = \A\x_1 + \bm{\epsilon}$, where $\x_1$ is the signal shown in Figure \ref{fig:app_kappa_marginal}(a) and where the signal to noise ratio is such that the true support of the signal can be recovered exactly. The dimensions of the forward model is $\A \in \mathbb{R}^{50 \times 100}$. Let $\mathbf{R}$ be the squared exponential kernel with lengthscale fixed to 8. Figure \ref{fig:app_kappa_marginal}(c) shows the surface of the marginal likelihood approximation as a function of $\nu_0$ and $\kappa^2_0$ while the remaining hyperparameters are kept fixed. The red dot indicates the maximum likelihood solution constrained to the domain shown in the figure. The red dashed line shows a plot of the implicit function $\hat{p}_{ML} = p(z_i = 1) = \phi\left(\frac{\nu_0}{\sqrt{1+\kappa^2_0}}\right)$ that intersects the maximum likelihood solution. It is clear that the likelihood surface has a ridge along the curve satisfying $\hat{p}_{ML} = \phi\left(\frac{\nu_0}{\sqrt{1+\kappa^2_0}}\right)$ and that the likelihood is increasing along that ridge as the magnitude of $\nu_0$ and $\kappa^2_0$ increase. Thus, the maximum likelihood solutions pushes to magnitude of $\nu_0$ and $\kappa^2_0$ to larger and larger values while keeping the sparsity level $\hat{p}_{ML}$ fixed and therefore, gradient-based optimization of the maximum likelihood w.r.t. $(\nu_0, \kappa^2_0)$ will never converge.
However, this problem only occurs when the support is separated as in Figure \ref{fig:app_kappa_marginal}(a). Figure \ref{fig:app_kappa_marginal}(f) shows the marginal likelihood approximation surface for $\y_2 = \A\x_2 + \bm{\epsilon}$, where $\x_2$ in Figure \ref{fig:app_kappa_marginal}(b). It is now seen that the maximum likelihood solution is well-defined within the interior of $\mathbb{R}^2$. The problem is easily fixed by imposing a weakly informative prior on $\kappa_0$ to ensure that the solution is always well-defined. To illustrate this, we re-run this experiment shown in Figure \ref{fig:app_kappa_marginal}(c) with two different priors on $\kappa_0$. Figures \ref{fig:app_kappa_marginal}(d)-(e) show the results for a standardized half student t prior with 4 degrees of freedom and a log-normal prior with mean 6 and standard deviation 3, respectively. Figures \ref{fig:app_kappa_marginal}(g)-(h) show the same plots for the signal $\x_2$. Figure \ref{fig:app_kappa_post} shows the resulting posterior distribution for both signals with and without priors distributions.

\vskip 0.2in
\bibliography{15-464}

\end{document}